\documentclass[12pt,onecolumn,twoside]{IEEEtran}

\usepackage{amssymb}
\usepackage{amsmath}
\usepackage{epsfig}
\usepackage{url}
\usepackage{booktabs}
\usepackage{multirow}

\usepackage{siunitx}

\usepackage{subfigure}
\usepackage{dsfont}
\usepackage{upgreek}

\usepackage{epstopdf}

\usepackage{bbm}

\usepackage{commath}

\def\dspace{\multiply\normalbaselineskip 160
            \divide\normalbaselineskip 100 \normalbaselines
            \csname @@normalbaselineskip\endcsname\normalbaselineskip}

\def\sspace{\multiply\normalbaselineskip 200
            \divide\normalbaselineskip 300 \normalbaselines
            \csname @@normalbaselineskip\endcsname\normalbaselineskip}

\def\1over4{{\textstyle{1\over 4}}}
\def\oneover8{{\textstyle{1\over 8}}}

\def\schi2{{\scriptscriptstyle {\chi^2}}}

\def\cQ{{ \cal Q }}
\def\cH{{ \cal H }}

\def\bh0{\mbox{\boldmath $h_0$}}

\def\bh{\mbox{\boldmath $h$}}

\def\bphi{\mbox{\boldmath $\phi$}}

\def\bz{\mbox{\boldmath $z$}}

\def\bmu{\mbox{\boldmath $\mu$}}

\def\bq{\mbox{\boldmath $q$}}
\def\bx{\mbox{\boldmath $x$}}

\def\bbeta{\mbox{\boldmath $\beta$}}

\def\etabold{\mbox{\boldmath $\eta$}}

\def\by{\mbox{\boldmath $y$}}

\def\bx{\mbox{\boldmath $x$}}

\def\b0{\textbf{0} }

\newtheorem{theorem}{Theorem}

\DeclareMathOperator{\cov}{cov}
\DeclareMathOperator{\ch}{ch}
\DeclareMathOperator{\diag}{diag}
\DeclareMathOperator{\Exp}{E}
\DeclareMathOperator{\Invchi2}{Inv-\chi^2}
\DeclareMathOperator{\NMSE}{NMSE}
\DeclareMathOperator{\rank}{rank}
\DeclareMathOperator{\Prob}{Pr}
\DeclareMathOperator{\sib}{sib}
\DeclareMathOperator{\gp}{gp}

\newcommand*{\bm}[1]{\ensuremath{\boldsymbol{#1}}}

\newcommand\Div[2]{\mathbbm{D}\del[1]{ #1 \,\Vert\, #2  }}

\def\tMAX{\textup{MAX}}
\def\tMIN{\textup{MIN}}
\def\troot{\textup{root}}
\def\tH{\textup{H}}
\def\tu{\textup{u}}
\def\td{\textup{d}}
\def\tL{\textup{L}}
\def\cA{\mathcal{A}}
\def\cT{\mathcal{T}}

\def\adots{\mathinner{\mkern1mu\raise1pt\vbox{\kern7pt\hbox{.}}
                      \mkern2mu\raise4pt\hbox{.}
                      \mkern2mu\raise7pt\hbox{.}\mkern1mu}}

\def\R{\mbox{I{\kern-0.2em}R}}

\def\tr{\mathop{\rm tr}\nolimits}

\def\boxit#1{\vbox{\hrule\hbox{\vrule\kern3pt
             \vbox{\kern3pt#1\kern3pt}\kern3pt\vrule}\hrule}}

\def\PSNR{\mbox{PSNR}}

\def\etabold{\mbox{\boldmath $\eta$}}

\def\bzero{\mathbf{0} }
\def\bone{\mathbbm{1}}

\def\sbq{{ \scriptstyle \boldsymbol{q}  }}

\def\sleaf{{\scriptstyle {\rm leaf}  }}
\def\ch{\mathop{\rm ch}\nolimits}

\def\scA{{\scriptstyle \mathcal{A}  }}

\def\bnu{\mbox{\boldmath $\nu$}}

\usepackage[acronym]{glossaries}

\makeglossaries

\newacronym{AMP}{\textsc{amp}}{approximate message passing}
\newacronym{GAMP}{\textsc{gamp}}{generalized approximate message passing}
\newacronym{CT}{\textsc{ct}}{computed tomography}
\newacronym{CPU}{\textsc{cpu}}{central processing unit}
\newacronym{DWT}{\textsc{dwt}}{discrete wavelet transform}
\newacronym{E}{\textsc{e}}{expectation}
\newacronym{EM}{\textsc{em}}{expectation-maximization}
\newacronym{iid}{i.i.d.}{independent, identically distributed}
\newacronym{FBP}{\textsc{fbp}}{filtered backprojection}
\newacronym{fMRI}{f{\large \textsc{mri}}}{functional magnetic resonance imaging}
\newacronym{KL}{\textsc{kl}}{Kullback-Leibler}
\newacronym{NDE}{\textsc{nde}}{nondestructive evaluation}
\newacronym{M}{\textsc{m}}{maximization}
\newacronym{MAC}{\textsc{mac}}{mass attenuation coefficient}
\newacronym{MAP}{\textsc{map}}{maximum \emph{a posteriori}}
\newacronym{MMSE}{\textsc{mmse}}{minimum mean-square error}
\newacronym{NP-hard}{\textsc{np}-hard}{non-deterministic
  polynomial-time hard} 
\newacronym{pdf}{pdf}{probability density function}
\newacronym{pmf}{pmf}{probability mass function}
\newacronym{turbo-AMP}{turbo-\textsc{amp}}{}
\newacronym{FPC_AS}{\textsc{fpc}$_{\textsc{as}}$}{fixed-point continuation active set}
\newacronym{PSNR}{\textsc{psnr}}{peak signal-to-noise ratio}
\newacronym{GPSR}{\textsc{gpsr}}{gradient-projection for sparse reconstruction}
\newacronym{DORE}{\textsc{dore}}{double overrelaxation thresholding}
\newacronym{NIHT}{\textsc{niht}}{normalized iterative hard thresholding}
\newacronym{MB-IHT}{\textsc{mb-iht}}{model-based iterative hard
  thresholding}
\newacronym{MCMC}{\textsc{mcmc}}{Markov chain Monte Carlo}
\newacronym{MP-EM}{\textsc{mp-em}}{max-product \textsc{em}}
\newacronym{MP-EM_OPT}{\textsc{mp-em}$_{\textsc{opt}}$}{max-product \textsc{em}}
\newacronym{MSE}{\textsc{mse}}{mean-square error}
\newacronym{NMSE}{\textsc{nmse}}{normalized mean-square error}
\newacronym{SNR}{\textsc{snr}}{signal-to-noise ratio}
\newacronym{HMT}{\textsc{hmt}}{hidden Markov tree}
\newacronym{VB}{\textsc{vb}}{variational Bayesian}

\title{A Max-Product EM Algorithm for\\ Reconstructing Markov-tree Sparse
  Signals\\ from Compressive Samples}

\author{Zhao Song and Aleksandar Dogand\v{z}i\'c,~\IEEEmembership{Senior Member,~IEEE}
\thanks{A portion of this work was presented at the SPIE
    Optics+Photonics Symposium, San Diego, CA, August 2012.}
\thanks{The authors are with
the Department of Electrical and Computer Engineering, Iowa State
University, Ames, IA 50011 USA (email: zhaosong@iastate.edu;
ald@iastate.edu).} }

\begin{document}
\maketitle

\vspace{-0.25in}

\begin{abstract}
  We propose a Bayesian \gls{EM} algorithm for reconstructing
  Markov-tree sparse signals via belief propagation.  The measurements
  follow an underdetermined linear model where the
  regression-coefficient vector is the sum of an unknown approximately
  sparse signal and a zero-mean white Gaussian noise with an unknown
  variance.  The signal is composed of large- and small-magnitude
  components identified by binary state variables whose probabilistic
  dependence structure is described by a Markov tree. Gaussian priors
  are assigned to the signal coefficients given their state variables
  and the Jeffreys' noninformative prior is assigned to the noise
  variance.  Our signal reconstruction scheme is based on an \gls{EM}
  iteration that aims at maximizing the posterior distribution of the
  signal and its state variables given the noise variance.   We
  construct the missing data for the \gls{EM} iteration so that the
  complete-data posterior distribution corresponds to a \gls{HMT}
  probabilistic graphical model that contains no loops and implement
  its \gls{M} step via a max-product algorithm.  This \gls{EM}
  algorithm estimates the vector of state variables \emph{as well as}
  solves iteratively a linear system of equations to obtain the
  corresponding signal estimate.  We select the noise variance so that
  the corresponding estimated signal and state variables obtained upon
  convergence of the \gls{EM} iteration have the largest marginal
  posterior distribution.  We compare the proposed and existing
  state-of-the-art reconstruction methods via signal and image
  reconstruction experiments.
\end{abstract}

\glsresetall



%

\dspace 12pt

\section{INTRODUCTION}
\label{sec:introduction}

\glsresetall
\glsunset{CPU}
\glsunset{turbo-AMP}
\glsunset{MP-EM_OPT}

\noindent
The advent of compressive sampling (compressed sensing) in the past
few years has sparked research activity in sparse signal
reconstruction, whose main goal is to estimate the \emph{sparsest} $p
\times 1$ signal coefficient vector $\bm{s}$ from the $N \times 1$
measurement vector $\by$ satisfying the following underdetermined
system of linear equations: 
\begin{equation}
  \label{eq:CS}
  \bm{y} = H \bm{s}
\end{equation}
where $H$ is an $N \times p$ \emph{sensing matrix} and $N \leq p$.

\IEEEpubidadjcol

A tree dependency structure is exhibited by the wavelet coefficients
of many natural images \cite{CrouseNowakBaraniuk,RombergChoiBaraniuk,
  CevherIndykCarinBaraniuk,HeCarin,HeChenCarin,SomSchniter,BaraniukCevherDuarteHegde}
(see also Fig.~\ref{fig:waveletcoefficientclustering} and
\cite[Fig. 2]{CevherIndykCarinBaraniuk}) as well as one-dimensional
signals
\cite{CrouseNowakBaraniuk,LuKimPearlman,BaraniukCevherDuarteHegde}.  A
probabilistic Markov tree structure has been introduced 
in \cite{CrouseNowakBaraniuk}
to model the
statistical dependency between the state variables of wavelet
coefficients. An approximate belief
propagation algorithm has been first applied to compressive sampling
by Baron, Sarvotham, and Baraniuk in \cite{BaronSarvothamBaraniuk},
which employs sparse Rademacher sensing matrices for Bayesian signal
reconstruction.  Donoho, Maleki, and Montanari
\cite{DonohoMalekiMontanari} simplified the sum-product algorithm by
approximating messages using a Gaussian distribution specified by
two scalar parameters, leading to their approximate message passing
\gls{AMP} algorithm. Following the \gls{AMP} framework, Schniter
\cite{Schniter} proposed a \gls{turbo-AMP} structured sparse signal
recovery method based on loopy belief propagation and turbo
equalization and applied it to reconstruct one-dimensional signals;
\cite{SomSchniter} applied the \gls{turbo-AMP} approach to reconstruct
compressible images.   A \gls{GAMP} algorithm that generalizes the
\gls{AMP} algorithm to arbitrary input and output channels and
incorporates both max-sum and sum-product loopy belief propagation
separately is proposed in \cite{Rangan}.  However, the above
references do not employ the exact form of the messages and also have
the following limitations: \cite{BaronSarvothamBaraniuk} relies on
sparsity of the sensing matrix, the methods in
\cite{BaronSarvothamBaraniuk, DonohoMalekiMontanari, Rangan} apply to
unstructured signals only, and the \gls{turbo-AMP} approach in
\cite{SomSchniter} and \cite{Schniter} needs sensing matrices to have
approximately \gls{iid} elements, see
\cite[Section.~III-C]{SomSchniter}.  Indeed, \gls{turbo-AMP} is
sensitive to the presence of correlations among the elements of the
sampling matrix and performs poorly if these correlations are
sufficiently high and if norms of the columns or rows of the sampling
matrix are sufficiently variable.

 In \cite{HeCarin} and \cite{HeChenCarin},
\gls{MCMC} and \gls{VB} schemes are used to reconstruct images that
follow probabilistic Markov tree structure from linear measurements;
however, \cite{HeCarin} and \cite{HeChenCarin} did not report
large-scale examples: these schemes are computationaly
demanding and do not scale with increasing dimensionality of the
reconstruction problem.

In this paper, we combine the hierarchical measurement model in
\cite{FigueiredoNowak} with a Markov tree prior on the binary state
variables that identify the large- and small-magnitude signal
coefficients and develop a Bayesian \gls{MAP} \gls{EM} signal
reconstruction scheme that aims at maximizing the posterior
distribution of the signal and its state variables given the noise
variance, where the \gls{M} step employs a max-product belief
propagation algorithm.  Unlike the \gls{turbo-AMP} scheme in
\cite{SomSchniter} and \cite{Schniter}, our reconstruction scheme
\emph{does not} require sensing matrices to have approximately
\gls{iid} elements and can handle correlations among these elements.
Unlike the previous work, we \emph{do not} approximate the message
form in our belief propagation scheme. Indeed, the \gls{M} step of our
\gls{EM} algorithm is exact because the expected complete-data
posterior distribution that we maximize in the \gls{M} step
corresponds to the \gls{HMT} graphical model that contains no loops. In
\cite{sd12}, we proposed a similar \gls{EM} algorithm for a random
signal model \cite{QiuDogandzic12b} with a purely sparse vector of
signal coefficients and a noninformative prior on this component given
the binary state variables. We apply a grid search to select the noise
variance so that the estimated signal and state variables have the
largest marginal posterior distribution.

In Section~\ref{sec:measuremodel}, we introduce our measurement and
prior models. We assume that the Markov tree prior distribution is
\emph{known}. To reduce the number of tuning
parameters for the tree prior, we further assume that these parameters
do not change between Markov tree levels. This is in contrast to other
approaches (e.g., \cite{HeCarin}, \cite{SomSchniter}, and
\cite{HeChenCarin}), which learn the Markov tree parameters from the
measurements and allow their variation across the tree levels, see
also the discussions in Sections~\ref{subsubsec:2dexamplesrm} and
\ref{Conclusion}.  Section~\ref{sec:BayesianEM} describes the proposed
\gls{EM} algorithm and establishes its properties; the implementation
of the \gls{M} step via the max-product algorithm is presented in
Section~\ref{sec:BP}.  The selection of the noise variance parameter
is discussed in Section~\ref{sec:sigma2choice}.  Numerical simulations
in Section~\ref{sec:example} compare reconstruction performances of
the proposed and existing methods.%

We introduce the notation: $I_n$ and $\bzero_{n \times 1}$ denote the
identity matrix of size $n$ and the $n \times 1$ vector of zeros,
respectively; ``$^T$'', $\det(\cdot)$, and $\left\| \cdot \right\|_p$ are the
transpose, determinant, and $\ell_p$ norm, respectively; 
$\mathcal{N}(\bx | \bmu, \Sigma)$
denotes the \gls{pdf} of a multivariate Gaussian random vector $\bx$
with mean $\bmu$ and covariance matrix $\Sigma$; $\Invchi2( \sigma^2 |
\nu, \sigma_0^2 )$ denotes the \gls{pdf} of a scaled inverse
chi-square distribution with $\nu$ degrees of freedom and a scale
parameter $\sigma_0^2$, see \cite[App. A]{Gelmanetal};
$\Div{p(\bm{x})}{q(\bm{x})}$ denotes the \gls{KL} divergence from
\gls{pdf} $p(\bm{x})$ to \gls{pdf} $q(\bm{x})$ \cite[Sec.
2.8.2]{Murphy}, \cite[Sec. 8.5]{CoverThomas}; $|\cT|$ is the
cardinality of the set $\cT$; $\upsilon(\cdot)$ is an invertible
operator that transforms the two-dimensional matrix element indices
into one-dimensional vector element indices.
Finally,
$\rho_H$ denotes the largest singular value of a matrix $H$, also
known as the spectral norm of $H$, and
``$\odot$'' denotes the Hadamard (elementwise) product.

\section{Measurement and Prior Models}
\label{sec:measuremodel}

\noindent
We model an $N \times 1$ real-valued measurement vector $\by$ using
the standard additive white Gaussian noise measurement model with the
likelihood function given by the following \gls{pdf}
\cite{CevherIndykCarinBaraniuk,SomSchniter}:
\begin{equation}
  \label{eq:awgn}
  p_{\bm{y} | \bm{s}, \sigma^2}( \bm{y} | \bm{s}, \sigma^2 )
  = \mathcal{N}(  \bm{y} |  H \bm{s}, \sigma^2 I_N )
\end{equation}
where $H$ is an $N \times p$ real-valued sensing matrix with $\rank(H)
= N$ satisfying the spectral norm condition
\begin{equation}
  \label{eq:spectralnormofH}
  \rho_H = 1
\end{equation}
$\bm{s} = [ s_1, s_2, \ldots, s_p ]^T$ is an unknown $p \times 1$
real-valued signal coefficient vector, and $\sigma^2$ is the unknown
noise variance. We assume \eqref{eq:spectralnormofH} without loss of
generality because it is easily satisfied by appropriate scaling of
the sensing matrix, measurements, and noise variance,\footnote{\label{scaling}For a
  generic sensing matrix $H'$ with $\rho_{H'} \neq 1$, data vector
  $\bm{y}'$ and noise variance $(\sigma^2)'$, this scaling is
  performed as follows: $H = H' / \rho_{H'}, \bm{y} = \bm{y}' /
  \rho_{H'}$, and $\sigma^2 = (\sigma^2)' / \rho_{H'}^2$, which
  guarantees that the new sensing matrix $H$ satisfies
  \eqref{eq:spectralnormofH}.} provided that the spectral norm of the
sensing matrix is easy to determine, see also footnote
\ref{adaptivestepsize} for comments on the case where the spectral
norm of the sensing matrix cannot be easily determined or estimated.

We adopt the Jeffreys' noninformative prior for the variance component
$\sigma^2$:
\begin{equation}
  \label{eq:psigma2}
  p_{\sigma^2}(\sigma^2) \propto (\sigma^2)^{-1}.
\end{equation}

Define the vector of binary state variables $\bq = [q_1, q_2, \dotsc,
q_p]^T \in \{0, 1\}^p$ that determine if the magnitudes of the signal
components $s_i, \, i=1,2,\ldots,p$ are small ($q_i=0$) or large
($q_i=1$).  Assume that $s_i$ are conditionally independent given
$q_i$ and assign the following prior \gls{pdf} to the signal coefficients:
\begin{subequations}
  \label{eq:sgivenqprior0}
  \begin{IEEEeqnarray}{rCl}
    \label{eq:sgivenqprior}
    p_{\bm{s} | \bm{q}, \sigma^2} (\bm{s} | \bm{q}, \sigma^2)
    =  \prod_{i = 1}^p && [\mathcal{N}( s_i | 0, \gamma^2 \sigma^2 )]^{q_i}
     [\mathcal{N}( s_i | 0, \epsilon^2 \sigma^2 )]^{1-q_i}
  \end{IEEEeqnarray}
  where $\gamma^2$ and $\epsilon^2$ are known positive constants and,
  typically, $\gamma^2 \gg \epsilon^2$. Hence, the large- and
  small-magnitude signal coefficients $s_i$ corresponding to $q_i=1$
  and $q_i=0$ are modeled as zero-mean Gaussian random variables with
  variances $\gamma^2 \sigma^2$ and $\epsilon^2 \sigma^2$,
  respectively.  Consequently, $\gamma^2$ and $\epsilon^2$ are
  relative variances (to the noise variance $\sigma^2$) of the large-
  and small-magnitude signal coefficients.  Equivalently,
  \begin{equation}
    \label{eq:sgivenqprior2}
    p_{\bm{s} | \bm{q}, \sigma^2}(\bm{s} | \bm{q}, \sigma^2) = \mathcal{N}( \bm{s} |
    \mathbf{0}_{p \times 1}, \sigma^2 D(\bm{q}) )
  \end{equation}
  where
\begin{equation}
  D(\bq) = \diag\bigl\{ (\gamma^2)^{q_1}  (\epsilon^2)^{1-q_1}, (\gamma^2)^{q_2}  (\epsilon^2)^{1-q_2}, \dotsc, (\gamma^2)^{q_p}  (\epsilon^2)^{1-q_p}
  \bigr\}.
 \end{equation}
\end{subequations}

\begin{figure}
\centering
\subfigure[]{\includegraphics[width=0.45\linewidth]{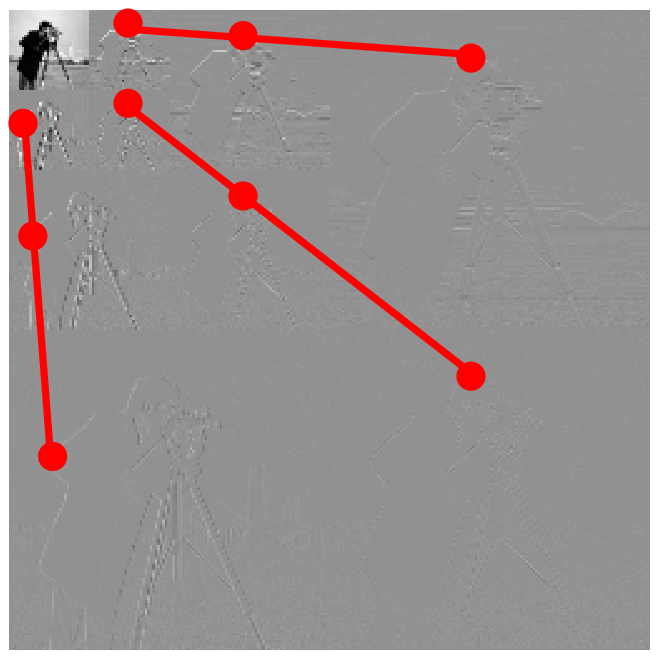}
  \label{fig:waveletcoefficientclustering}
} \hfil 
\subfigure[]{\includegraphics[width=0.445\linewidth]{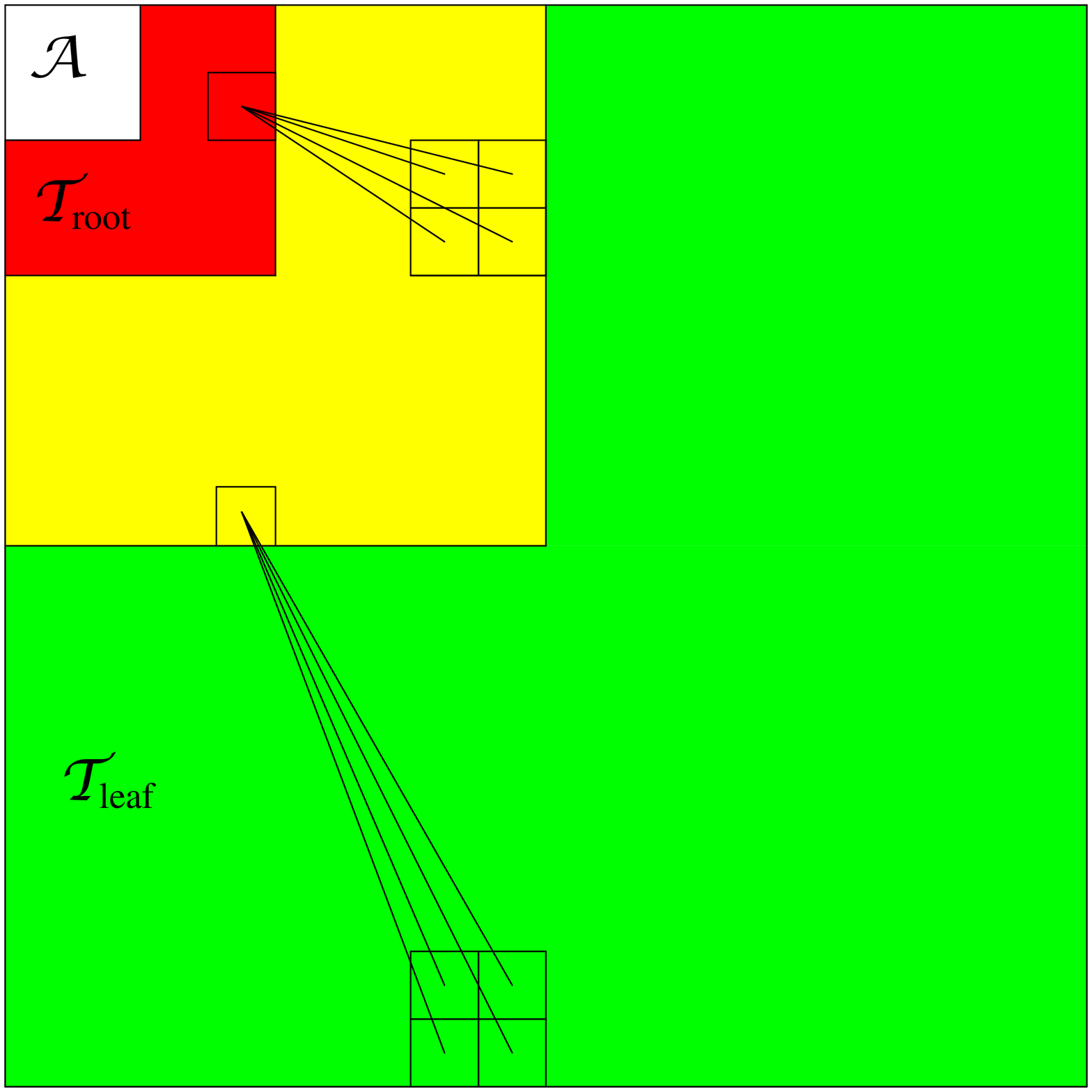}
   \label{fig:typesofwaveletdecompositioncoefficients}
}

\caption{(a) Clustering of significant discrete wavelet transform
  coefficients of a compressed `Cameraman' image and (b) types of
  wavelet decomposition coefficients: approximation, root, and leaf,
  whose sets are denoted by $\cA, \cT_{\troot}$, and $\cT_{\sleaf}$,
  respectively.}
\end{figure}

We now introduce the Markov tree prior \gls{pmf} on the state
variables $q_i$ \cite{CrouseNowakBaraniuk,SomSchniter}.  To make this
probability model easier to understand, we focus on the image
reconstruction scenario where the elements of $\bm{s}$ are the
two-dimensional \gls{DWT} coefficients of the underlying image that we
wish to reconstruct. Hence, we introduce two-dimensional signal
element indices $(i_1, i_2)$. Recall that the conversion operator
$\upsilon( \cdot )$ is invertible; hence, there is a one-to-one
correspondence between the corresponding one- and two-dimensional
signal element indices. A parent wavelet coefficient with a
two-dimensional position index $(i_1, i_2)$ has four children in the
finer wavelet decomposition level with two-dimensional indices $(2 i_1
- 1, 2 i_2 - 1)$, $(2 i_1 - 1, 2 i_2)$, $(2 i_1, 2 i_2 - 1)$ and $(2
i_1, 2 i_2)$, see Fig.~\ref{fig:typesofwaveletdecompositioncoefficients}.  The
parent-child dependency assumption implies that, if a parent
coefficient in a certain wavelet decomposition level has small (large)
magnitude, then its children coefficients in the next finer wavelet
decomposition level tend to have small (large) magnitude as
well. Denote by $\rho$ and $\kappa$ the numbers of rows and columns of the
image, and by $L$ the number of wavelet decomposition levels (tree
depth).

We set the prior \gls{pmf} $p_{\bm{q}} (\bm{q})$ as follows.  In the
first wavelet decomposition level ($l = 1$), assign
\begin{subequations}
\label{eq:priorforq}
\begin{equation}
  \label{def:1stPhappx}
  p_{q_i}(1) = \Prob\{ q_i = 1 \} =
    \begin{cases}
      1, & i \in \cA\\
      P_{\troot},  & i \in \cT_{\troot}
    \end{cases}
\end{equation}
where
\begin{IEEEeqnarray}{rCl}
  \label{eq:approximationcoefficientindices}
  \cA &=& \upsilon \Bigl( \Bigl\{1, 2, \dotsc, \frac{\rho}{2^L}\Bigr\}
  \times \Bigl\{1, 2, \ldots, \frac{\kappa}{2^L}\Bigr\}\Bigr)
  \\
  \cT_{\troot} &=& \upsilon \Bigl( \Bigl\{1,2, \dotsc,
  \frac{\rho}{2^{L-1}}\Bigr\} \times \Bigl\{1,2, \dotsc,
  \frac{\kappa}{2^{L-1}}\Bigr\} \Bigr) \,\big\backslash\, \mathcal{A}
\end{IEEEeqnarray}
are the sets of indices of the approximation and root node
coefficients and $P_{\troot} \in (0,1)$ is a known constant
denoting the prior probability that a root node signal coefficient
has large magnitude, see
Fig.~\ref{fig:typesofwaveletdecompositioncoefficients}.  In the
levels $l = 2, 3, \ldots, L$, assign
\begin{equation}
  p_{q_i |  q_{\pi(i)}} (1 |  q_{\pi (i)} ) =
    \begin{cases}
      P_{\tH}, & q_{\pi (i)} = 1 \\
      P_{\tL}, & q_{\pi (i)} = 0
    \end{cases}
\end{equation}
\end{subequations}
where $\pi(i)$ denotes the index of the parent of node $i$. Here,
$P_{\tH} \in (0,1)$ and $P_{\tL} \in (0,1)$ are known constants
denoting the probabilities that the signal coefficient $s_i$ is large
if the corresponding parent signal coefficient is large or small,
respectively.

The expected number of large-magnitude signal coefficients is
\begin{subequations}
\begin{IEEEeqnarray}{rCl}
  \Exp\biggl[\sum_{i=1}^p q_i\biggr] = \frac{p}{4^L} \biggl( 1 + 3
  \sum_{l=0}^{L - 1} 4^l P_l \biggr)
   \label{eq:numhighstate}
\end{IEEEeqnarray}
where $P_l$ is the marginal probability that a state variable in the
$l$th tree level is equal to one, computed recursively as follows:
\begin{equation}
  P_l = P_{l - 1} P_{\tH} + (1 - P_{l - 1}) P_{\tL}
\end{equation}
\end{subequations}
initialized by $P_0 = P_{\troot}$.

Our wavelet tree structure consists of $|\cT_{\troot}|$ trees and
spans all signal wavelet coefficients except the approximation
coefficients; hence, the set of indices of the wavelet
coefficients within the trees is
\begin{subequations}
\begin{equation}
  \label{eq:treecoefficientindices}
  \mathcal{T} = \upsilon \bigl( \{1, 2, \ldots, \rho\}
  \times \{1, 2, \ldots, \kappa\} \bigr) \, \backslash \, \mathcal{A} .
\end{equation}
Define also the set of leaf variable node indices within the tree
structure as
\begin{IEEEeqnarray}{rCl}
  \cT_{\sleaf} &=& \upsilon \Bigl( \bigl[\{1, 2, \ldots, \rho\} \times \{1,
  2, \ldots, \kappa \}\bigr] \, \big\backslash \, \Bigl[\Bigl\{1, 2, \dotsc, \frac{\rho}{2}\Bigr\}
  \times \Bigl\{1, 2, \dotsc, \frac{\kappa}{2} \Bigr\}\Bigr] \Bigr)
\end{IEEEeqnarray}
see Fig.~\ref{fig:typesofwaveletdecompositioncoefficients}.
\end{subequations}
We have $5$ tuning parameters $P_{\troot}, P_{\tH}$,
$P_{\tL}$, $\gamma^2$, and $\epsilon^2$, each with a clear meaning. A
fairly crude choice of these parameters is sufficient for achieving
good reconstruction performance, see Section~\ref{sec:example}.

The logarithm of the prior \gls{pmf} $p_{\sbq}(\bq)$ is
\begin{IEEEeqnarray}{rCl}
\ln p_{\bm{q}}(\bq) &=& \text{const} +
    \biggl[ \sum_{i \in \cA} \ln \bone(q_i = 1) \biggr] + \biggl[ \sum_{i
      \in \cT_{\troot}} q_i  \ln P_{\troot} + (1-q_i)
    \ln(1-P_{\troot}) \biggr]
\notag
  \\
    &&  + \biggl[ \sum_{i \in \cT \backslash \cT_{\troot}} q_i
  q_{\pi(i)}  \ln P_{\tH} + (1-q_i)  q_{\pi(i)}  \ln(1-P_{\tH})
\notag
  \\
   && + q_i  (1-q_{\pi(i)})  \ln P_{\tL}  + (1-q_i)  (1-q_{\pi(i)})
   \ln(1-P_{\tL}) \biggr]
\label{eq:prior}
\end{IEEEeqnarray}
where const denotes the terms that are not functions of $\bm{q}$.

\subsection{Bayesian Inference}
\label{sec:BayesianInference}

\noindent
Define the vectors of state variables and signal coefficients
\begin{equation}
  \label{eq:theta}
  \bm{\theta}=\begin{bmatrix} \bm{\theta}_1^T & \bm{\theta}_2^T & \cdots &
    \bm{\theta}_p^T \end{bmatrix}^T, \quad 
\bm{\theta}_i = [ q_i,  s_i ]^T.
\end{equation}
The joint posterior distribution of $\bm{\theta}$ and $\sigma^2$ is
\begin{IEEEeqnarray}{rCl}
  p_{\bm{\theta}, \sigma^2 | \bm{y}}( \bm{\theta}, \sigma^2 | \bm{y} )
  &\propto& p_{\bm{y} | \bm{s}, \sigma^2}( \bm{y} | \bm{s}, \sigma^2 ) \,
  p_{\bm{s} | \bm{q}, \sigma^2}(\bm{s} | \bq, \sigma^2) \,
  p_{\bm{q}}(\bq) \, p_{\sigma^2}(\sigma^2) \nonumber
  \\
  &\propto& (\sigma^2)^{-(p+N+2)/2} \exp\biggl[ - 0.5 \frac{\| \bm{y} -
    H \bm{s} \|_2^2}{\sigma^2} - 0.5 \frac{\bm{s}^T D^{-1}(\bq)
    \bm{s}}{\sigma^2} \biggr] \Bigl(\frac{\epsilon^2}{\gamma^2}\Bigr)^{0.5
    \sum_{i=1}^p q_i} p_{\bm{q}}(\bq) \hspace{0.25in}
 \label{eq:post}
\end{IEEEeqnarray}
which implies
\begin{subequations}
\begin{IEEEeqnarray}{rCl}
  \label{eq:postsigma2}
  p_{\sigma^2 | \bm{\theta},  \bm{y}}( \sigma^2 | \bm{\theta},  \bm{y} ) &=&
  \Invchi2\Bigl( \sigma^2 \,\Big|\, p+N, \frac{ \| \bm{y} - H
    \bm{s} \|_2^2 + \bm{s}^T  D^{-1}(\bq)  \bm{s} }{p+N} \Bigr)
  \\
 \label{eq:pxigivensigma2y}
  p_{\bm{\theta} | \sigma^2, \bm{y} }( \bm{\theta} | \sigma^2, \bm{y} )
  &\propto& \exp\Bigl[ - 0.5 \frac{\| \bm{y} - H \bm{s} \|_2^2 + \bm{s}^T
    D^{-1}(\bq) \bm{s}}{\sigma^2} \Bigr]
  \Big(\frac{\epsilon^2}{\gamma^2}\Big)^{0.5 \sum_{i=1}^p q_i}
  p_{\bm{q}}(\bq).
\end{IEEEeqnarray}
\end{subequations}
We integrate the noise variance parameter from the joint posterior
distribution as follows (see also \cite[(5.5) on p. 126]{Gelmanetal}):
\begin{subequations}
\begin{IEEEeqnarray}{rCl}
   \label{eq:pxigiveny}
  p_{\bm{\theta} | \bm{y} }( \bm{\theta} | \bm{y} ) &=& \frac{p_{\bm{\theta},
       \sigma^2 | \bm{y}}( \bm{\theta},  \sigma^2 | \bm{y} )}{
    p_{\sigma^2 | \bm{\theta},  \bm{y}}( \sigma^2 | \bm{\theta},
    \bm{y} ) } \propto p_{\bm{q}}(\bq)
  \Bigl(\frac{\epsilon^2}{\gamma^2}\Bigr)^{0.5  \sum_{i=1}^p q_i}
  \Big/ \biggl[ \frac{ \| \bm{y} - H  \bm{s} \|_2^2 + \bm{s}^T  D^{-1}(\bq)
     \bm{s} }{p+N} \biggr]^{(p+N)/2}.
\end{IEEEeqnarray}
For a fixed $\bq$, \eqref{eq:pxigiveny} is maximized with respect to
$\bm{s}$ at
\begin{equation}
  \label{eq:shat}
  \bar{\bm{s}}(\bq) = D(\bq)  H^T  [I_N + H  D(\bq)  H^T]^{-1}  \bm{y}
\end{equation}
which is the Bayesian linear-model \gls{MMSE}
estimator of $\bm{s}$ for a given $\bq$ \cite[Theorem 11.1]{Kay}.
As $\epsilon^2$ decreases to zero, $\bar{\bm{s}}(\bq)$ becomes
more sparse (becoming exactly sparse for $\epsilon^2=0$); as
$\epsilon^2$ increases, $\bar{\bm{s}}(\bq)$ becomes less sparse.

Substituting \eqref{eq:shat} into \eqref{eq:pxigiveny} yields
the \emph{concentrated (profile) marginal posterior distribution}:
\begin{equation}
  \label{eq:concentratedposteriorwrts}
  \max_{\bm{s}} p_{\bm{\theta} | \bm{y}} (\bm{\theta} | \bm{y}) \propto
  p_{\bm{q}}(\bq)  \Bigl(\frac{\epsilon^2}{\gamma^2}\Bigr)^{0.5
    \sum_{i=1}^p q_i} \Big/
\biggl\{\frac{\bm{y}^T  [I_N + H
    D(\bq)  H^T]^{-1}  \bm{y}  }{p+N} \biggr\}^{(p+N)/2}
\end{equation}
\end{subequations}
Which is a function of the state variables $\bq$ only.

We wish to maximize (\ref{eq:pxigiveny}) with respect to
$\bm{\theta}$, but cannot perform this task directly.  Consequently,
we adopt an indirect approach: We first develop an \gls{EM} algorithm
for maximizing $p_{\bm{\theta} | \sigma^2, \bm{y}}( \bm{\theta} |
\sigma^2, \bm{y} )$ in (\ref{eq:pxigivensigma2y}) for a given
$\sigma^2$ (Section~\ref{sec:BayesianEM}) and then apply a grid
search scheme for selecting the best noise variance parameter
$\sigma^2$ so that the estimated signal and state variables have the
largest marginal posterior distribution \eqref{eq:pxigiveny}
(Section~\ref{sec:sigma2choice}).

\section{An EM Algorithm for Maximizing $p_{\bm{\theta} | \sigma^2,
    \bm{y}} (\bm{\theta} | \sigma^2, \bm{y})$}
\label{sec:BayesianEM}

\noindent
Motivated by \cite[Sec.~V.A]{FigueiredoNowak},
we introduce the following hierarchical two-stage model:
\begin{subequations}
\label{eq:hierarchicalmodel}
\begin{IEEEeqnarray}{rCl}
  \label{eq:pygivenzsigma2}
  p_{ \bm{y} | \bm{z}, \sigma^2 }( \bm{y} | \bz, \sigma^2 ) &=& 
\mathcal{N}\bigl( \bm{y} | H  \bz, \sigma^2  (I_N - H  H^T)
  \bigr)
  \\
  \label{eq:pz}
  p_{ \bm{z} | \bm{s}, \sigma^2 }( \bz | \bm{s}, \sigma^2 ) &=&
  \mathcal{N}( \bz | \bm{s}, \sigma^2 I_p )
\end{IEEEeqnarray}
where $\bz$ is a $p \times 1$ vector of \emph{missing data}.
\end{subequations}
Observe that the spectral norm condition \eqref{eq:spectralnormofH}
guarantees that the covariance matrix $\sigma^2 (I_N - H H^T)$ in
\eqref{eq:pygivenzsigma2} is positive semidefinite.

Our \gls{EM} algorithm for maximizing $p_{\bm{\theta} | \sigma^2,
  \bm{y}}( \bm{\theta} | \sigma^2, \bm{y} )$ in
(\ref{eq:pxigivensigma2y}) consists of iterating between the following
\gls{E} and \gls{M} steps (see
Appendix~\ref{app:EMalgorithm}):\footnote{\label{adaptivestepsize}If
  the spectral norm of the sensing matrix $H$ cannot be easily
  determined or estimated [and, therefore, \eqref{eq:spectralnormofH}
  cannot be ensured], we can introduce an adaptive positive step size that
  multiplies the second summand in the \gls{E} step
  \eqref{eq:EstepBayes}; we also need to divide the first summand in
  \eqref{eq:MstepBayes1} by this quantity. Then, the step size adaptation can be performed
  along the lines of \cite{dgq11}, with goal to ensure monotonicity of
  the \gls{EM} iteration. Such a step size adaptation (which, in
  effect, estimates the spectral norm of $H$) is typically completed
  within the first few \gls{EM} iterations.}
\begin{IEEEeqnarray}{rCl}
\kern-8em \text{\textsc{e} step:} \kern1.5em
  \bz^{(j)} &\triangleq& \Exp_{\bm{z} | \sigma^2, \bm{y}, \bm{s}} [\bz |
  \sigma^2, \bm{y}, \bm{s}^{(j)}] =  [ z_1^{(j)},   z_2^{(j)},  \dotsc,
    z_p^{(j)} ]^T = \bm{s}^{(j)} +
  H^T   (\bm{y} -  H  \bm{s}^{(j)})
  \label{eq:EstepBayes}
\end{IEEEeqnarray}
\begin{subequations}
\label{eq:MstepBayes}
\begin{IEEEeqnarray}{rCl}
  \kern-6em \text{\textsc{m} step:} \kern0.5em \bm{\theta}^{(j+1)} &=& \arg
  \max_{\bm{\theta}} \biggl\{ - 0.5 \frac{\| \bz^{(j)} - \bm{s} \|_2^2 + \bm{s}^T
    D^{-1}(\bq) \bm{s}}{\sigma^2} + \ln[p_{\bm{q}}(\bq)] + 0.5
  \ln\Big(\frac{\epsilon^2}{\gamma^2}\Big) \sum_{i=1}^p q_i \biggr\}
  \label{eq:MstepBayes1}
\\
&=& \arg \max_{\bm{\theta}} \ln p_{\bm{\theta} | \sigma^2, \bm{z}}
(\bm{\theta} | \sigma^2, \bz^{(j)})
\label{eq:MstepBayes2}
\end{IEEEeqnarray}
\end{subequations}
where $j$ denotes the iteration index.  See, e.g.,
\cite[Sec. 11.4]{Murphy}, \cite{DLR}, and \cite{McLachlanKrishnan} for a
general exposition on the \gls{EM} algorithm and its properties and
\cite[Chapter 12.3]{Gelmanetal} for its Bayesian version.  To
simplify the notation, we omit the dependence of the iterates
on $\sigma^2$ in this section. Denote by
$\bm{\theta}^{(+\infty)}, \bm{s}^{(+\infty)}$, and $\bq^{(+\infty)}$
the estimates of $\bm{\theta},\bm{s}$, and $\bq$ obtained upon
convergence of the above \gls{EM} iteration.

For any two consecutive iterations $j$ and $j+1$, this \gls{EM}
algorithm ensures that the objective posterior function \emph{does
  not} decrease, i.e.,
\begin{IEEEeqnarray}{rCl}
  p_{\bm{\theta} | \sigma^2, \bm{y}}(\bm{\theta}^{(j+1)} | \sigma^2,
  \bm{y}) \geq p_{\bm{\theta} | \sigma^2, \bm{y}}(\bm{\theta}^{(j)} |
  \sigma^2, \bm{y})
\label{eq:monotonicity}
\end{IEEEeqnarray}
see Appendix~\ref{app:EMalgorithm}.
Monotonic convergence is also a key general property of the \gls{EM}-type
algorithms \cite{McLachlanKrishnan}.

\begin{theorem} 
  \label{theorem}
  The signal and binary state variable estimates $\bm{s}^{(+\infty)}$
  and $\bq^{(+\infty)}$ obtained upon convergence of the \gls{EM}
  iteration \eqref{eq:EstepBayes}--\eqref{eq:MstepBayes} satisfy
\begin{equation}
  \label{eq:sinftyandshat}
  \bm{s}^{(+\infty)} = \bar{\bm{s}}(\bq^{(+\infty)}).
\end{equation}
Hence, this iteration provides an estimate $\bq^{(+\infty)}$ of the
  vector of state variables $\bq$ \emph{as well as} finds the solution
  \eqref{eq:shat} of the underlying linear system to obtain the
  corresponding signal estimate.
\end{theorem}
\begin{IEEEproof}
  See Appendix~\ref{app:EMalgorithm}.
\end{IEEEproof}
Consequently, as $\epsilon^2$ decreases to zero, $\bm{s}^{(+\infty)}$
becomes more sparse; as $\epsilon^2$ increases, $\bm{s}^{(+\infty)}$
becomes less sparse.

Note that the \gls{M} step in (\ref{eq:MstepBayes2}) is equivalent to
maximizing $p_{\bm{\theta} | \sigma^2, \bm{z}} (\bm{\theta} | \sigma^2,
\bz)$ for the missing data vector $\bz=\bz^{(j)}$. In the following
section, we describe efficient maximization of $p_{\bm{\theta} | \sigma^2,
  \bm{z}} (\bm{\theta} | \sigma^2, \bz)$.

\subsection{M Step: Maximizing $p_{\bm{\theta} | \sigma^2, \bm{z}} (\bm{\theta} | \sigma^2, \bz)$}
\label{sec:BP}

\noindent
Before we proceed, define
\begin{equation}
  \label{eq:sihat}
  \widehat{s}_i(0)  = \frac{\epsilon^2}{1 + \epsilon^2}  z_i, \quad
  \widehat{s}_i(1) = \frac{\gamma^2}{1 + \gamma^2}  z_i
\end{equation}
where we omit the dependence of $\widehat{s}_i(0)$ and
$\widehat{s}_i(1)$ on $z_i$ to simplify the notation.

\begin{figure}
  \centering
  \includegraphics[width=\linewidth]{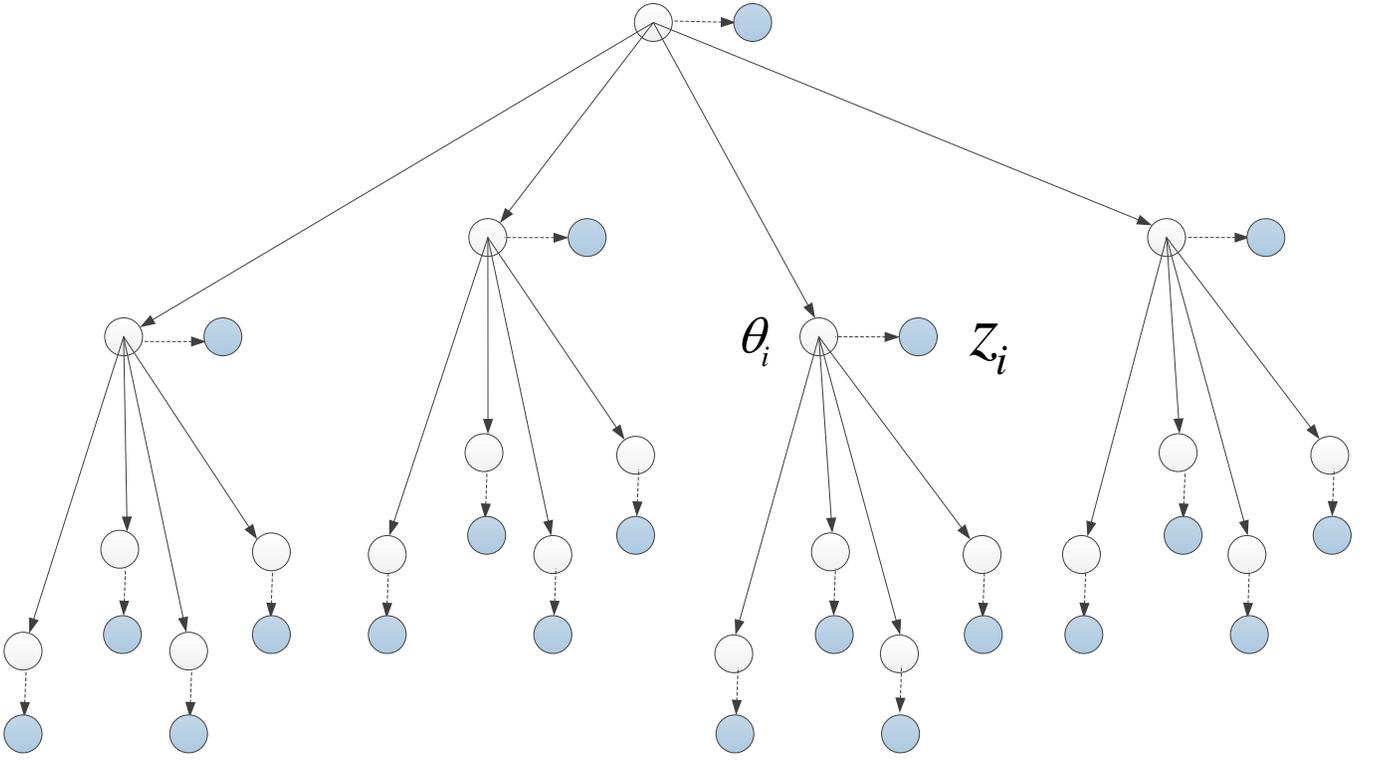}
  \caption{A \gls{HMT}, part of the probabilistic model
    (\ref{eq:qsgivenztree}).}
  \label{fig:hmt}
  \vspace{-0.2in}
\end{figure}

Observe that
\begin{IEEEeqnarray}{rCl}
  p_{\bm{\theta} | \sigma^2, \bm{z}} (\bm{\theta} | \sigma^2, \bz) &\propto&
  p_{\bm{\theta}_{\cA} | \sigma^2, \bm{z}} (\bm{\theta}_{\scA} | \sigma^2, \bz) 
  p_{\bm{\theta}_{\cT} | \sigma^2, \bm{z}} (\bm{\theta}_{\cT} | \sigma^2, \bz)
  \label{eq:qsgivenz}
\end{IEEEeqnarray}
where $\bm{\theta}_{\scA}$ and $\bm{\theta}_{\cT}$ consist of $\bm{\theta}_i, i \in \cA$
and $\bm{\theta}_i, i \in \cT$, respectively,
 and
\begin{subequations}
  \begin{IEEEeqnarray}{rCL}
    \label{eq:qsgivenzappx}
    p_{\bm{\theta}_{\cA} | \sigma^2, \bm{z}} (\bm{\theta}_{\scA} |
    \sigma^2, \bz) & \propto & \prod_{i \in \cA} \mathcal{N}(
    z_i | s_i, \sigma^2 )  \mathcal{N}( s_i | 0, \gamma^2 
    \sigma^2 )  \bone(q_i = 1) 
    \\
    \label{eq:qsgivenztree}
    p_{\bm{\theta}_{\cT} | \sigma^2, \bm{z}} (\bm{\theta}_{\cT} |
    \sigma^2, \bz) & \propto & \biggl\{ \prod_{i \in \cT} \mathcal{N}(
    z_i | s_i, \sigma^2 )  [\mathcal{N}( s_i | 0, \gamma^2 
    \sigma^2 )]^{q_i}  [\mathcal{N}( s_i | 0, \epsilon^2  \sigma^2
    )]^{1-q_i} \biggr\} p_{\bm{q}_{\cT}} (\bq_{\cT}).
  \end{IEEEeqnarray}
\end{subequations}
Here, (\ref{eq:qsgivenzappx}) follows from (\ref{def:1stPhappx}) and
(\ref{eq:qsgivenztree}) corresponds to the \gls{HMT}
probabilistic model that contains no loops. Fig.~\ref{fig:hmt} depicts
an \gls{HMT} that is a part of the probabilistic model
(\ref{eq:qsgivenztree}). Maximizing $p_{\bm{\theta}_{\cA} | \sigma^2,
  \bm{z}} (\bm{\theta}_{\scA} | \sigma^2, \bz^{(j)})$ in
\eqref{eq:qsgivenzappx} with respect to $\bm{\theta}_i, i \in \cA$ yields
\begin{equation}
  \label{eq:Mstepforapproximationcoefficients}
\widehat{\bm{\theta}}_i =  [ 1,  \widehat{s}_i(1) ]^T, \qquad i \in \mathcal{A}
\end{equation}
where we have used the identity \eqref{eq:appmodeofproductofGaussians}
in Appendix~\ref{app:max-productalgorithmderivation}.

We now apply the max-product belief propagation
algorithm \cite{KollerFriedman,WeissFreeman,Pearl} to each tree in our
wavelet tree structure, with the goal to find the mode of
$p_{\bm{\theta}_{\cT} | \sigma^2, \bm{z}} (\bm{\theta}_{\cT} | \sigma^2,
\bz)$.
We represent the \gls{HMT} probabilistic model for $p_{\bm{\theta}_{\cT} |
  \sigma^2, \bm{z}} (\bm{\theta}_{\cT} | \sigma^2, \bz)$ via
\emph{potential functions} as [see \eqref{eq:qsgivenztree}]
\begin{equation}
  \label{eq:xigivenztreepotential}
  p_{\bm{\theta}_{\cT} | \sigma^2, \bm{z}} (\bm{\theta}_{\cT} | \sigma^2, \bz) \propto \biggl[ \prod_{i
    \in \cT \backslash \cT_{\troot}} \psi_i(\bm{\theta}_i)  \psi_{i,\pi(i)}(
  q_i, q_{\pi(i)}) \biggr]  \biggl[ \prod_{i \in \cT_{\troot}}
  \psi_i(\bm{\theta}_i) \biggr]
\end{equation}
where
\begin{subequations}
  \label{eq:potentialfunctions}
  \begin{IEEEeqnarray}{rCl}
    \psi_i(\bm{\theta}_i) &=&
    \mathcal{N}(z_i | s_i, \sigma^2 ) 
        [\mathcal{N}(s_i | 0, \gamma^2  \sigma^2 )]^{q_i} 
         [\mathcal{N}(s_i | 0, \epsilon^2  \sigma^2 )]^{1-q_i}
    \label{eq:nodepotential1}
 \end{IEEEeqnarray}
for $i \in \cT \backslash \cT_{\troot}$,
  \begin{IEEEeqnarray}{rCl}
  \psi_i(\bm{\theta}_i) &=& \mathcal{N}( z_i | s_i, \sigma^2 ) 
        [P_{\troot}  \mathcal{N}( s_i | 0, \gamma^2  \sigma^2
        )]^{q_i} 
         [(1-P_{\troot})  \mathcal{N}( s_i | 0,
        \epsilon^2  \sigma^2 )]^{1-q_i}
        \label{eq:nodepotential2}
  \end{IEEEeqnarray}
 for $i \in \cT_{\troot}$,
  and
  \begin{IEEEeqnarray}{rCl}
    \psi_{i,\pi(i)}(q_i, q_{\pi(i)}) &=& [{P_{\tH}}^{q_i} 
    (1-P_{\tH})^{1-q_i}]^{q_{\pi(i)}} 
     [{P_{\tL}}^{q_i}  (1 -
    P_{\tL})^{1-q_i}]^{1-q_{\pi(i)}}
    \label{eq:edgepotential}
  \end{IEEEeqnarray}
for $i \in \cT \backslash \cT_{\troot}$.
\end{subequations}

Our algorithm for maximizing (\ref{eq:xigivenztreepotential}) consists
of computing and passing upward and downward messages and calculating
and maximizing beliefs.

\subsubsection{Computing and Passing Upward Messages}
\label{subsubsec:upmsg}

\begin{figure}
\centering
\subfigure[]{\includegraphics[width=0.35\linewidth]{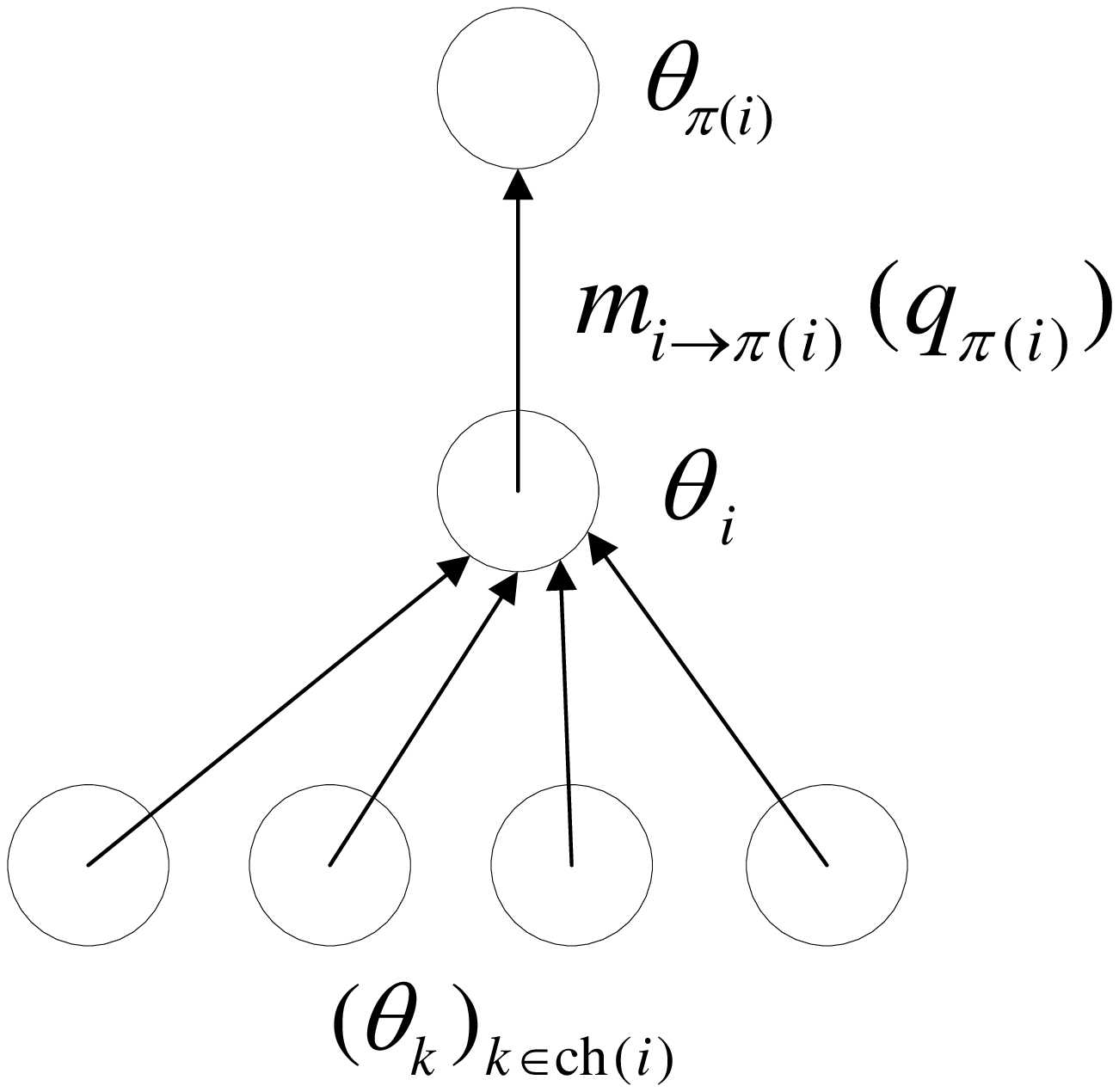}
\label{fig:msgup} } \hfil
\subfigure[]{\includegraphics[width=0.35\linewidth]{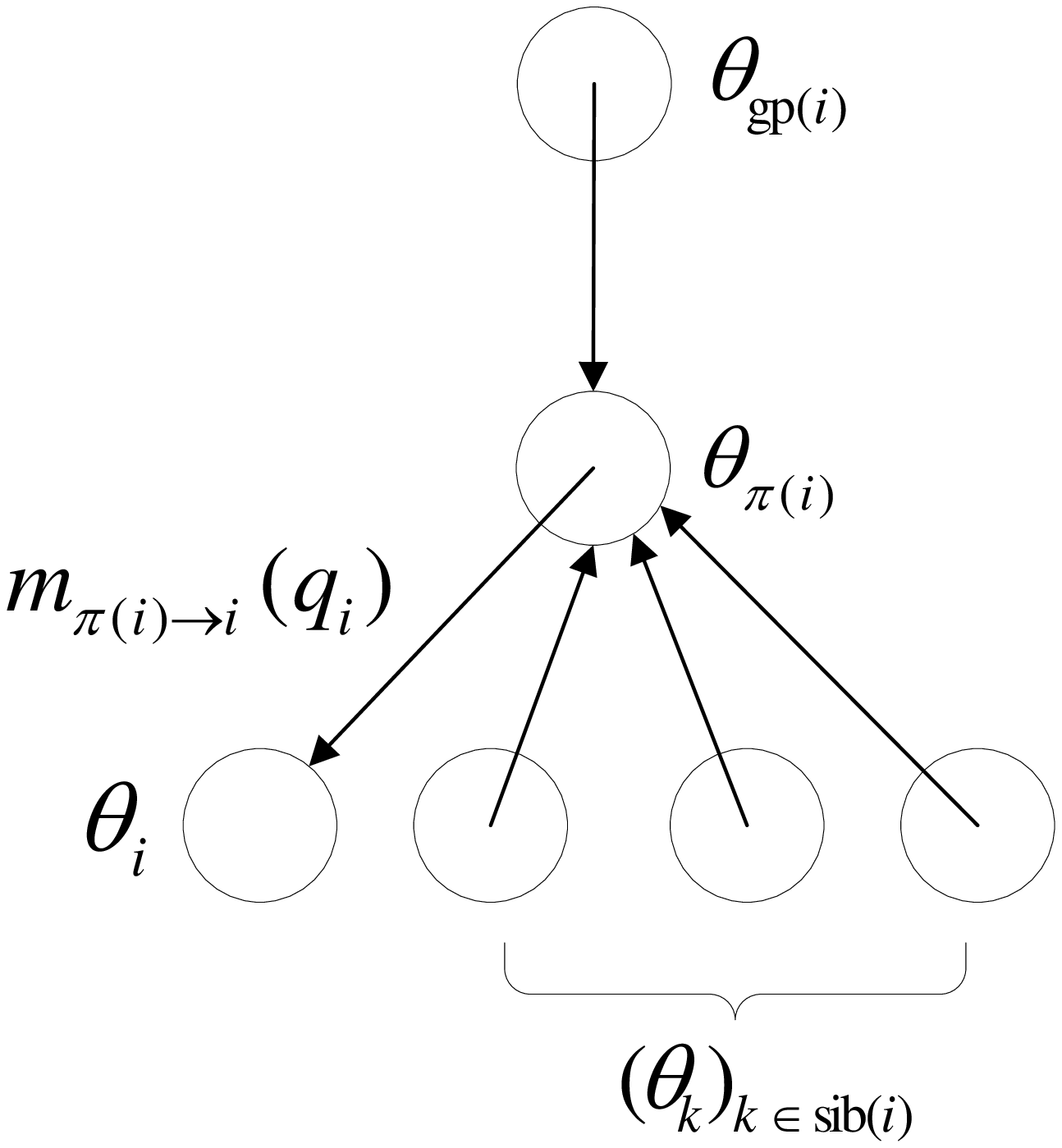}
\label{fig:msgdown} }  \caption{Computing and passing (a)
upward and (b) downward messages.} \label{fig:msg}
\end{figure}

\noindent
We propagate the upward messages from the lowest decomposition level
(i.e., the leaves) towards the root of the tree.  Fig.~\ref{fig:msgup}
depicts the computation of the upward message from variable node
$\bm{\theta}_i$ to its parent node $\bm{\theta}_{\pi (i)}$ wherein we also define a
\emph{child} of $\bm{\theta}_i$ as a variable node $\bm{\theta}_k$ with index $k \in
\ch(i)$, where $\ch(i)$ is the index set of the children of $i$: for
$i = \upsilon (i_1,i_2)$, $\ch(i) = \{ \upsilon \big( (2  i_1 - 1, 2
 i_2 - 1), (2  i_1 - 1, 2  i_2), (2  i_1, 2  i_2 - 1), (2 
i_1, 2  i_2) \big) \}$. Here, we use a circle and an edge with an
arrow to denote a variable node and a message, respectively.  The
upward messages have the following general form \cite{WeissFreeman}:
\begin{equation}
  \label{eq:upwardmsg}
  m_{i \rightarrow \pi(i)}( q_{\pi (i)} ) = \alpha \max_{\bm{\theta}_i}
  \biggl\{ \psi_i(\bm{\theta}_i)  \psi_{i,\pi(i)}( q_i, q_{\pi(i)}) 
  \prod_{k \in \ch (i)} m_{k \rightarrow i}( q_i ) \biggr\}
\end{equation}
where $\alpha > 0$ denotes a normalizing constant used for computational
stability \cite{WeissFreeman}.  For nodes with no children
(corresponding to level $L$, i.e., $i \in \cT_{\sleaf}$), we set
the multiplicative term $\prod_{k \in \ch (i)} m_{k \rightarrow
i}(\bm{\theta}_i)$ in \eqref{eq:upwardmsg} to one.

In Appendix~\ref{app:Upwardmessagederivation}, we show that
the only two candidates for $\bm{\theta}_i$ in the maximization of
(\ref{eq:upwardmsg}) are $[0, \widehat{s}_i(0)]^T$
and $[1, \widehat{s}_i(1)]^T$, see also
\eqref{eq:sihat}.

Substituting these candidates into (\ref{eq:upwardmsg}) and
normalizing the messages yields (see
Appendix~\ref{app:Upwardmessagederivation})
\begin{subequations}
  \label{eq:upwardmsgform}
  \begin{equation}
    \label{eq:msgupwardform}
    m_{i \rightarrow \pi(i)}( q_{\pi (i)} ) = [\mu^{\tu}_i(0)]^{1-q_{\pi (i)}} 
    [\mu^{\tu}_i(1)]^{q_{\pi
        (i)}}
  \end{equation}
  where $[\mu_i^{\tu}(0), \mu_i^{\tu}(1)]^T = \bmu_i^{\tu}$,
  \begin{IEEEeqnarray}{rCL}
    \label{eq:muiu}
    \bmu_i^{\tu} &=& \frac{ [\max \{ \bnu_{0,i}^{\tu} \odot
      \etabold_i^{\tu} \},  \max \{ \bnu_{1,i}^{\tu} \odot
      \etabold_i^{\tu} \} ]^T}{ \max \{ \bnu_{0,i}^{\tu} \odot
      \etabold_i^{\tu} \} + \max \{ \bnu_{1,i}^{\tu} \odot
      \etabold_i^{\tu} \} }
\nonumber
\\
&=& \frac{ \begin{bmatrix} \exp \bigl( \ln ( \max \{
        \bnu_{0,i}^{\tu} \odot \etabold_i^{\tu} \} ) - \ln ( \max \{
        \bnu_{1,i}^{\tu} \odot \etabold_i^{\tu} \} ) \bigr), &
        1 \end{bmatrix}^T}{ 1 + \exp \bigl( \ln ( \max \{ \bnu_{0,i}^{\tu}
      \odot \etabold_i^{\tu} \} ) - \ln ( \max \{ \bnu_{1,i}^{\tu} \odot
      \etabold_i^{\tu} \} ) \bigr) }
    \\
    \label{eq:nu0i}
    \bnu_{0,i}^{\tu} &=&
    [1-P_{\tL}, P_{\tL}]^T \odot
    \bphi(z_i)
    \\
    \label{eq:nu1i}
    \bnu_{1,i}^{\tu} &=& [1-P_{\tH}, P_{\tH}]^T \odot
    \bphi(z_i)
    \\
    \label{eq:zetaiu}
    \etabold_i^{\tu} &=& 
    \begin{cases}
      \bigodot_{k \in \ch (i)} \bmu^{\tu}_k , &   i \in \cT \backslash
      \cT_{\sleaf}
      \\
      [ 1, 1 ]^T, & i \in \cT_{\sleaf}
    \end{cases}
    \\
    \label{eq:phi}
    \bphi(z) &=& \begin{bmatrix} \exp( - 0.5  \frac{z^2}{\sigma^2+\sigma^2 \epsilon^2})/\epsilon, & \exp( - 0.5 
      \frac{z^2}{\sigma^2+\sigma^2 \gamma^2})/\gamma
    \end{bmatrix}^T
  \end{IEEEeqnarray}
\end{subequations}
and $\epsilon = \sqrt{\epsilon^2} > 0$ and $\gamma = \sqrt{\gamma^2} >
0$.  A numerically stable implementation of \eqref{eq:muiu} that we
employ is illustrated in the second expression in
\eqref{eq:muiu}. Similarly, the elementwise products in
\eqref{eq:nu0i}--\eqref{eq:zetaiu} are implemented as exponentiated
sums of logarithms of the product terms.

\subsubsection{Computing and Passing Downward Messages}
\label{subsubsec:downmsg}

\noindent
Upon obtaining all the upward messages, we now compute the downward
messages and propagate them from the root towards the lowest level
(i.e., the leaves). Fig.~\ref{fig:msgdown} depicts the computation of
the downward message from the parent $\bm{\theta}_{\pi (i)}$ to
the variable node $\bm{\theta}_i$, which involves upward messages to
$\bm{\theta}_{\pi (i)}$ from its other children, i.e.\ the
\emph{siblings} of $\bm{\theta}_i$, marked as $\bm{\theta}_k,  k \in {\rm
  sib}(i)$. This downward message also requires the message sent to
$\bm{\theta}_{\pi (i)}$ from its parent node, which is the
\emph{grandparent} of $\bm{\theta}_i$, denoted by $\bm{\theta}_{\gp
(i)}$. The downward messages have the following general form
\cite{WeissFreeman}:
\begin{IEEEeqnarray}{rCl}
\label{eq:downwardmsg}
  m_{\pi(i) \rightarrow i}( q_i ) = \alpha \max_{\bm{\theta}_{\pi(i)}}
  \Big\{ \psi_{\pi(i)}(\bm{\theta}_{\pi(i)})  \psi_{i,\pi(i)}( q_i,
  q_{\pi(i)})  m_{\gp(i) \rightarrow \pi(i)}(q_{\pi(i)})  \prod_{k
    \in \sib(i)} m_{k \rightarrow \pi(i)}( q_{\pi(i)} ) \Big\}
\hspace{0.25in}
\end{IEEEeqnarray}
where $\alpha > 0$ denotes a normalizing constant used for
computational stability.  For the variable nodes $i$ in the second
decomposition level that have no grandparents (i.e., $\pi (i) \in
\cT_{\troot}$), we set the multiplicative term $m_{\gp(i) \rightarrow
  \pi(i)}(q_{\pi(i)})$ in (\ref{eq:downwardmsg}) to one.

In Appendix~\ref{app:Downwardmessagederivation}, we show that the only
two candidates for $\bm{\theta}_{\pi(i)}$ in the maximization of
\eqref{eq:downwardmsg} are $[0, \widehat{s}_{\pi(i)}(0)]^T$ and $[1,
\widehat{s}_{\pi(i)}(1)]^T$, see also \eqref{eq:sihat}.  Substituting
these candidates into (\ref{eq:downwardmsg}) and normalizing the
messages yields (see Appendix~\ref{app:Downwardmessagederivation})
\begin{subequations}
  \begin{equation}
    \label{def:downmsgsummary}
    m_{\pi(i) \rightarrow i}(q_i) = [\mu_i^{\td}(0)]^{1-q_i}  [\mu^{\td}_i(1)]^{q_i}
\end{equation}
for $\pi (i) \in \cT \backslash \cT_{\sleaf}$, where $
[\mu_i^{\td}(0), \mu_i^{\td}(1)]^T = \bmu_i^{\td}$ and
\begin{IEEEeqnarray}{rCL}
\label{eq:muid}
  \bmu_i^{\td} &=& \frac{ [\max \{ \bnu_{0,i}^{\td} \odot
    \etabold_i^{\td} \},   \max \{ \bnu_{1,i}^{\td} \odot
    \etabold_i^{\td} \}]^T } { \max \{ \bnu_{0,i}^{\td} \odot
    \etabold_i^{\td} \} + \max \{ \bnu_{1,i}^{\td} \odot \etabold_i^{\td}
    \} }
\nonumber
\\
    &=& \frac{ \begin{bmatrix} \exp\bigl(  \ln ( \max \{ \bnu_{0,i}^{\td} \odot
\etabold_i^{\td} \} ) - \ln ( \max \{ \bnu_{1,i}^{\td} \odot
\etabold_i^{\td}
    \} ) \bigr), & 1  \end{bmatrix}^T}{ 1 + \exp\bigl(  \ln ( \max 
    \{ \bnu_{0,i}^{\td} \odot \etabold_i^{\td} \} ) - \ln ( \max 
    \{ \bnu_{1,i}^{\td} \odot \etabold_i^{\td} \} ) \bigr) }
  \\
  \bnu_{0,i}^{\td} &=& [1-P_{\tL}, 1-P_{\tH}]^T \odot
  \bphi(z_{\pi(i)}) \odot \biggl[ \bigodot_{k \in \sib (i)}
  \bmu^{\tu}_k \biggr]
  \\
  \bnu_{1,i}^{\td} &=& [P_{\tL}, P_{\tH}]^T \odot \bphi(z_{\pi(i)})
  \odot \biggl[ \bigodot_{k \in \sib (i)} \bmu^{\tu}_k \biggr]
  \\
  \etabold_i^{\td} &=&  \begin{cases}
   [  1 - P_{\troot},  P_{\troot} ]^T, & \pi (i) \in \cT_{\troot}
   \\
   \bmu_{\pi(i)}^{\td}, & \pi(i) \in ( \cT \backslash \cT_{\troot} ) \backslash \cT_{\sleaf}
\end{cases}.
\end{IEEEeqnarray}
\label{eq:downwardmsgsummary}
\end{subequations}
A numerically stable implementation of \eqref{eq:muid} that we
employ is illustrated in the second expression in
\eqref{eq:muid}.\looseness=-1

The above upward and downward messages have discrete representations,
which is practically important and is a consequence of the fact that
we use a Gaussian prior on the signal coefficients, see
(\ref{eq:sgivenqprior0}). Indeed, in contrast with the existing
message passing algorithms for compressive sampling
\cite{BaronSarvothamBaraniuk,DonohoMalekiMontanari,Schniter,SomSchniter},
our max-product scheme employs \emph{exact} messages.

\subsubsection{Maximizing Beliefs}
\label{subsubsec:belief}

\noindent
Upon computing and passing all the upward and downward messages, we
maximize the beliefs, which have the following general form
\cite{WeissFreeman}:
\begin{equation}
  \label{eq:belief}
  b(\bm{\theta}_i) = \alpha  \psi_i(\bm{\theta}_i)  m_{\pi(i) \rightarrow i}( q_i ) 
  \prod_{k \in \ch (i)} m_{k \rightarrow i}( q_i )
\end{equation}
for each $i \in \cT$, where $\alpha > 0$ is a normalizing constant.
[In \eqref{eq:belief}, we set $m_{\pi(i) \rightarrow i}( q_i ) = 1$ if
$i \in \cT_{\troot}$ and $\prod_{k \in \ch (i)} m_{k \rightarrow i}(
q_i ) = 1$ if $i \in \cT_{\sleaf}$.]  We then use these beliefs to
obtain the mode
\begin{equation}
  \label{eq:modexiTgivenz}
  \widehat{\bm{\theta}}_{\cT} = \arg \max_{\bm{\theta}_{\cT}} p_{\bm{\theta}_{\cT}
    | \sigma^2,  \bm{z}}(\bm{\theta}_{\cT} | \sigma^2, \bz)
\end{equation}
where the elements of $\widehat{\bm{\theta}}_{\cT}$ are [see
\eqref{eq:sihat}]
\begin{subequations}
  \label{eq:mapofxiT}
  \begin{equation}
    \label{eq:maximizedbeliefs}
    \widehat{\bm{\theta}}_i =
[\widehat{q}_i,  \widehat{s}_i(\widehat{q}_i)]^T
 = \arg
    \max_{\bm{\theta}_i} b(\bm{\theta}_i) =
    \begin{cases} [1,  \widehat{s}_i(1) ]^T, &
        \beta_i (1) \geq \beta_i (0) \\
        [ 0, \widehat{s}_i(0) ]^T, & \text{otherwise}
      \end{cases}, \qquad
    i \in \cT
  \end{equation}
  and
  \begin{equation}
    \label{eq:betas}
    \bbeta_i =  [ \beta_i (0),  \beta_i (1) ]^T =
    \begin{cases} \alpha_1 [ 1 - P_{\troot},  P_{\troot} ]^T \odot \bphi(z_i)  \odot \etabold_i^{\tu}, & i \in \cT_{\troot}
        \\
        \alpha_1  \bphi(z_i) \odot \bmu_i^{\td} \odot \etabold_i^{\tu}, & i \in
        \cT  \backslash  \cT_{\troot}
      \end{cases}.
  \end{equation}
\end{subequations}
Here, $\alpha_1 > 0$ is a normalizing constant.
The detailed derivation for the forms of $\widehat{\bm{\theta}}_i$ and
$\bbeta_i$ in \eqref{eq:mapofxiT} is provided in
Appendix~\ref{app:Beliefsderivation}.

\section{Selecting $\sigma^2$ via Grid Search}
\label{sec:sigma2choice}

\noindent We can integrate $\sigma^2$ out, yielding the marginal
posterior of $\bm{\theta}$ in \eqref{eq:pxigiveny}, and derive an
`outer' \gls{EM} iteration for maximizing $p_{\bm{\theta} | \bm{y} }(
\bm{\theta} | \bm{y} )$:
\begin{enumerate}[(i)]
\item 
\label{fixsigma2esttheta}
fix $\sigma^2$ and apply the \gls{EM} iteration proposed in
  Section~\ref{sec:BayesianEM} to obtain an estimate
  $\bm{\theta}^{(+\infty)}(\sigma^2)$ of $\bm{\theta}$;
\item 
\label{fixthetaestsigma2}
fix $\bm{\theta}$ to the value obtained in (i) and estimate
  $\sigma^2$ as
    \begin{equation}
  \widehat{\sigma}^2(\bm{\theta}) = \frac{\|\bm{y} - H  \bm{s}\|_2^2 + \bm{s}^T  D^{-1}(\bq)  \bm{s}}{p+N}.
    \end{equation}
\end{enumerate}
Even though it guarantees monotonic increase of the marginal posterior
$p_{\bm{\theta} | \bm{y} }( \bm{\theta} | \bm{y} )$, the `outer' \gls{EM}
iteration (\ref{fixsigma2esttheta})--(\ref{fixthetaestsigma2})
does not work well in practice because it gets
stuck in an undesirable local maximum of $p_{\bm{\theta} | \bm{y} }(
\bm{\theta} | \bm{y} )$. To find a better (generally local) maximum of
$p_{\bm{\theta} | \bm{y} }( \bm{\theta} | \bm{y} )$, we apply a grid
search over $\sigma^2$ as follows.

\begin{figure}
\centering
\includegraphics[width=0.45\linewidth]{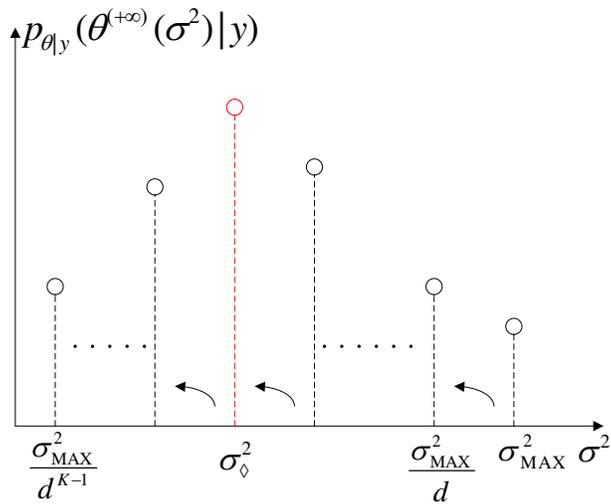}
\caption{Grid search for selecting $\sigma^2$.}
\label{fig:GridSearch}
\end{figure}

We apply the \gls{EM} algorithm in Section~\ref{sec:BayesianEM} using
a range of values of the regularization parameter $\sigma^2$. We
traverse the grid of $K$ values of $\sigma^2$ sequentially and use the
signal estimate from the previous grid point to initialize the signal
estimation at the current grid point (as depicted in
Fig.~\ref{fig:GridSearch}): in particular, we move from a larger
$\sigma^2$ (say $\sigma^2_{\text{old}}$) to the next smaller
$\sigma^2_{\text{new}} ( < \sigma^2_{\text{old}})$ and use
$\bm{s}^{(+\infty)} (\sigma^2_{\text{old}})$ (obtained upon
convergence of the \gls{EM} iteration in Section~\ref{sec:BayesianEM}
for $\sigma^2=\sigma^2_{\text{old}}$) to initialize the \gls{EM}
iteration at $\sigma^2_{\text{new}}$.  The largest $\sigma^2$ on the
grid and the initial signal estimate at this grid point are selected
as
\begin{subequations}
\label{eq:grid4sigma2}
\begin{equation}
  \label{eq:sigma2MAXsinit}
  \sigma_{\tMAX}^2 = \frac{\| \bm{y} \|_2^2}{p+N}, \quad \bm{\theta}^{(0)}( \sigma_{\tMAX}^2 ) = \bzero_{2p \times 1}.
\end{equation}
The consecutive grid points $\sigma^2_{\text{new}}$ and $\sigma^2_{\text{old}}$ satisfy
\begin{equation}
  \sigma^2_{\text{new}} = \frac{\sigma^2_{\text{old}}}{d}
  \label{eq:sigma2relationship}
\end{equation}
where $d > 1$ is a constant determining the search resolution.
\end{subequations}
Finally, we select the $\sigma^2$ from the above grid of candidates
that yields the largest marginal posterior distribution
\eqref{eq:pxigiveny}:
\begin{equation}
  \sigma_{\diamondsuit}^2 = \arg \max_{\sigma^2 \in \{ \sigma_{\tMAX}^2,
    \sigma_{\tMAX}^2/d, \ldots, \sigma_{\tMAX}^2/d^{K-1} \} }
  p_{\bm{\theta} | \bm{y} }( \bm{\theta}^{(+\infty)}(\sigma^2) | \bm{y} )
  \label{eq:sigma2star}
\end{equation}
and the final estimates of $\bm{\theta}$ and $\bm{s}$ as
$\bm{\theta}^{(+\infty)}(\sigma_{\diamondsuit}^2)$ and
$\bm{s}^{(+\infty)}(\sigma_{\diamondsuit}^2)$, respectively, see
Fig.~\ref{fig:GridSearch}.

\section{Numerical Examples}
\label{sec:example}

\noindent
We compare the reconstruction performances of the
following methods:
\begin{itemize}

\item our proposed \gls{MP-EM} algorithm in
  Section~\ref{sec:BayesianEM} with the variance parameter $\sigma^2$
  selected via grid search using the marginal-posterior based criterion in
  Section~\ref{sec:sigma2choice}, search resolution
  $d=2$, and zero initial signal estimate:
\begin{equation}
  \label{eq:s0}
  \bm{s}^{(0)} = {\bf 0}_{p \times 1}
\end{equation}
with Matlab implementations available at
  \url{http://home.eng.iastate.edu/~ald/MPEM.html};
\item our \gls{MP-EM} algorithm in Section~\ref{sec:BayesianEM} with
  $\sigma^2$ \emph{tuned manually} for good performance (labeled
  \gls{MP-EM_OPT}) with $d=2$ and zero $\bm{s}^{(0)}$ in
  (\ref{eq:s0}), used as a benchmark;
\item the Gaussian-mixture version of the \gls{turbo-AMP} approach
   \cite{SomSchniter} with a Matlab implementation in
  \cite{TurboAMPLink} and the tuning hyperparameters chosen as the
  default values\footnote{These default values were designed 
for a set of approximately sparse wavelet coefficients of natural 
images, see \cite{SomSchniter}, which differ from the simulated signals
in Section~\ref{subsec:1dexample}.} in this implementation;

\item the \gls{FPC_AS} algorithm
  \cite{WenYinGoldfarbZhang}
  that aims at minimizing the Lagrangian cost function
\begin{subequations}
\begin{equation}
  \label{eq:BPDNcostfunction}
  0.5  \| \bm{y} - H  \bm{s} \|_2^2 + \tau  \| \bm{s} \|_1
\end{equation}
with the regularization parameter $\tau$ computed as
\begin{equation}
  \label{def:tuningparameter}
  \tau = 10^{a}  \| H^T  \bm{y} \|_{\infty}
\end{equation}
\end{subequations}
where $a$ is a tuning parameter chosen manually to achieve good reconstruction
performance;

\item the Barzilai-Borwein version of the \gls{GPSR} method with
  debiasing in \cite[Sec.\ III.B]{FigueiredoNowakWright} with the
  convergence threshold $\texttt{tolP} = 10^{-5}$ and tuning parameter
  $a$ in (\ref{def:tuningparameter}) chosen manually to achieve good
  reconstruction performance;

\item the \gls{NIHT} scheme
\cite{BlumensathDavies} initialized by
the zero $\bm{s}^{(0)}$ in (\ref{eq:s0});

\item the \gls{MB-IHT}
algorithm \cite{BaraniukCevherDuarteHegde} using a greedy tree
approximation \cite{Baraniuk}, initialized by
the zero $\bm{s}^{(0)}$ in (\ref{eq:s0});

\item the \gls{VB} tree-structured compressive sensing
  \cite{HeChenCarin} with a Matlab implementation in \cite{VBLink} and
  the tuning hyperparameters chosen as the default values in this
  implementation.\footnote{We scaled the sensing matrix $H$ by
    $\sqrt{p/\tr(H H^T)}$ prior to applying the \gls{VB} method, which
    helped improve its performance compared with using the unscaled
    $H$. This scaling is also applied in the \gls{turbo-AMP} implementation
    \cite{TurboAMPLink}.}

\end{itemize}
For the \gls{MP-EM}, \gls{NIHT}, and \gls{MB-IHT} iterations, we use the following
convergence criterion:
\begin{equation}
  \label{eq:convcrit}
  \frac{\| \bm{s}^{(j+1)} - \bm{s}^{(j)} \|_2^2}{p} < \delta
\end{equation}
where $\delta > 0$ is the convergence threshold selected in the
following examples so that the performances of the above methods do
not change significantly by further decreasing $\delta$.

For \gls{MP-EM}, we set the tuning constants in all following examples
as\footnote{The selections of $\gamma^2$ and $\epsilon^2$ in
\eqref{eq:tuningMP-EM} enforce a purely sparse signal model because
$\gamma^2 \gg \epsilon^2$.  When selecting $P_{\troot}, P_{\tH}$, and
$P_{\tL}$, we suggest to use \eqref{eq:numhighstate} and check that the
expected number of large-magnitude signal coefficients is roughly of
the order of the signal sparsity level that we expect. For example,
the selections in
\eqref{eq:tuningMP-EM} lead to the normalized expected
number of large-magnitude signal coefficients $\Exp[\sum_{i=1}^p q_i]/p = 0.0108$.}
\begin{equation}
  \gamma^2 = 1000, \quad
  \epsilon^2 = 0.1, \quad P_{\troot} = P_{\tH} =
  0.2, \quad P_{\tL} = 10^{-5}
  \label{eq:tuningMP-EM}
\end{equation}
which leads to $\frac{\Exp\left[\sum_{i=1}^p q_i\right]}{p} = 0.0108$. 

The sensing matrix $H$ has the following structure:
\begin{equation}
  \label{eq:H}
  H = \frac{1}{\rho_{\Phi}}  \Phi  \Psi
\end{equation}
where $\Phi$ is the $N \times p$ sampling matrix and $\Psi$ is the $p
\times p$ orthogonal transform matrix (satisfying $\Psi \Psi^T =
I_p$). Note that $H$ in \eqref{eq:H} satisfies the spectral norm
condition \eqref{eq:spectralnormofH}. We set the tree depth 
\begin{equation}
  \label{eq:L}
  L = 4.
\end{equation}

\subsection{Small-scale Structured Sparse Signal Reconstruction}
\label{subsec:1dexample}

\noindent We generated the binary state variables $\bq$ of length $p =
1024$ using the Markov tree model in
Section~\ref{sec:measuremodel}. Conditional on $q_i$, $s_i$ are
generated according to \eqref{eq:sgivenqprior2}. Here, the
matrix-to-vector conversion operator $\upsilon(\cdot)$ corresponds to
simple columnwise conversion, except for \gls{VB} whose implementation
\cite{VBLink} requires the use of Matlab's $\texttt{wavedec2}$
function for this purpose.  The sampling matrices $\Phi$ in \eqref{eq:H} have been
simulated using
\begin{enumerate}[(i)]
\item 
\label{whiteGaussiansensingmatrix}
a \emph{white Gaussian matrix} whose entries are \gls{iid} standard
Gaussian random variables,
\item 
\label{correlatedGaussiansensingmatrix}
a \emph{row-correlated Gaussian matrix} with \gls{iid} zero-mean
Gaussian columns (indexed by $k=1,2,\ldots,p$) having covariance
matrix whose $(i,j)$th element is 
\begin{subequations}
\begin{equation}
  \label{eq:rowcormatrix}
  \cov( \Phi_{i,k}, \Phi_{j,k} ) =
r^{|i-j|}, \qquad i,j = 1,2,\ldots,N
\end{equation}
useful, e.g., in modeling time-series data
\cite[Sec. 5]{RaskuttiWainwrightYu}, and
\item 
\label{correlatedGaussiansensingmatrix2}
a \emph{column-correlated Gaussian matrix} with \gls{iid}
  zero-mean Gaussian rows (indexed by $k=1,2,\ldots,N$)
having covariance matrices whose $(i,j)$th
  element is 
\begin{equation}
  \label{eq:colcormatrix}
\cov( \Phi_{k,i}, \Phi_{k,j} ) = c^{|i-j|}, \qquad i,j =
  1,2,\ldots,p.
\end{equation}
\end{subequations}
\end{enumerate}
The general column correlation model
(\ref{correlatedGaussiansensingmatrix2}) for the design (sensing)
matrices is analyzed in \cite{RaskuttiWainwrightYu,Wainwright2009a,Wainwright2009b}, see also
\cite{Bach,ObozinskiWainwrightJordan,NguyenTran}, which employ this
correlation structure. Correlations among columns of the design
matrices occur e.g., in genomic applications
\cite[Sec. 18.4]{HastieTibshiraniFriedman} and spatially correlated
designs are relevant to \gls{fMRI} \cite{VaroquauxGramfortThirion}.

The transform matrix $\Psi$ in \eqref{eq:H} is chosen to be identity:
\begin{equation}
  \label{eq:Psismallscale}
  \Psi = I_p
\end{equation}
hence, in this example, the sampling and sensing matrices $\Phi$ and
$H$ are the same up to a proportionality constant.

We simulate the observation vectors $\bm{y}$ using the measurement and
prior models in \eqref{eq:awgn}, \eqref{eq:sgivenqprior0}, and
\eqref{eq:priorforq} and following model parameters:
\begin{equation}
  \label{eq:modelparameters}
  \epsilon_\star^2 = 1, \quad \sigma_\star^2 = 10^{-6}, \quad 
( P_{\troot} )_{\star} =  (  P_{\tH} )_{\star} = 0.5, \quad ( P_{\tL}
)_{\star} = 10^{-4}, 
\quad \gamma_\star^2 \in \{ 10^3,  10^4,  10^5 \}
\end{equation}
where the subscripts $\star$ emphasize that these selections are the
true model parameters employed to simulate the measurements and are
generally \emph{different} from the tuning constants
\eqref{eq:tuningMP-EM} employed by the \gls{MP-EM} method.  Here, our
goal is to show the performance of the \gls{MP-EM} method in the case
where there is a mismatch between the tuning parameters and
corresponding true model parameters.  The choices $(P_{\tH})_{\star},
(P_{\troot})_{\star}$, and $( P_{\tL} )_{\star}$ in
\eqref{eq:modelparameters} correspond to the normalized expected
number of large-magnitude signal coefficients
\begin{equation}
  \label{eq:Eq}
  \frac{\Exp\bigl[\sum_{i=1}^p q_i\bigr]}{p} = 0.0919
\end{equation}
computed using \eqref{eq:numhighstate}.  We vary the values of $\gamma_\star^2$
to test the performances of various methods at different \glspl{SNR}.

Our performance metric is the \emph{average} \gls{NMSE} of an estimate
$\widetilde{\bm{s}}$ of the signal coefficient vector 
(used also in e.g., \cite{SchniterPotterZiniel}):
\begin{eqnarray}
  \label{def:NMSE1} \NMSE\{\widetilde{\bm{s}}\} =
  \Exp_{\Phi,\bm{s},\bm{y}} \biggl[ \frac{\|\widetilde{\bm{s}}  -  \bm{s} \|^2_2}	{\| \bm{s} \|_2^2} \biggr]
\end{eqnarray}
computed using $500$ Monte Carlo trials, where \emph{averaging} is
performed over the random Gaussian sampling matrices $\Phi$, signal
$\bm{s}$, and measurements $\bm{y}$.

We select the convergence threshold in
\eqref{eq:convcrit} to 
\begin{equation}
  \label{eq:deltaex1}
  \delta = 10^{-10}.
\end{equation}
For \gls{MP-EM} and \gls{MP-EM_OPT}, we set the grid length $K = 16$.
The tuning parameters for \gls{MP-EM} are given in \eqref{eq:tuningMP-EM}.\looseness=-1

The \gls{NIHT} and \gls{MB-IHT} methods require knowledge of the
signal sparsity level (i.e., an upper bound on the number of nonzero
coefficients); in this example, we set the signal sparsity level for
these methods to the exact number of large-magnitude signal
coefficients $\sum_{i=1}^p q_i$.  For \gls{GPSR} and \gls{FPC_AS}, we
vary $a$ in \eqref{def:tuningparameter} within the set $\{-1, -2, -3,
-4, -5, -6, -7, -8, -9\}$ and, for each $N / p$ and each of the two
methods, we use the optimal $a$ that achieves the smallest \gls{NMSE}.

The \gls{turbo-AMP} implementation  in \cite{TurboAMPLink}
requires a function input \texttt{xRange} that corresponds to the
range of the input signal $\Psi \bm{s}$. In this example, we set the
value of this tuning constant to six standard deviations of the signal
coefficients in $\bm{s}$: 
\begin{equation}
  \texttt{xRange} = 6  \sigma_{\star}  \sqrt{\frac{\Exp\bigl[\sum_{i=1}^p q_i\bigr]}{p}  \gamma_{\star}^2  + \Bigl( 1 - \frac{\Exp\bigl[\sum_{i=1}^p q_i\bigr]}{p} \Bigr)   \epsilon_{\star}^2 }
\label{eq:xRange}
\end{equation}
where $\sigma_{\star} = \sqrt{\sigma_{\star}^2}$; \gls{turbo-AMP} with
this selection performs well compared with other choices of
\texttt{xRange} that we tested. Selecting too small or too large
\texttt{xRange} would lead to deteriorated performance of \gls{turbo-AMP}.
 \Gls{turbo-AMP} is particularly sensitive to underestimation of this quantity
and less sensitive to selecting larger values than optimal.

\subsubsection{White and Row-correlated Sensing Matrices}
\label{sec:whiteandrow-correlatedsensingmatrices}

\begin{figure}[!t]
  \subfigure[]{\includegraphics[width=0.475\linewidth]{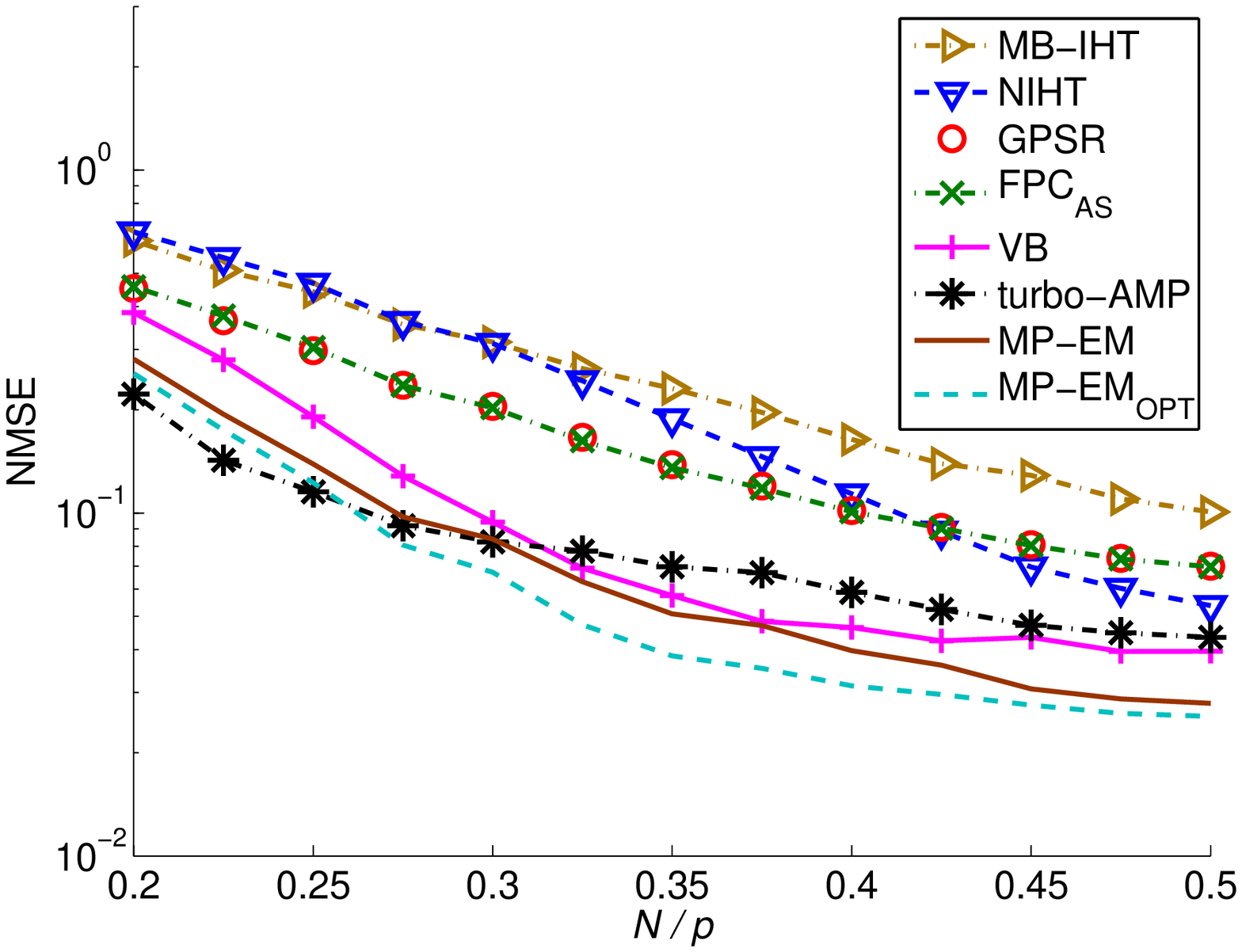}
    \label{fig:NMSE_White_gamma2_1e3} }
  \subfigure[]{\includegraphics[width=0.475\linewidth]{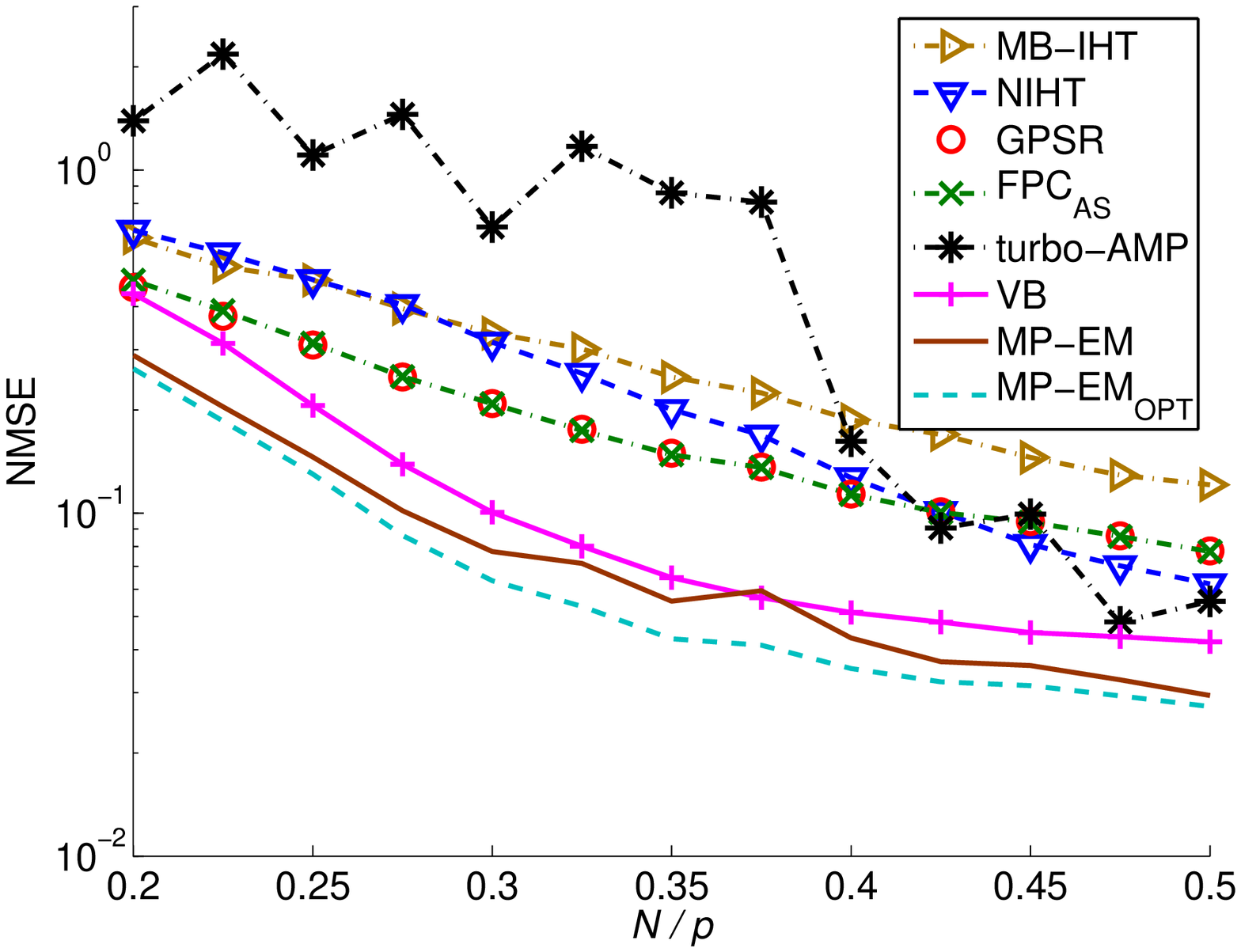}
    \label{fig:NMSE_Corr_gamma2_1e3} }\hfil
  \subfigure[]{\includegraphics[width=0.475\linewidth]{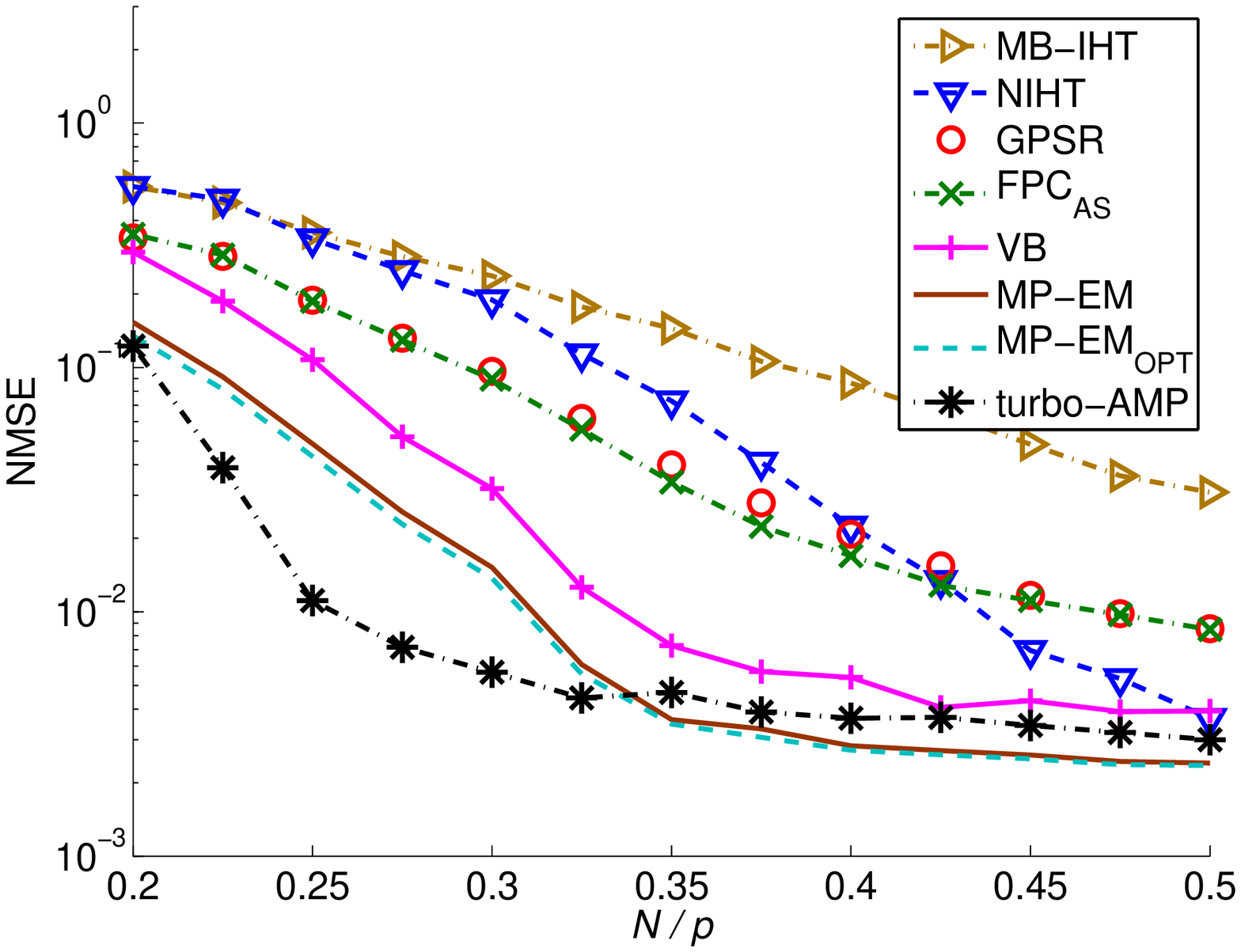}
    \label{fig:NMSE_White_gamma2_1e4} }
  \subfigure[]{\includegraphics[width=0.475\linewidth]{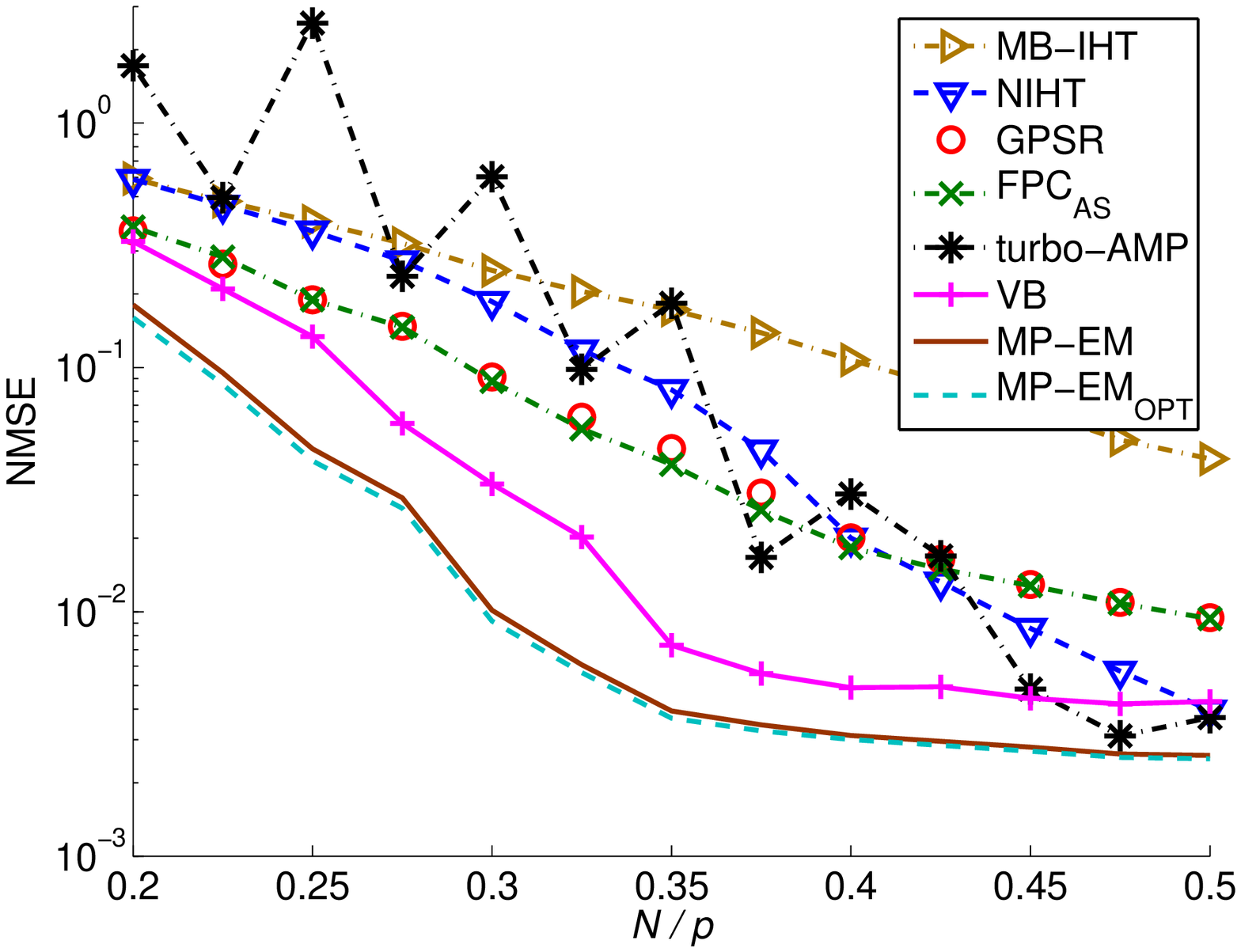}
    \label{fig:NMSE_Corr_gamma2_1e4} } \hfil
  \subfigure[]{\includegraphics[width=0.475\linewidth]{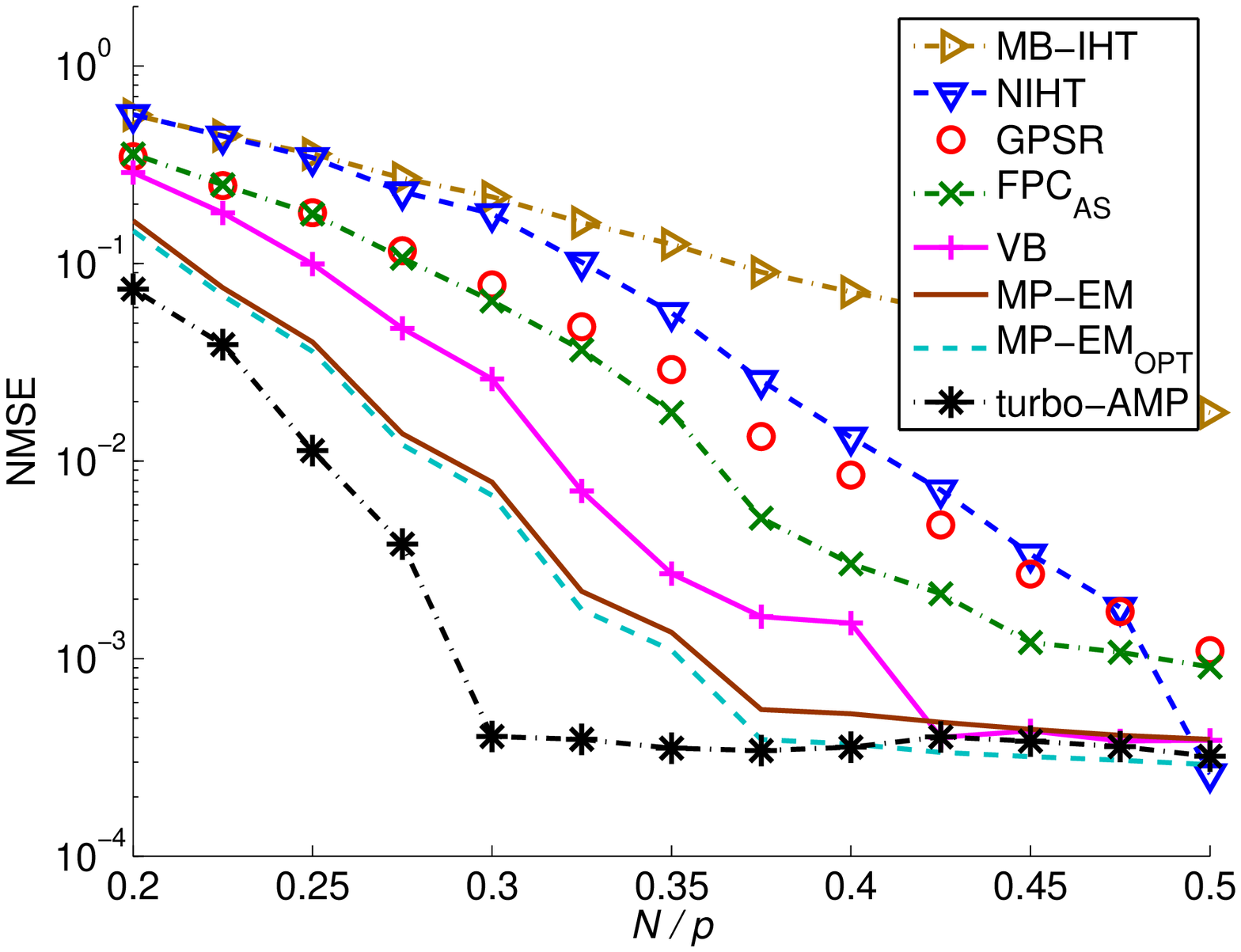}
    \label{fig:NMSE_White_gamma2_1e5} }
  \subfigure[]{\includegraphics[width=0.475\linewidth]{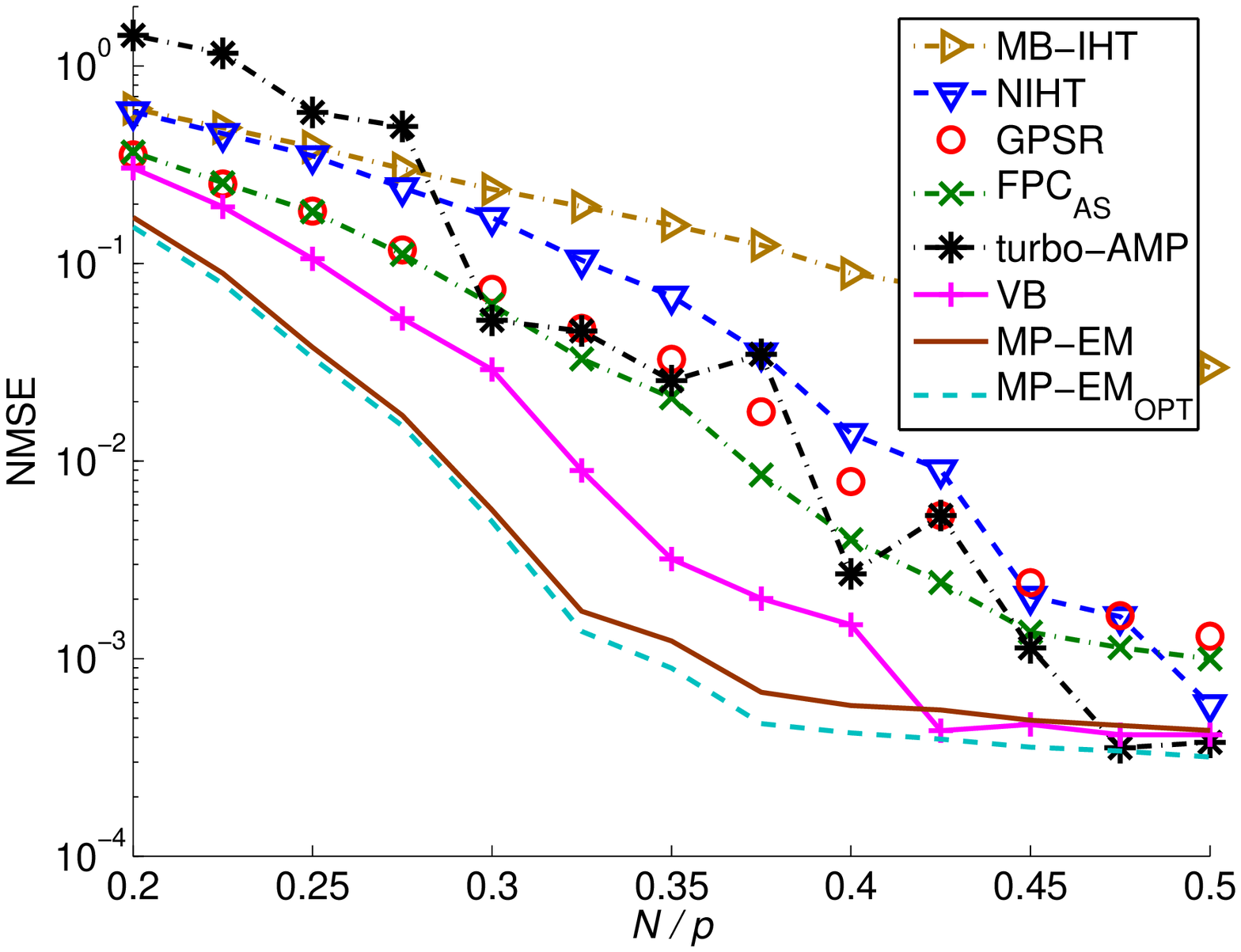}
    \label{fig:NMSE_Corr_gamma2_1e5} } \hfil \centering
  \caption{\glspl{NMSE} as functions of the subsampling factor $N / p$
    for (a)-(b) low \gls{SNR} with $\gamma_\star^2 = 10^3$, (c)-(d)
    medium \gls{SNR} with $\gamma_\star^2 = 10^4$, and (e)-(f) high
    \gls{SNR} with $\gamma_\star^2 = 10^5$ using [left: (a), (c), (e)]
    white and [right: (b), (d), (f)] row-correlated sensing matrices
    with correlation parameter $r=0.2$, respectively.}
  \label{fig:NMSE_smallscale}
  \vspace{-0.2in}
\end{figure}

Fig.~\ref{fig:NMSE_smallscale} shows the \glspl{NMSE} of different
methods as functions of the subsampling factor $N / p$ for the three
choices of $\gamma_\star^2$ in \eqref{eq:modelparameters}, corresponding to relatively low, medium, and high
\glspl{SNR}, and white and row-correlated sensing matrices with
correlation parameter $r=0.2$ in \eqref{eq:rowcormatrix}.  Here, a
larger value of the high-signal relative variance $\gamma_\star^2$
implies a relatively higher \gls{SNR}.  Indeed, for each method, the
signal with higher \gls{SNR} can be reconstructed with a smaller
\gls{NMSE} than the signal with lower \gls{SNR}: Compare
Figs.~\ref{fig:NMSE_White_gamma2_1e3}, \ref{fig:NMSE_White_gamma2_1e4},
and \ref{fig:NMSE_White_gamma2_1e5} as well as
Figs.~\ref{fig:NMSE_Corr_gamma2_1e3}, \ref{fig:NMSE_Corr_gamma2_1e4},
and \ref{fig:NMSE_Corr_gamma2_1e5}.

For white Gaussian sampling matrices, the methods that employ the
probabilistic tree structure of the signal coefficients
(\gls{turbo-AMP}, \gls{MP-EM}, \gls{MP-EM_OPT}, and \gls{VB}) clearly
outperform all other approaches, see
Figs.~\ref{fig:NMSE_White_gamma2_1e3},
\ref{fig:NMSE_White_gamma2_1e4}, and \ref{fig:NMSE_White_gamma2_1e5}.
For row-correlated Gaussian sampling matrices, \gls{MP-EM} and
\gls{MP-EM_OPT} achieve the best overall performances, followed by the
\gls{VB} method; \gls{turbo-AMP} is sensitive to introducing
correlation among elements of the sampling matrix $\Phi$ and performs
poorly for smaller $N / p$, see Figs.~\ref{fig:NMSE_Corr_gamma2_1e3},
\ref{fig:NMSE_Corr_gamma2_1e4}, and \ref{fig:NMSE_Corr_gamma2_1e5}.

For white Gaussian sampling matrices, we observe the following: 
\begin{itemize}
\item  at low \gls{SNR}, \gls{MP-EM}
and \gls{MP-EM_OPT} outperform other approaches when 
 $N / p > 0.275$,  see
Fig.~\ref{fig:NMSE_White_gamma2_1e3};
\item at medium and high \glspl{SNR}, \gls{turbo-AMP} achieves the
  best overall performance, followed by \gls{MP-EM_OPT} and
  \gls{MP-EM}, see Figs.~\ref{fig:NMSE_White_gamma2_1e4} and
  \ref{fig:NMSE_White_gamma2_1e5}.
\end{itemize}

The \gls{NIHT} method performs relatively poorly for smaller $N / p$,
but improves as $N / p$ increases.  For sufficiently high $N / p$,
\gls{NIHT} achieves smaller \glspl{NMSE} than other methods that do not
exploit the probabilistic tree structure.

In Fig.~\ref{fig:NMSE_smallscale}, the \glspl{NMSE} of \gls{MP-EM} are
close to those of \gls{MP-EM_OPT}, which implies that the
marginal-posterior based criterion in Section~\ref{sec:sigma2choice}
selects the noise variance parameter well in this example.

The performance of \gls{turbo-AMP} deteriorates with introduction of
correlation among elements of the sampling matrix $\Phi$: The
\glspl{NMSE} of \gls{turbo-AMP} for some subsampling factors are more
than an order of magnitude larger for row-correlated sampling matrices
than for white sampling matrices. In contrast, the \glspl{NMSE} for
all the other methods increase only slightly when we introduce
sampling matrix correlation \eqref{eq:rowcormatrix}, compare the left
and right-hand sides of Fig.~\ref{fig:NMSE_smallscale}. Increasing
this correlation by increasing $r$ to $0.3$ in \eqref{eq:rowcormatrix}
results in further performance deterioration of \gls{turbo-AMP} (i.e.,
\gls{turbo-AMP} has very high \glspl{NMSE} for all $N / p$ in this
case), whereas the competing methods continue to perform well.

\begin{figure}[!t]
  \subfigure[]{\includegraphics[width=0.475\linewidth]{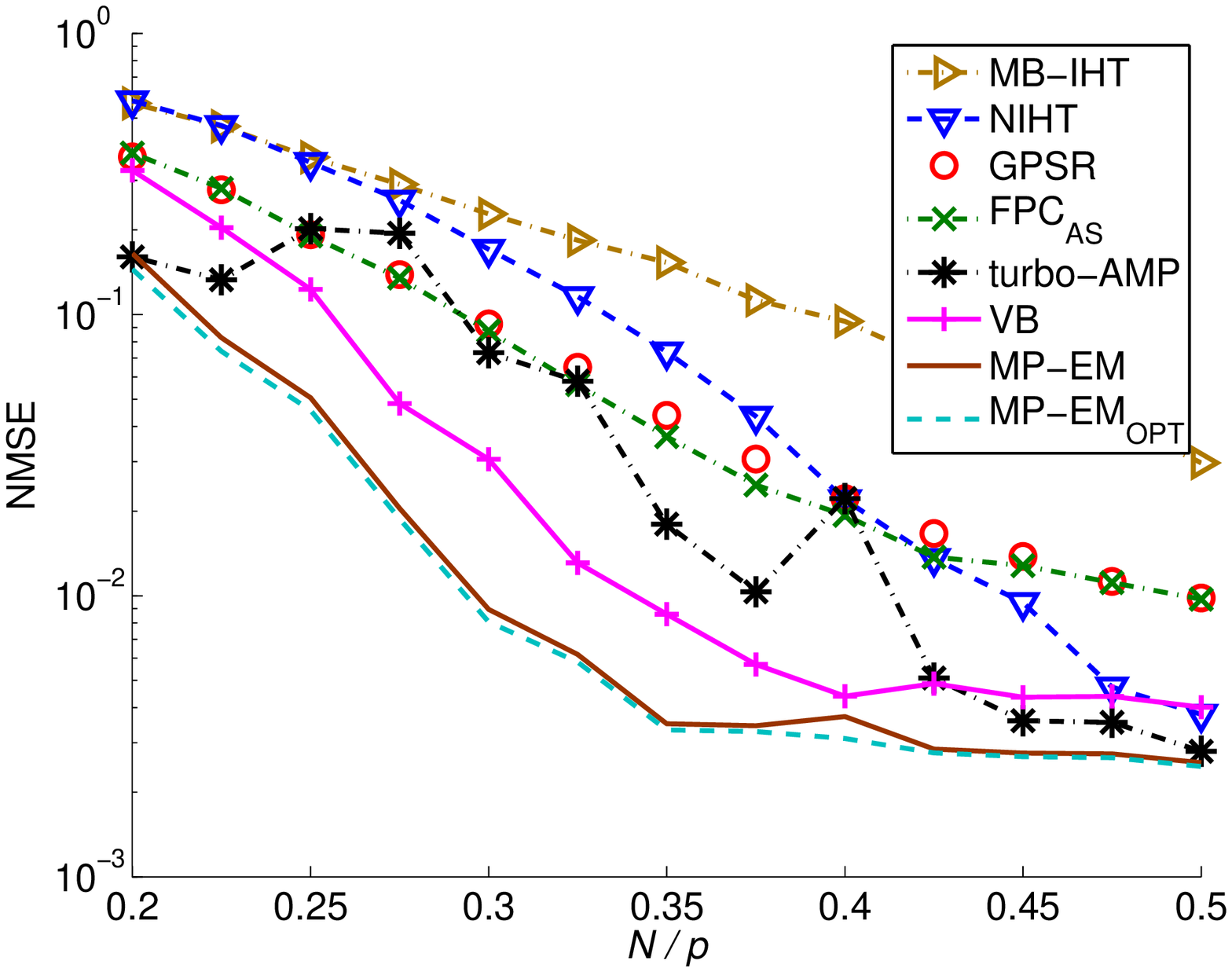}
    \label{fig:NMSE_CorrCOL_gamma2_1e4} }
  \subfigure[]{\includegraphics[width=0.475\linewidth]{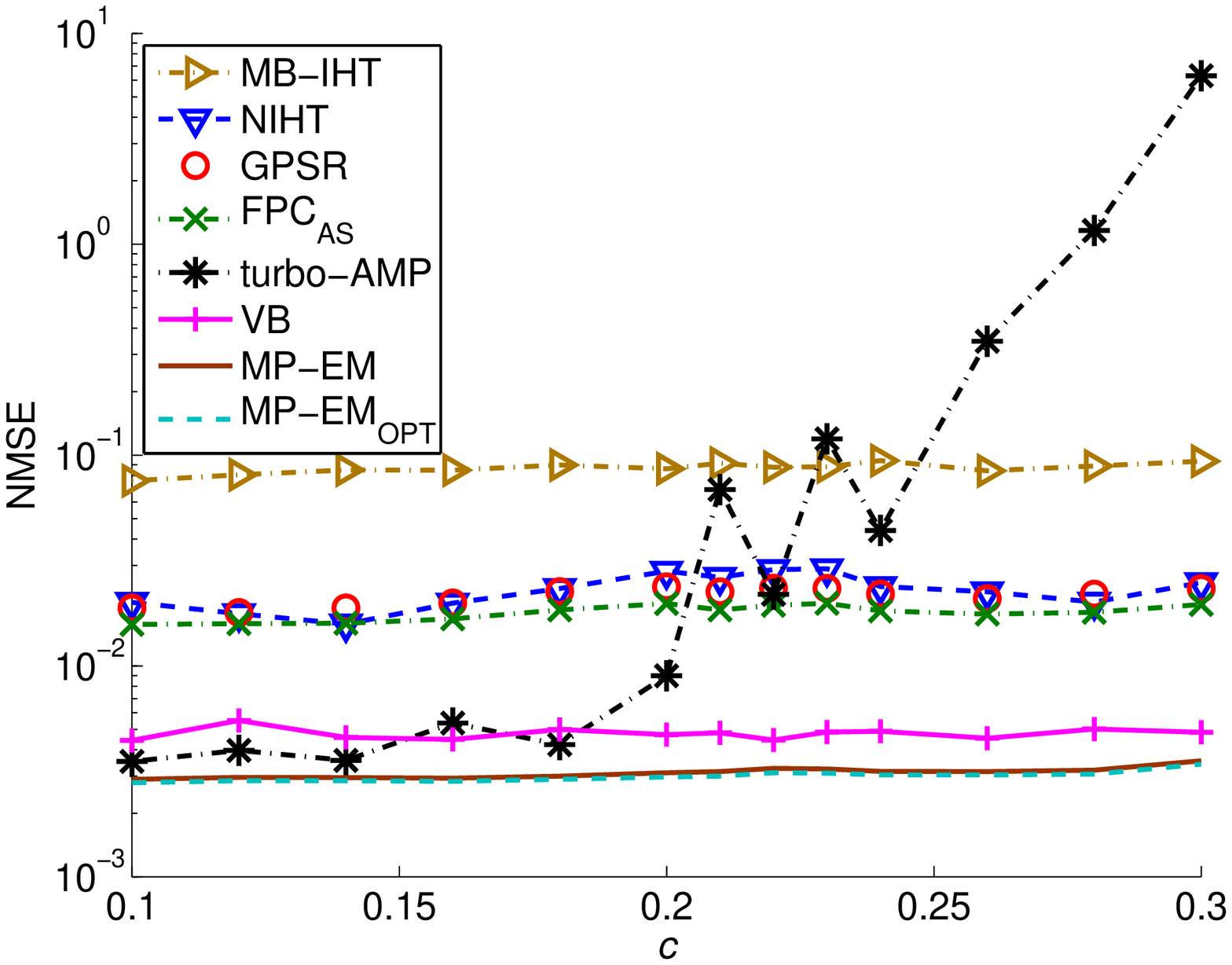}
    \label{fig:NMSE_CorrCOL_gamma2_1e4_diffC} }\hfil
  \centering
  \caption{Column-correlated sensing matrices: \glspl{NMSE} as
    functions of (a) the subsampling factor $N / p$ for correlation
    parameter $c = 0.2$ and (b) $c$ for $N / p = 0.4$ under the medium
    \gls{SNR} scenario with $\gamma_\star^2 = 10^4$.}
  \label{fig:NMSE_smallscale2}
  \vspace{-0.2in}
\end{figure}

The \gls{VB} method performs well under both white and row-correlated
sensing matrix scenarios and \gls{turbo-AMP} has a superior
reconstruction performance under the white sensing matrix
scenario. These good performances are likely facilitated by the fact
that \gls{VB} and \gls{turbo-AMP} \emph{learn} the Markov tree
parameters from the measurements.

\subsubsection{Column-correlated Sensing Matrices}
\label{sec:column-correlatedsensingmatrices}

Fig.~\ref{fig:NMSE_CorrCOL_gamma2_1e4} shows the \glspl{NMSE} of
different methods as functions of the subsampling factor $N / p$ 
for  column-correlated sampling matrices
 having the correlation constant $c = 0.2$ in
\eqref{eq:colcormatrix}
under
the medium \gls{SNR} scenario. Here, \gls{MP-EM} and \gls{MP-EM_OPT} have the
smallest \glspl{NMSE} over nearly the entire range of $N / p$ considered. 
Fig.~\ref{fig:NMSE_CorrCOL_gamma2_1e4_diffC} shows the \glspl{NMSE} as
functions of $c$ for $N / p$ fixed at $0.4$.  Here, only
\gls{turbo-AMP} is very sensitive to the presence of correlations
among the elements of the sampling matrix, whereas all other methods
vary only slightly as functions of $c$.

We also observe numerical instability of \gls{turbo-AMP} when
correlated Gaussian sampling matrices are employed, which is exhibited
by the oscillatory behavior of its \glspl{NMSE} in the right side of
Fig.~\ref{fig:NMSE_smallscale} and in Fig.~\ref{fig:NMSE_smallscale2}
[demanding more averaging than the $500$ Monte Carlo trials that we
employ to estimate \eqref{def:NMSE1}].

We simulated sampling matrices $\Phi$ that have variable
column norms or row norms, which led to deteriorating performances of
\gls{turbo-AMP} in both cases, whereas the competing methods perform
well. The fact that \gls{turbo-AMP} has been derived assuming Gaussian
sensing matrices with \gls{iid} elements explains its poor
performance for sensing matrices that deviate sufficiently from this
assumption.

The \gls{MB-IHT} method, which employs a greedy tree approximation and
deterministic tree structure, achieves quite a poor \gls{NMSE}
performance in Figs.~\ref{fig:NMSE_smallscale} and \ref{fig:NMSE_smallscale2}.
A relatively poor
performance of \textsc{mb-cosamp} (which employs the same
deterministic tree structure) has also been reported in
\cite[Sec.~IV.B]{SomSchniter}.

\subsection{Image Reconstruction}
\label{subsec:2dexample}

We reconstruct  $128 \times 128$ and $256 \times 256$ test images
 from noiseless compressive samples
($\sigma_\star^2 = 0$).
Here, the matrix-to-vector conversion operator $\upsilon(\cdot)$ is
based on the columnwise conversion for $128 \times 128$ images, and
Matlab wavelet decomposition function $\texttt{wavedec2}$ with Haar
wavelet for $256 \times 256 $ images, which has also been used in
\cite{HeCarin} and \cite{SomSchniter}.  Before taking the wavelet
transform, we subtract the mean of original image to ensure that $\Psi
\bm{s}$ has zero mean.

For \gls{turbo-AMP}, we set the function input \texttt{xRange} to
$255$, which is the difference between the minimum and maximum
possible image values in this example.\footnote{The
  authors thank Dr.\ Subhojit Som from Microsoft Inc.\ for the
  correspondence with regard to setting this parameter. \label{Som}}  Observe that
the \gls{turbo-AMP} implementation in \cite{TurboAMPLink} needs
additional prior information about the signal range, which is not
required by other methods.

\subsubsection{Medium scale with row-correlated Gaussian sampling matrices}
\label{subsubsec:2dexampleiid}

\noindent We reconstruct the $128 \times 128$ `Cameraman' image (cropped from 
the original $256 \times 256$ image in Fig.~\ref{Img:Cameraman}, as was
also done in \cite{VBLink, TurboAMPLink} and corresponding papers \cite{HeChenCarin, SomSchniter}) from 
compressive samples  generated using
row-correlated Gaussian sampling matrices
with covariances between the elements described by \eqref{eq:rowcormatrix}.
 Our performance metric is the \gls{NMSE} in \eqref{def:NMSE1}
computed using $10$ Monte Carlo trials, where the averaging is
performed only over the random Gaussian sampling matrices $\Phi$.

In this example, the convergence threshold in \eqref{eq:convcrit} is set to
\begin{equation}
  \label{eq:deltaex3}
  \delta = 0.01.
\end{equation}
For \gls{MP-EM} and \gls{MP-EM_OPT}, we set the grid length $K =
16$. The tuning parameters for \gls{MP-EM} are given in \eqref{eq:tuningMP-EM}.\looseness=-1

We set the sparsity level $r$ for \gls{NIHT} as $2000 N / p$
and $2500 N / p$ for \gls{MB-IHT}, tuned for good \gls{NMSE} performance.

\begin{figure}
\subfigure[]{\includegraphics[width=0.475\linewidth]{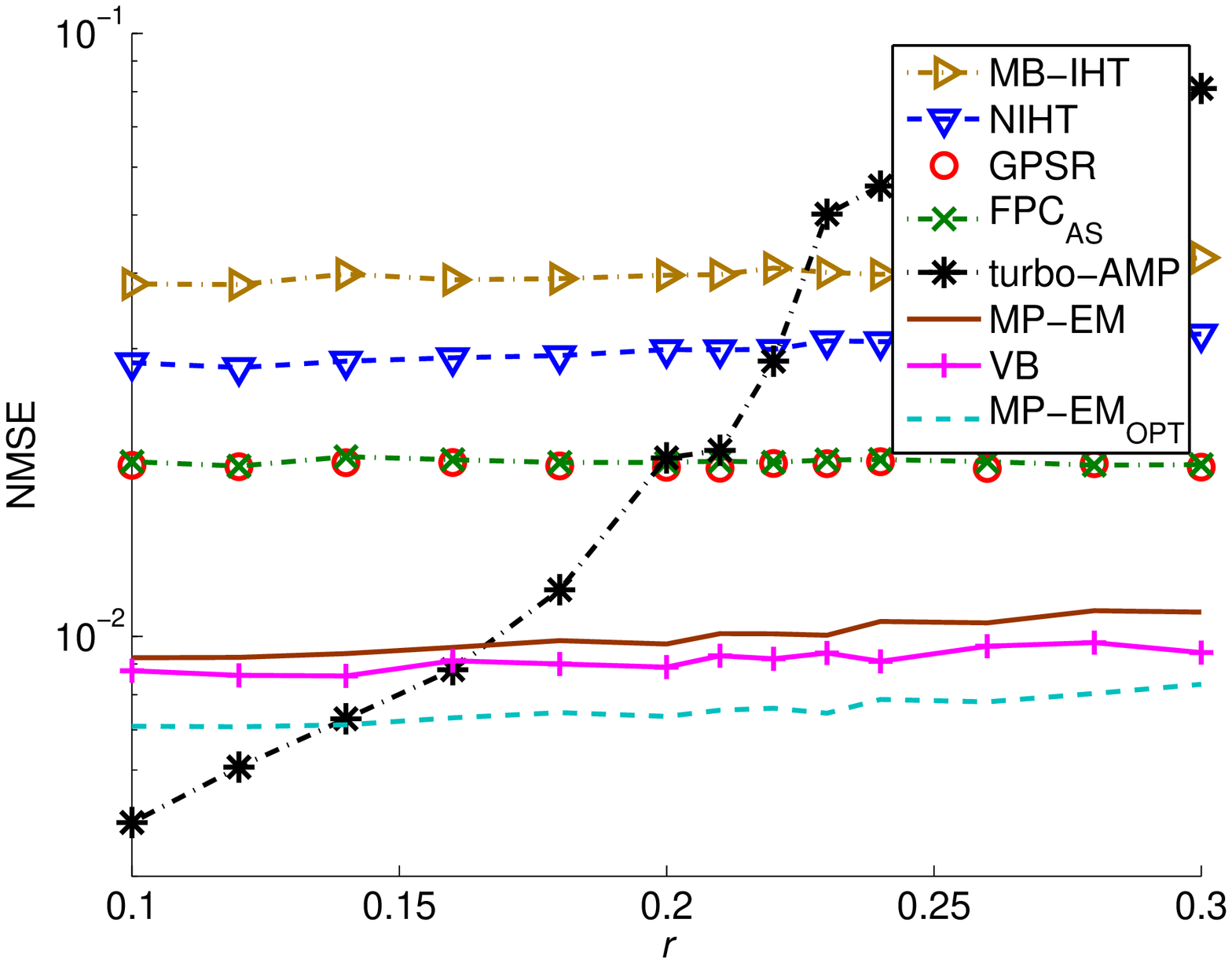}
\label{fig:NMSE128Cameraman4level} } \hfil \centering
\subfigure[]{\includegraphics[width=0.475\linewidth]{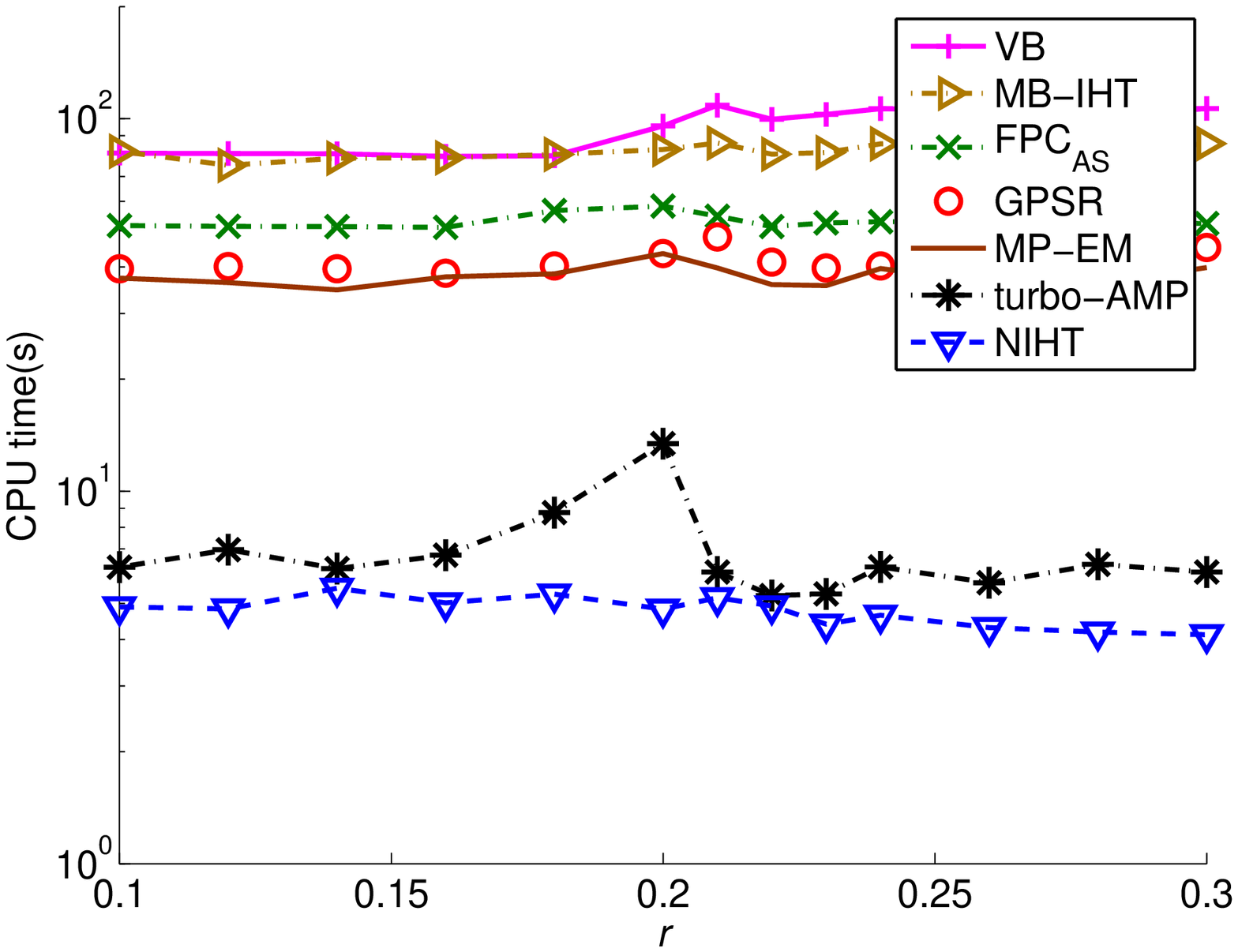}
\label{fig:CPUTime128Cameraman4level} } 
\caption{(a) \glspl{NMSE} and (b) \gls{CPU} times as functions of the
  correlation parameter $r$ for the $128 \times 128$ `Cameraman' image
  when $N / p = 0.3$.}
\label{fig:CameramanIID}
\end{figure}

Fig.~\ref{fig:CameramanIID} shows the \glspl{NMSE} and \gls{CPU} times
of different methods reconstructing the $128 \times 128$ `Cameraman'
image as functions of the correlation parameter $r$ in
\eqref{eq:rowcormatrix} with $N/p = 0.3$.  Since \gls{MP-EM} and
\gls{MP-EM_OPT} have the same runtime, we report only that of
\gls{MP-EM} in Fig.~\ref{fig:CPUTime128Cameraman4level}.
\Gls{turbo-AMP} has the smallest \gls{NMSE} when $N / p \leq
0.12$. However, its \gls{NMSE} increases sharply as $r$ becomes
larger: \gls{turbo-AMP} has the largest \gls{NMSE} when $N / p > 0.22
$. In contrast, the \glspl{NMSE} for all the other methods keep nearly
constants as we increase $r$. The \gls{MP-EM}, \gls{MP-EM_OPT}, and
\gls{VB} methods have smaller \glspl{NMSE} than \gls{GPSR},
\gls{FPC_AS}, \gls{NIHT}, and \gls{MB-IHT} for all the correlation
coefficients $r$ considered. The \gls{VB} approach performs slightly
better than \gls{MP-EM}, but is slower than \gls{MP-EM} and
\gls{MP-EM_OPT}. In terms of \gls{CPU} time, \gls{NIHT} is the fastest
among all the methods compared and \gls{turbo-AMP} requires
\SIrange{0.3}{8.6}{\second} more than \gls{NIHT}, both of which are
faster than the remaining methods.\footnote{Regarding the reported
  \gls{CPU} time, note that the \gls{turbo-AMP} code does not use
  Matlab only, but combines Matlab and JAVA codes.}  The \gls{VB}
scheme consumes the largest amount of \gls{CPU} time among all the
methods for all the correlation coefficient $r$ considered;
\gls{MP-EM} and \gls{MP-EM_OPT} are faster than \gls{GPSR},
\gls{FPC_AS}, \gls{MB-IHT}, and \gls{VB}.

As before, the good performance of \gls{VB} is likely facilitated by
the fact that it learns the Markov tree parameters from the
measurements.

\begin{figure}
\centering
\subfigure[\gls{MP-EM_OPT} (\gls{PSNR} \SI{26.3}{\decibel})]{\includegraphics[width=1.67in]{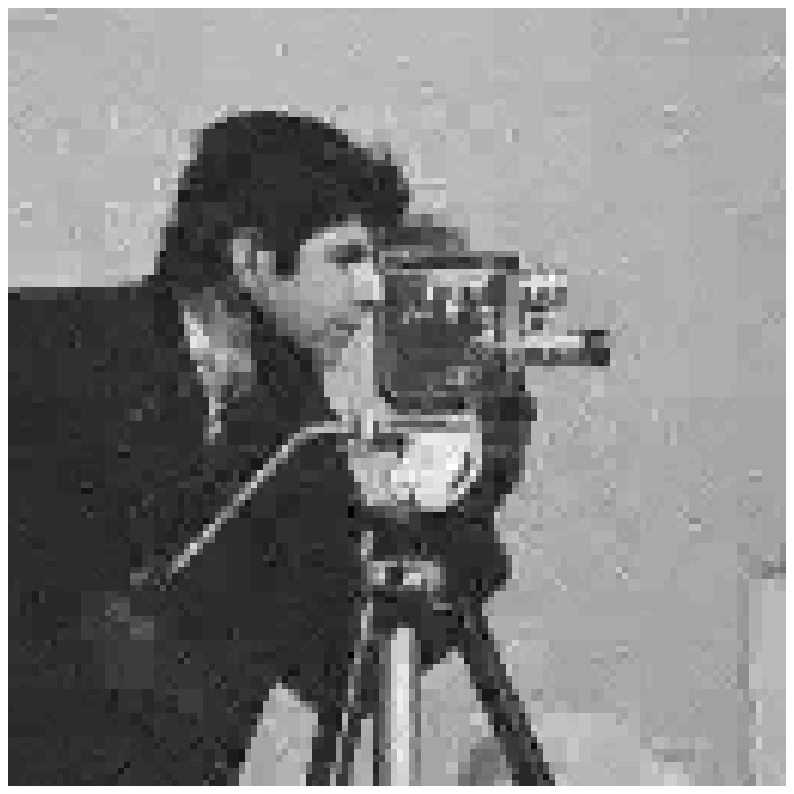}
\label{Img:Cameraman128MPEM-OPT} } 
\subfigure[\gls{VB} (\gls{PSNR} \SI{25.8}{\decibel})]{\includegraphics[width=1.67in]{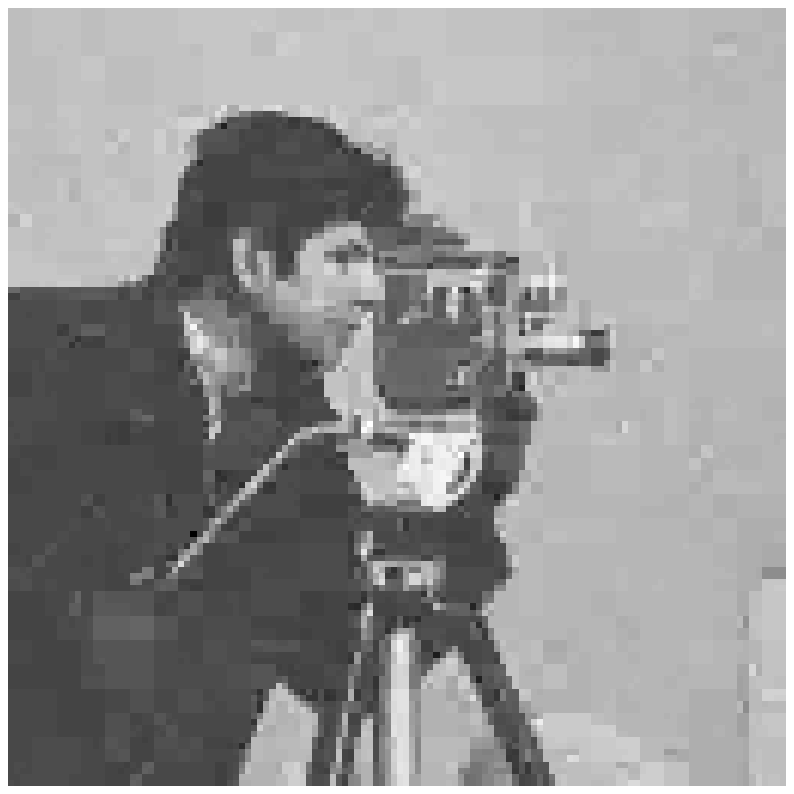}
\label{Img:Cameraman128VB} } 
\subfigure[\gls{MP-EM} (\gls{PSNR} \SI{24.9}{\decibel})]{\includegraphics[width=1.67in]{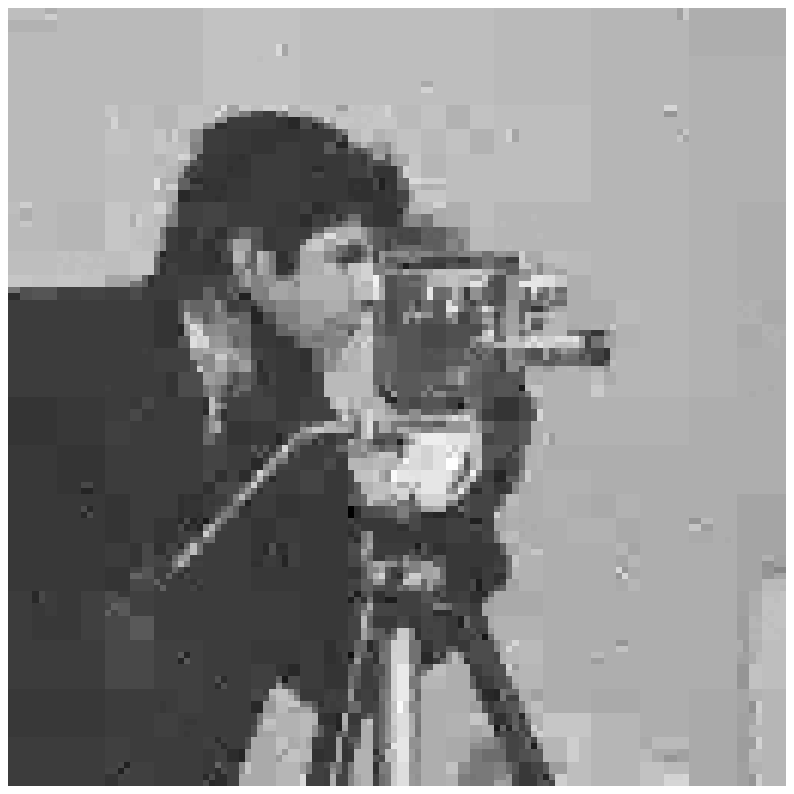}
\label{Img:Cameraman128MPEM} } 
\subfigure[\gls{turbo-AMP} (\gls{PSNR} \SI{23.8}{\decibel})]{\includegraphics[width=1.67in]{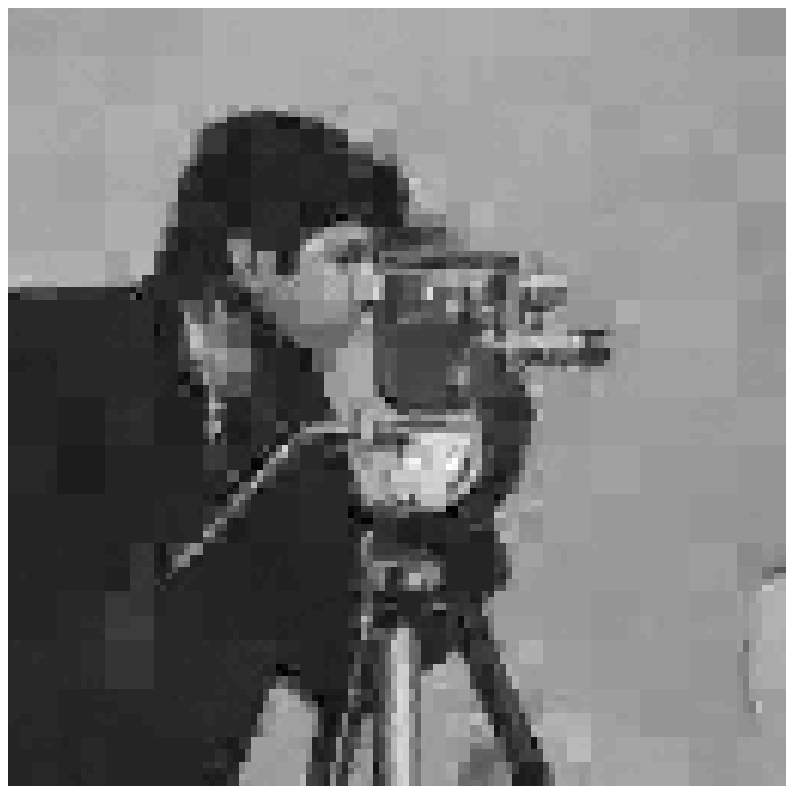}
\label{Img:Cameraman128TurboAMP} } 
\subfigure[\gls{FPC_AS} (\gls{PSNR} \SI{22.0}{\decibel})]{\includegraphics[width=1.67in]{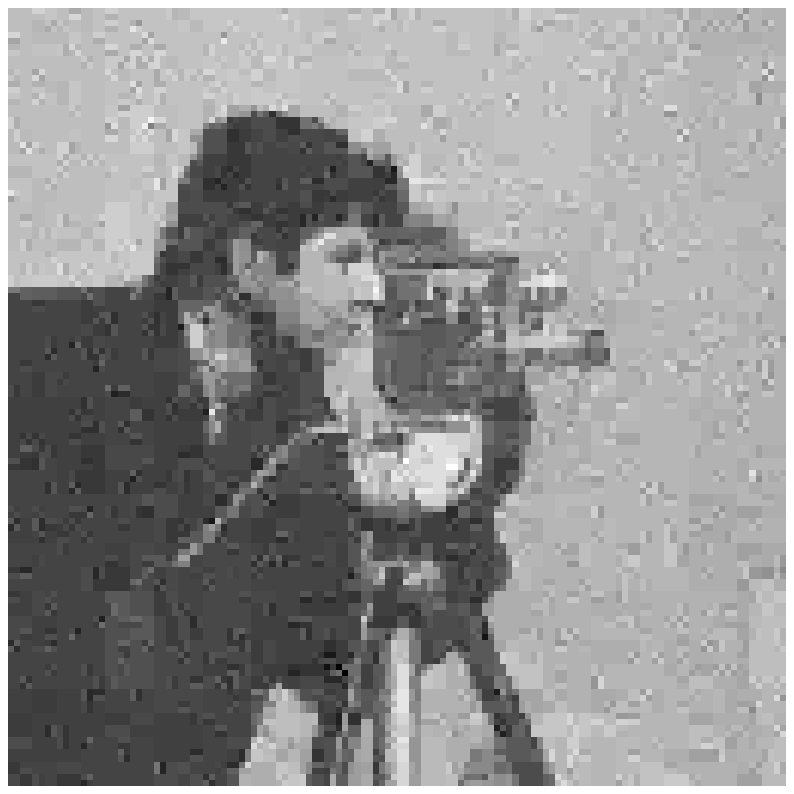}
\label{Img:Cameraman128FPCAS} } 
\subfigure[\gls{GPSR} (\gls{PSNR} \SI{22.0}{\decibel})]{\includegraphics[width=1.67in]{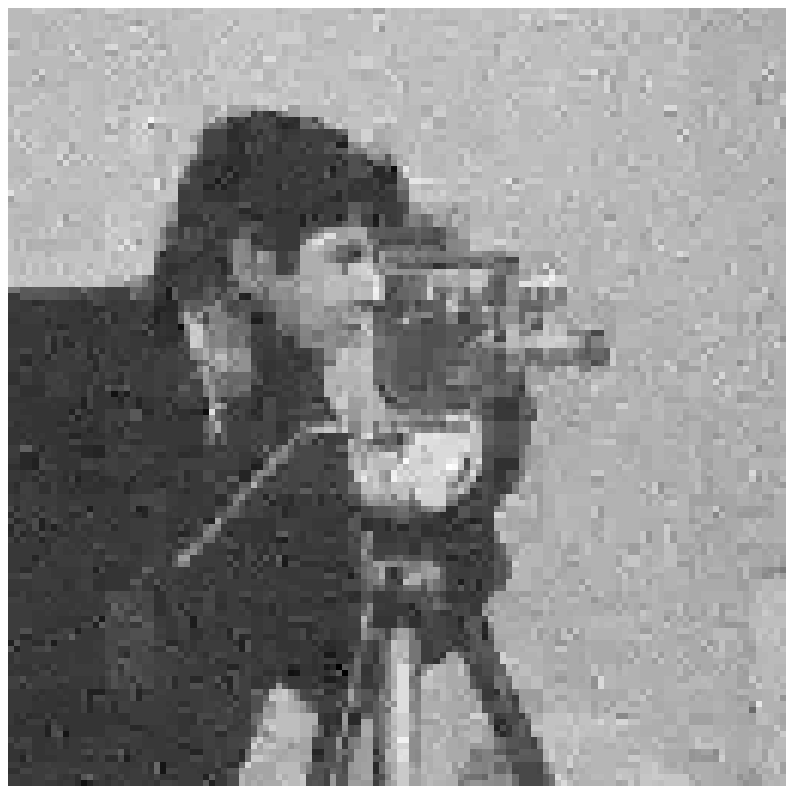}
\label{Img:Cameraman128GPSR} } 
\subfigure[\gls{NIHT}  (\gls{PSNR} \SI{20.2}{\decibel})]{\includegraphics[width=1.67in]{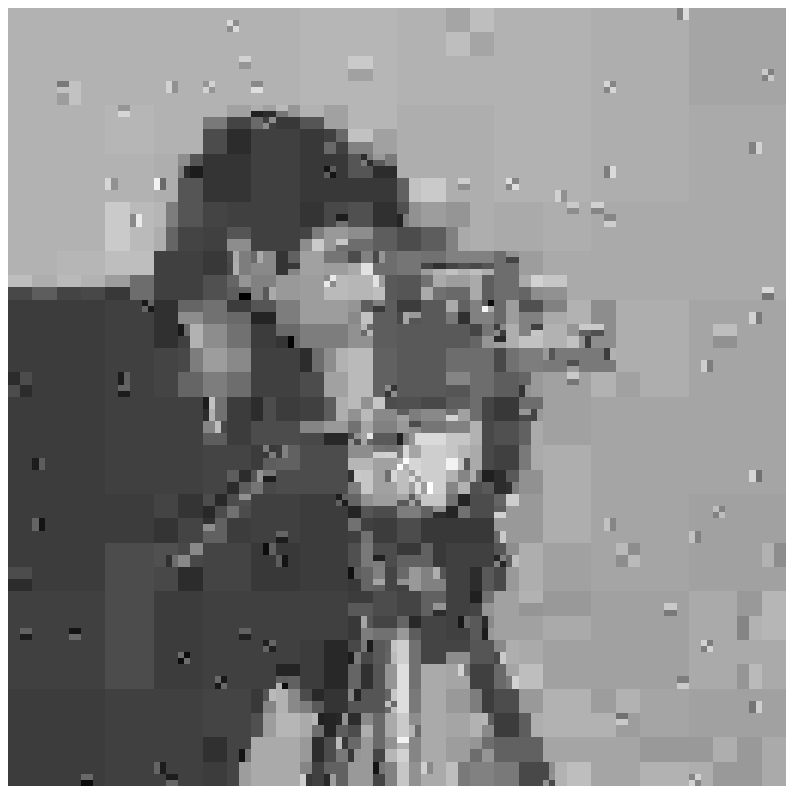}
\label{Img:Cameraman128NIHT} } 
\subfigure[\gls{MB-IHT} (\gls{PSNR} \SI{18.7}{\decibel})]{\includegraphics[width=1.67in]{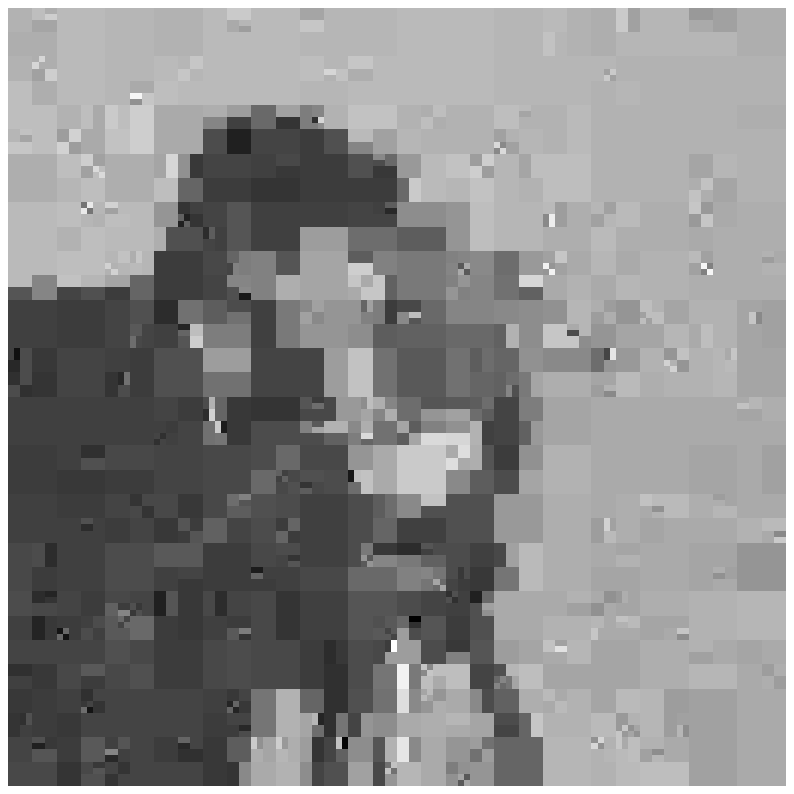}
\label{Img:Cameraman128MBIHT} } 

\caption{The $128 \times 128$ `Cameraman' image reconstructed by various methods
for $r = 0.2$ and $N / p = 0.3$.}

\label{img:cameraman128} \vspace{-0.2in}
\end{figure}

Fig.~\ref{img:cameraman128} shows the reconstructed $128 \times 128$
`Cameraman' image by different methods for $N/p = 0.3$ and $r = 0.2$
using one realization of the sampling matrix $\Phi$.
In Fig.~\ref{img:cameraman128}, we also report the 
\glspl{PSNR} of these methods, where the \gls{PSNR}
of an estimated signal
$\widetilde{\bm{s}}$ is defined as \cite[eq. (3.7)]{StarckMurtaghFadili}:
\begin{equation}
  \label{def:PSNR} \PSNR~\text{(dB)} = 10  \log_{10}   \biggl\{
  \frac{[(\Psi \bm{s})_{\tMAX}  -  (\Psi \bm{s})_{\tMIN}]^2}{\|\widetilde{\bm{s}}  - \bm{s}\|^2_2 / p} \biggr\}.
\end{equation}

\subsubsection{Large scale with structurally random sampling matrices}
\label{subsubsec:2dexamplesrm}

\begin{figure}
\subfigure[Lena]{\includegraphics[width=0.13\linewidth]{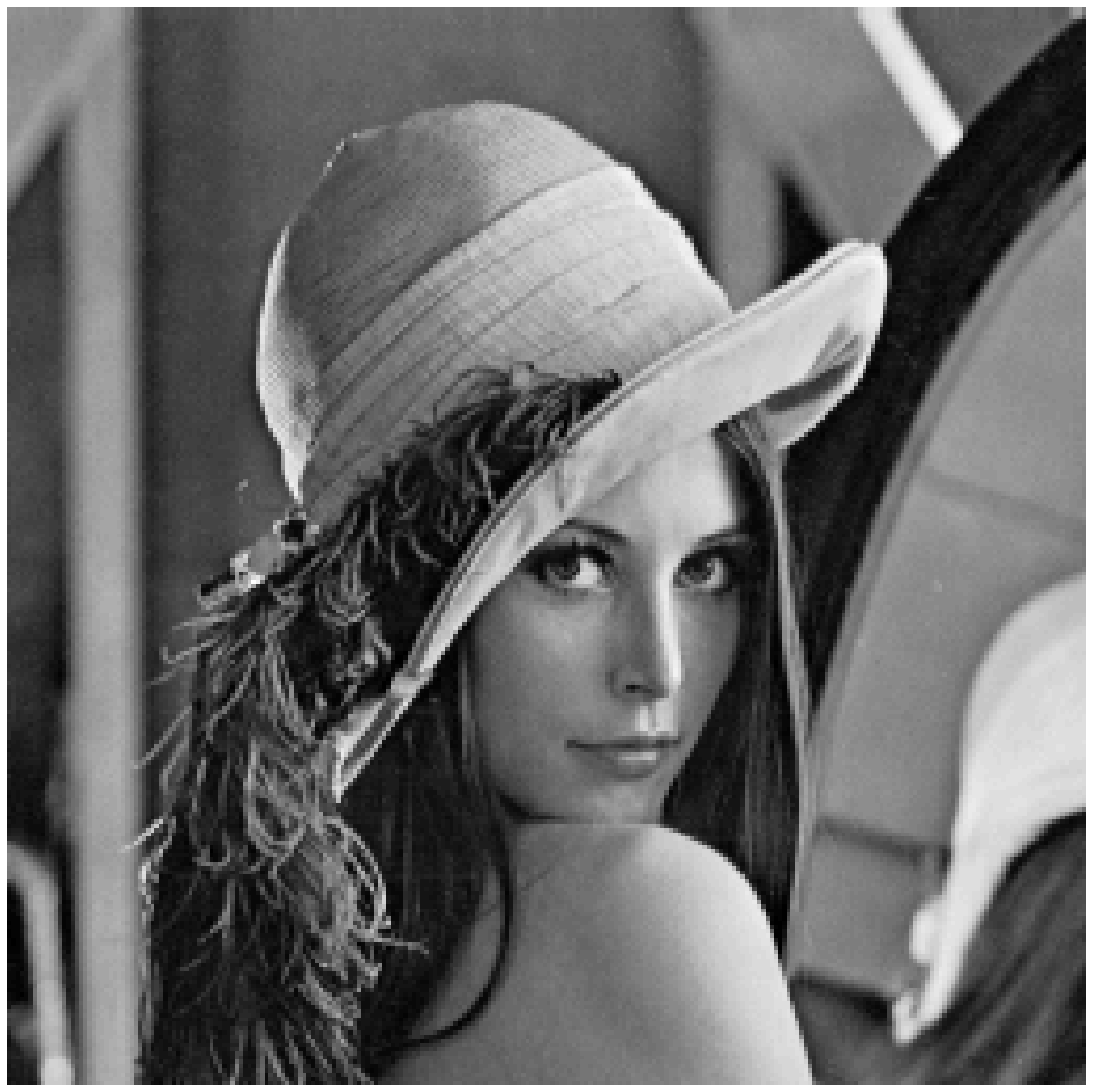}
\label{Img:Lena} 
}
\subfigure[Cameraman]{\includegraphics[width=0.13\linewidth]{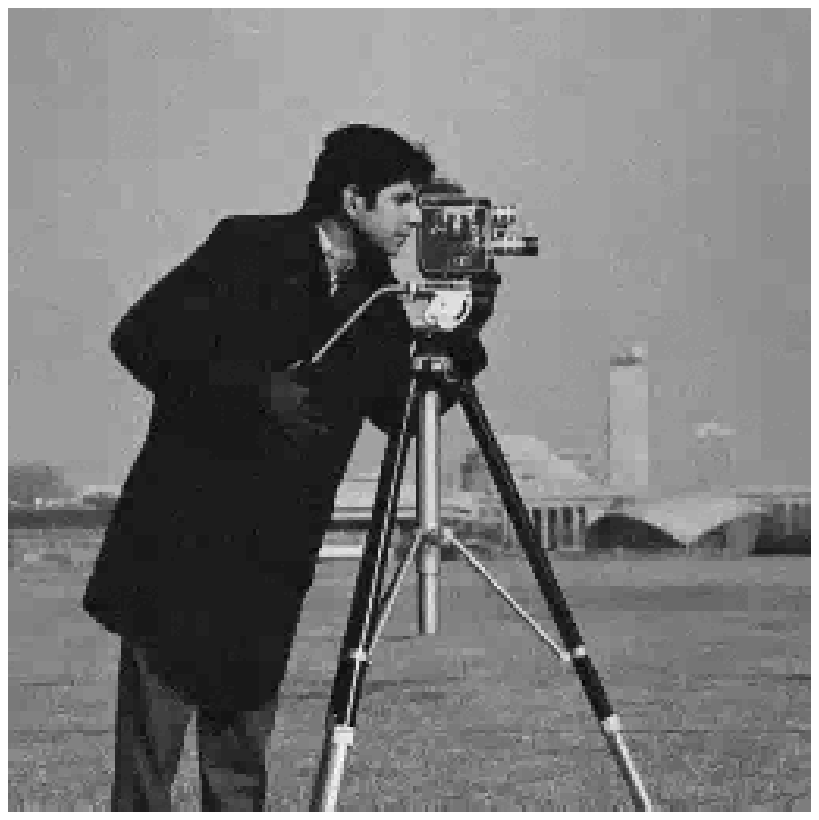}
\label{Img:Cameraman} } 
\subfigure[House]{\includegraphics[width=0.13\linewidth]{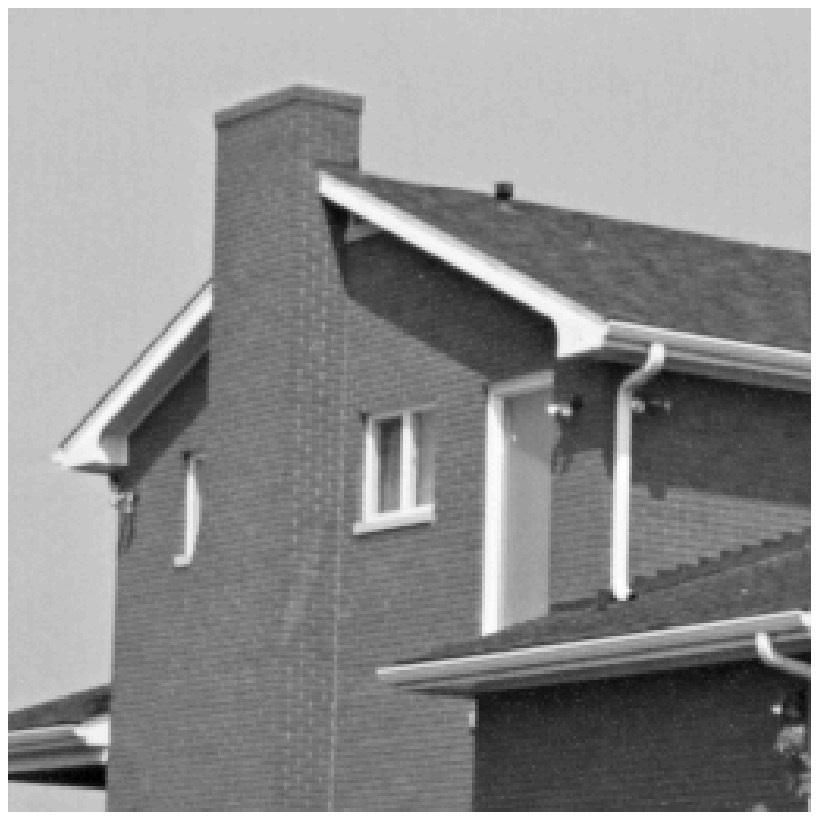}
\label{Img:House} } 
\subfigure[Boat]{\includegraphics[width=0.13\linewidth]{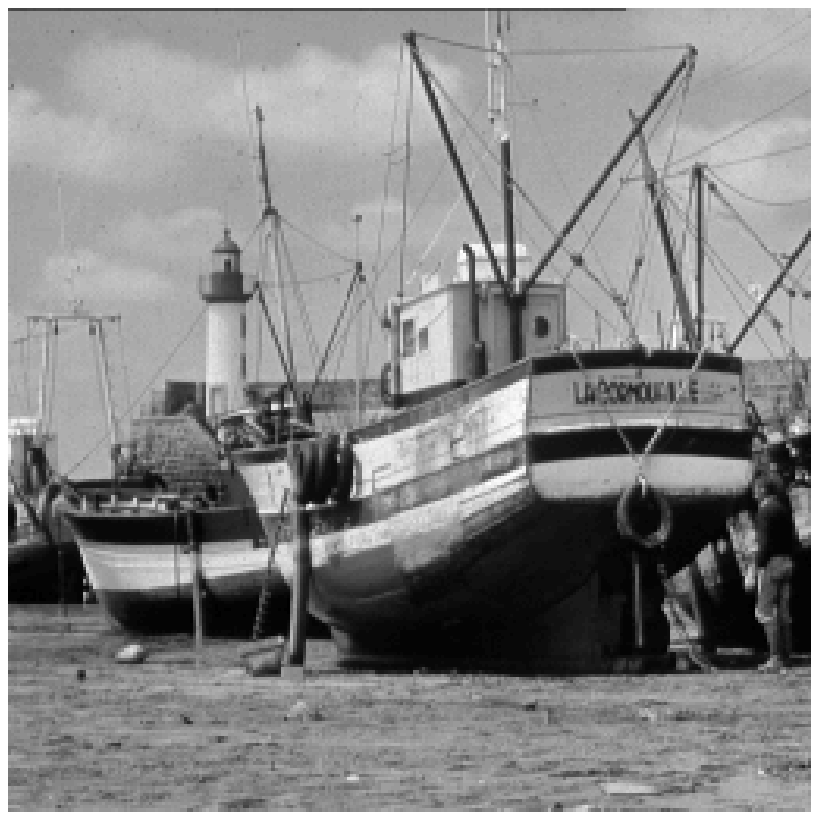}
\label{Img:Boat} } 
\subfigure[Einstein]{\includegraphics[width=0.13\linewidth]{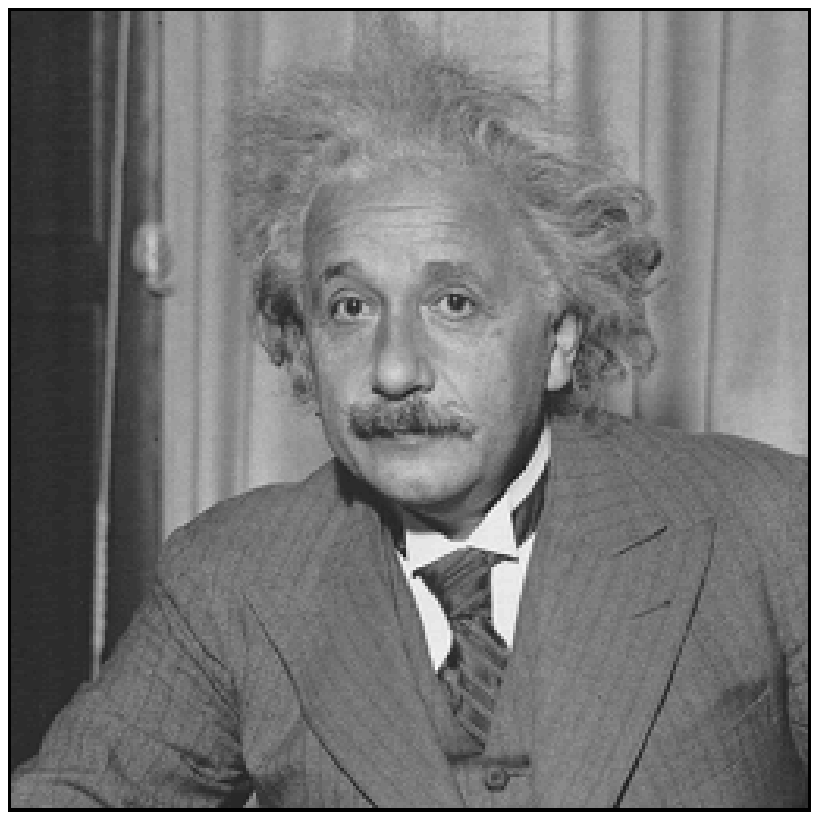}
\label{Img:Einstein} } 
\subfigure[Peppers]{\includegraphics[width=0.13\linewidth]{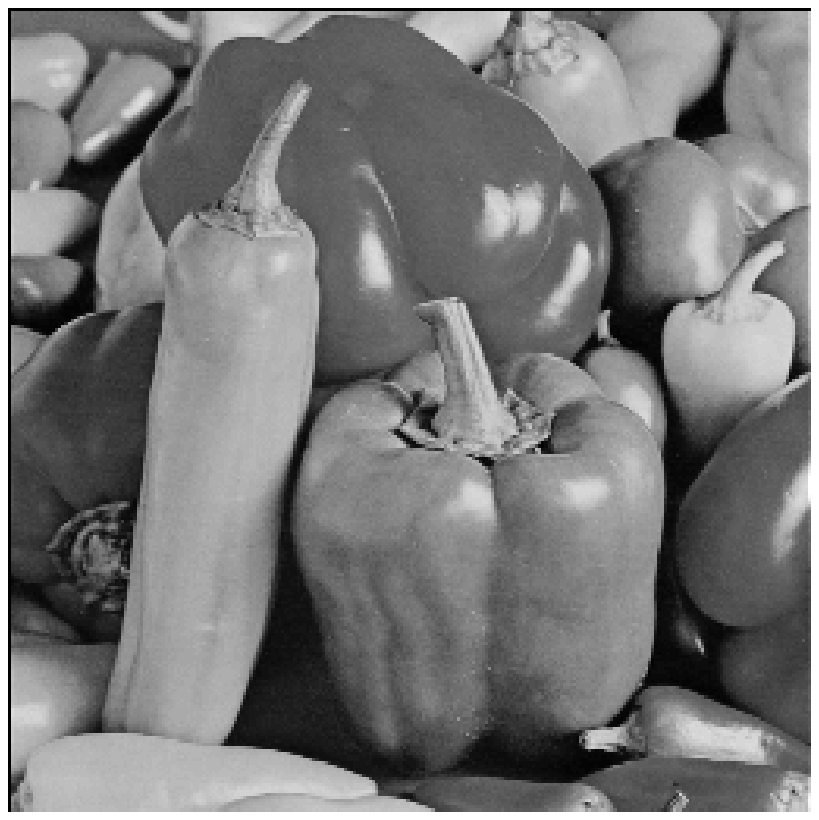}
\label{Img:Peppers} } 
\subfigure[Couple]{\includegraphics[width=0.13\linewidth]{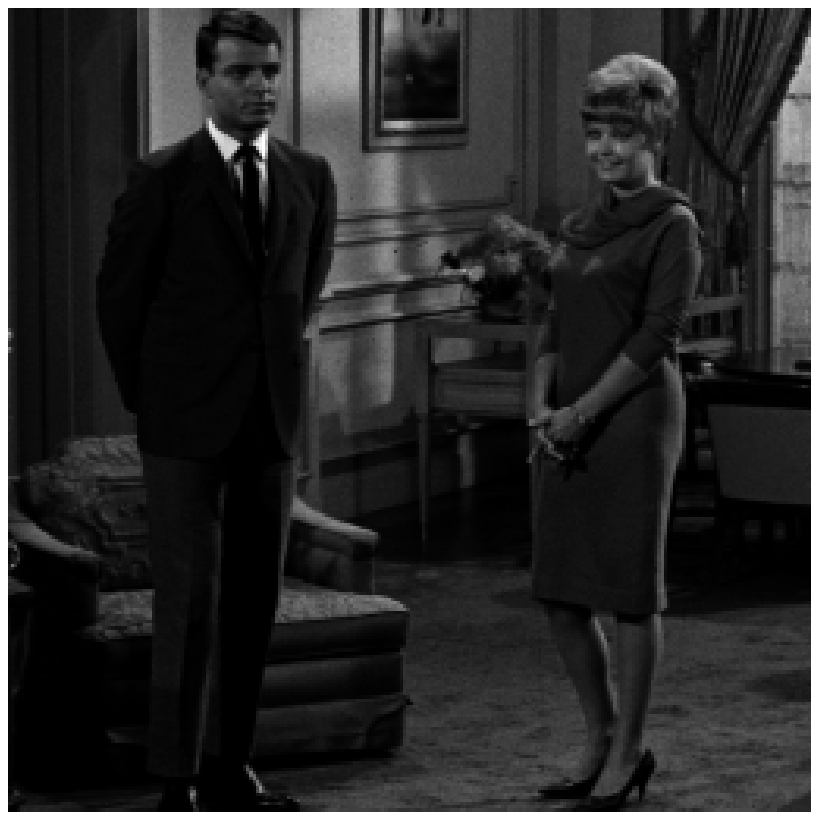}
\label{Img:Couple} }

\caption{The $256 \times 256$ test images.}
\label{fig:test} 
\vspace{-0.2in}
\end{figure}

\noindent We now reconstruct several $256 \times 256$ test images
shown in Fig.~\ref{fig:test} from compressive samples.   The sampling matrix
$\Phi$ is generated using structurally random compressive samples
\cite{DoGanNguyenTran} and the transform matrix $\Psi$ in \eqref{eq:H}
is the $p \times p$ orthogonal inverse Haar wavelet transform matrix,
which implies that the sensing matrix $H$ has orthonormal rows: $H H^T
= I_N$ and, consequently, $\rho_{\Phi} = \rho_H = 1$.  Our performance
metric in this example is the \gls{PSNR}, see \eqref{def:PSNR}.

In this example, the convergence threshold in \eqref{eq:convcrit} is set to
\begin{equation}
  \label{eq:deltaex2}
  \delta = 0.1.
\end{equation}
For \gls{MP-EM} and \gls{MP-EM_OPT}, we set the grid length $K =
12$. The tuning parameters for \gls{MP-EM} are the same as before and
given in \eqref{eq:tuningMP-EM}.\looseness=-1

We set the signal sparsity levels for \gls{NIHT} and \gls{MB-IHT} to
$10000 N / p$ and $15000 N / p$, respectively, tuned for good
\gls{PSNR} performance. For \gls{FPC_AS} and \gls{GPSR}, we set the
regularization parameter $a = -3$ [see \eqref{def:tuningparameter}],
which yields generally the best \gls{PSNR} performance for these two
methods.

We do not include the \gls{VB} method in this example because its implementation
\cite{VBLink} cannot be applied to reconstruct the large-scale images in
Fig.~\ref{fig:test}.

\begin{figure}
\subfigure[]{\includegraphics[width=0.475\linewidth]{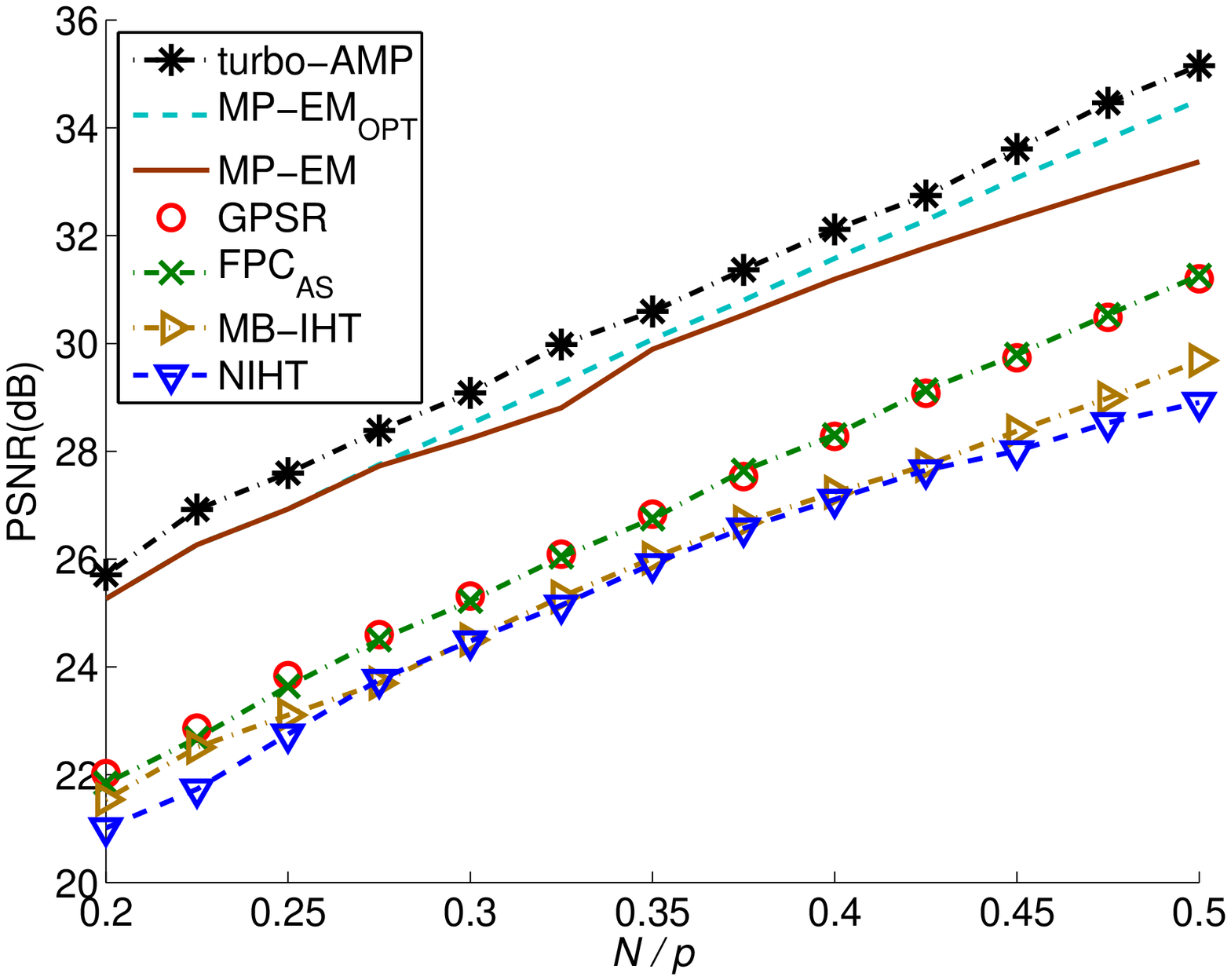}
\label{fig:PSNR256Cameraman4level} } \hfil \centering
\subfigure[]{\includegraphics[width=0.475\linewidth]{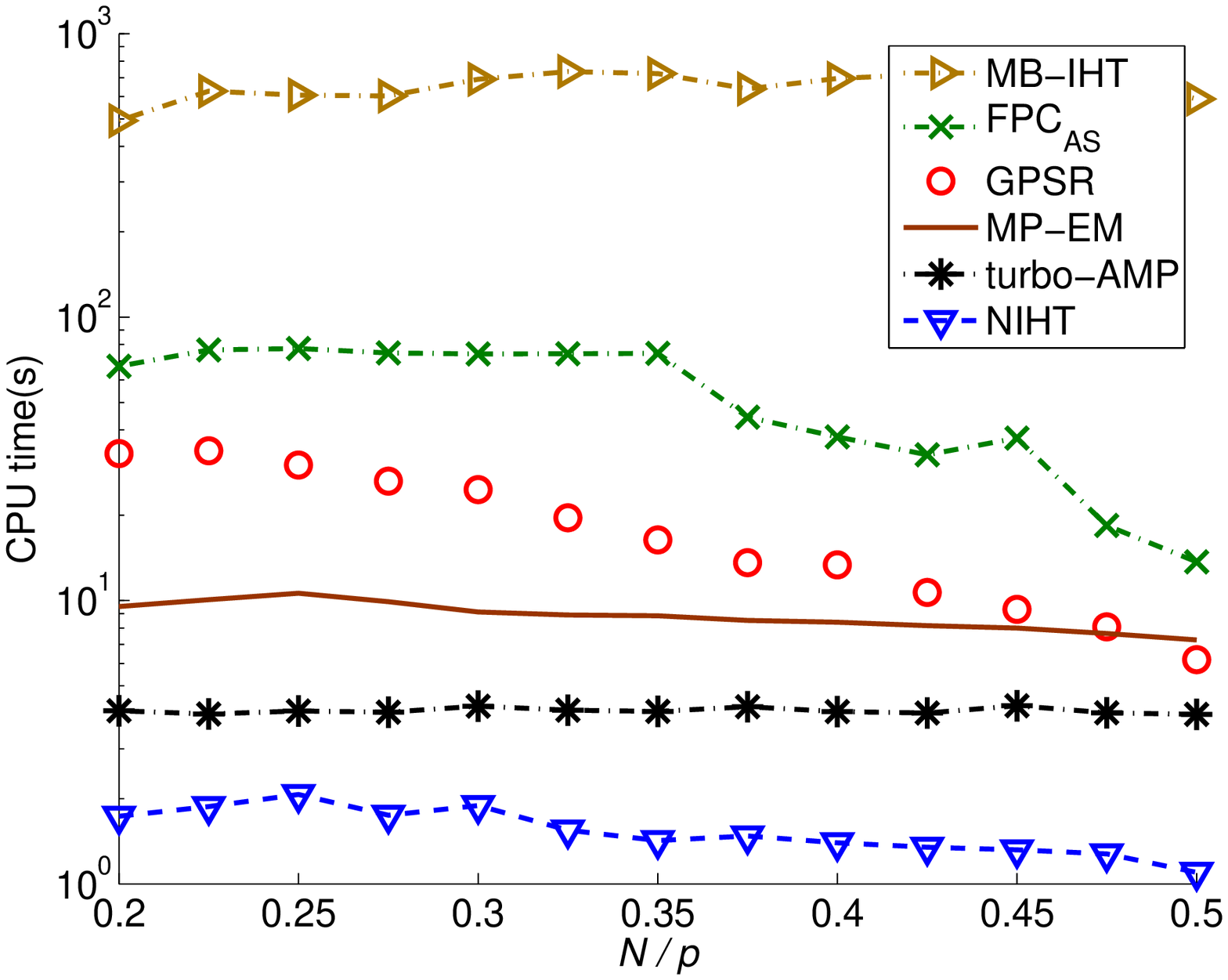}
\label{fig:CPUTime256Cameraman4level} } 
\caption{(a) \glspl{PSNR} and (b) \gls{CPU} times as functions of the
  subsampling factor $N / p$ for the $256 \times 256$ `Cameraman'
  image.}
\label{fig:CameramanSRM}
\end{figure}

Fig.~\ref{fig:CameramanSRM} shows the \glspl{PSNR} and \gls{CPU} times
of different methods reconstructing the $256 \times 256$
`Cameraman' image, as functions of the subsampling factor $N / p$. \Gls{turbo-AMP}
has the highest \glspl{PSNR} for all $N / p$. The performances of \gls{MP-EM}
and \gls{MP-EM_OPT} are close to that of \gls{turbo-AMP}: the \glspl{PSNR} of
\gls{MP-EM_OPT} are \SIrange{0.4}{0.7}{\decibel} less than those of
\gls{turbo-AMP}. Moreover, the \gls{PSNR} improvement for \gls{MP-EM}
against its other closest competitors varies between \SIrange{2.1}{3.2}{\decibel}.    
In terms of \gls{CPU} time, \gls{NIHT} is the fastest among all the
methods compared; \gls{turbo-AMP} is the second fastest and takes
around \SI{4}{\second} for each $N / p$. The \gls{MP-EM} method
requires \SIrange{3.3}{6.5}{\second} more than \gls{turbo-AMP}, but is
clearly faster than \gls{GPSR}, \gls{FPC_AS}, and \gls{MB-IHT} for
nearly all measurement points. As before,
\gls{MP-EM} and \gls{MP-EM_OPT} have the same runtime and we report
only that of \gls{MP-EM} in Fig.~\ref{fig:CPUTime256Cameraman4level}.

Table~\ref{tab:ImgComparison1} shows the
\glspl{PSNR} of the compared methods for different images and $N / p$
equal to $0.35$. 
The \gls{MP-EM}, \gls{MP-EM_OPT}, and
\gls{turbo-AMP} methods clearly outperform the other methods for every
image. In Table~\ref{tab:ImgComparison1}, \gls{turbo-AMP} is better
than \gls{MP-EM_OPT} and \gls{MP-EM} for all the
images: The improvement in terms of \gls{PSNR} varies between
\SI{0.3}{\decibel} and \SI{1.4}{\decibel}.

\begin{table}[t]
\centering
\caption{\glspl{PSNR} for  $N / p = 0.35$.
}
\begin{tabular}{l c c c c c c c}
& \gls{NIHT} & \gls{MB-IHT} & \gls{FPC_AS} & \gls{GPSR} &
\gls{turbo-AMP} & \gls{MP-EM}  & \gls{MP-EM_OPT}  \\
\midrule
Lena  & 24.3 & 24.8 & 25.3 & 25.5 & 
\textbf{29.2} & 27.8  & 27.9\\
Cameraman & 26.0 & 26.0 & 26.8 & 26.8 & \textbf{30.6} &  29.9  & 30.1\\
House & 29.8 & 29.7 & 30.5 & 30.5  & \textbf{33.4} & 32.6  & 33.1\\
Boat & 22.5 & 22.9 & 23.7 & 24.0 & \textbf{27.1} & 26.1  & 26.1\\
Einstein  & 26.9 & 27.4 & 27.4  & 27.7 & \textbf{30.4} & 30.0 & 30.0\\
Peppers & 25.8 & 26.2 & 26.1 & 26.2 & \textbf{30.2} & 29.2  & 29.3\\
Couple & 28.8 & 29.1 & 30.3 & 30.2 & \textbf{33.6} & 32.6  & 32.7
\end{tabular}
\label{tab:ImgComparison1}
\end{table}

In Fig.~\ref{fig:CameramanSRM} and Table~\ref{tab:ImgComparison1}, 
\gls{MB-IHT} achieves a fair performance and
consumes the the largest amount of \gls{CPU} time.  \Gls{turbo-AMP} performs well
for all $N / p$ and images and outperforms all competitors, which is likely
because
\begin{itemize}
\item it uses a more general prior on the binary state variables
(than our \gls{MP-EM} method), which allows the tree probability
parameters $P_{\tH}$, $P_{\tL}$, $\gamma^2$, and $\epsilon^2$ to
vary between the signal decomposition levels, and \item
\emph{learns} the tree probability parameters parameters from
  the measurements.
\end{itemize}
In contrast, our \gls{MP-EM} method employs the crude choices of the
tree and other tuning parameters in
\eqref{eq:tuningMP-EM}.\looseness=-1

\begin{figure}[!h]
\centering
\subfigure[True Image]{\includegraphics[width=1.67in]{Cameraman.eps}
\label{Img:CameramanTrue} } 
\subfigure[\gls{turbo-AMP} ($\gls{PSNR} = 30.59$~dB)]{\includegraphics[width=1.67in]{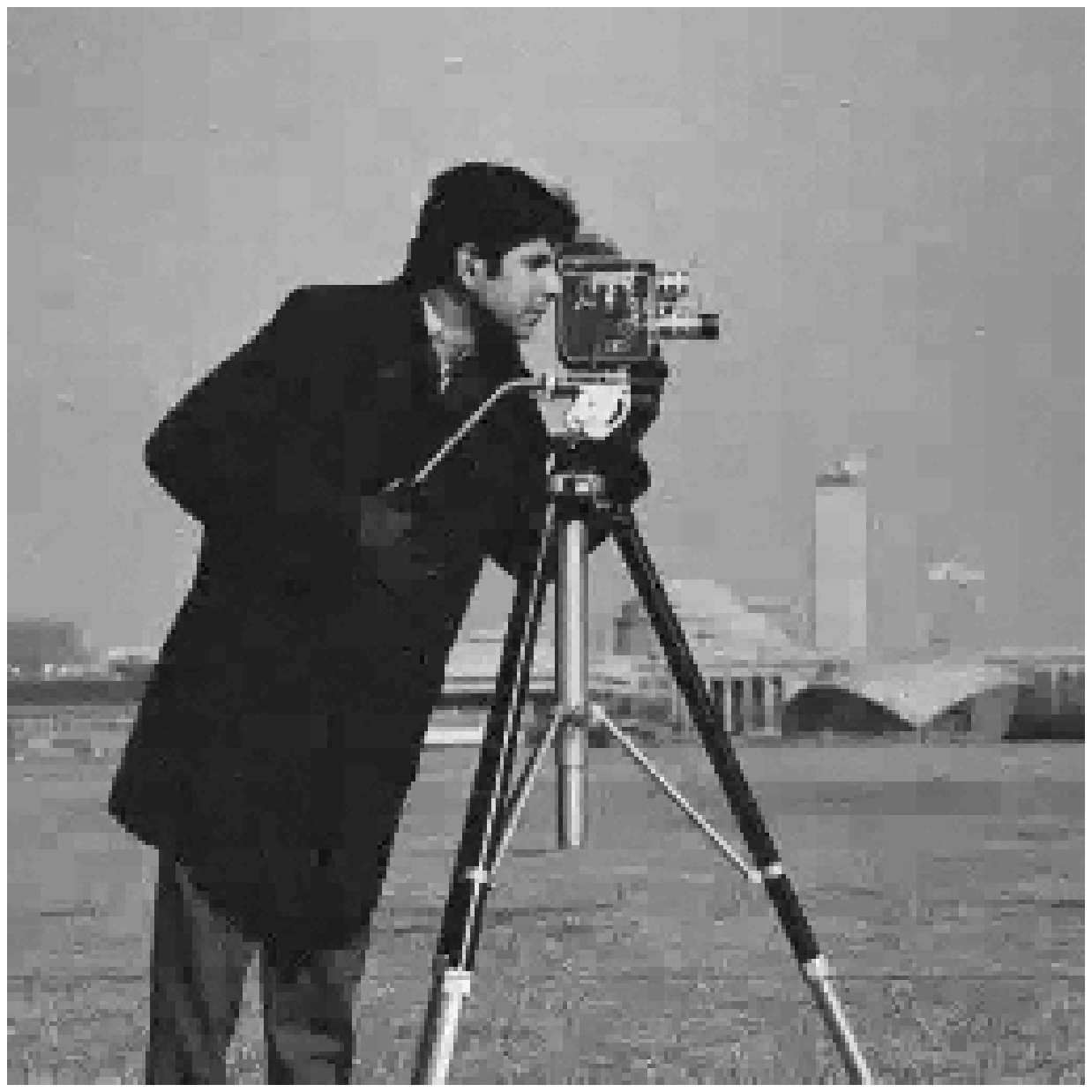} \label{Img:CameramanTurboAMP} } 
\subfigure[\gls{MP-EM_OPT} ($\gls{PSNR} = 30.08$~dB)]{\includegraphics[width=1.67in]{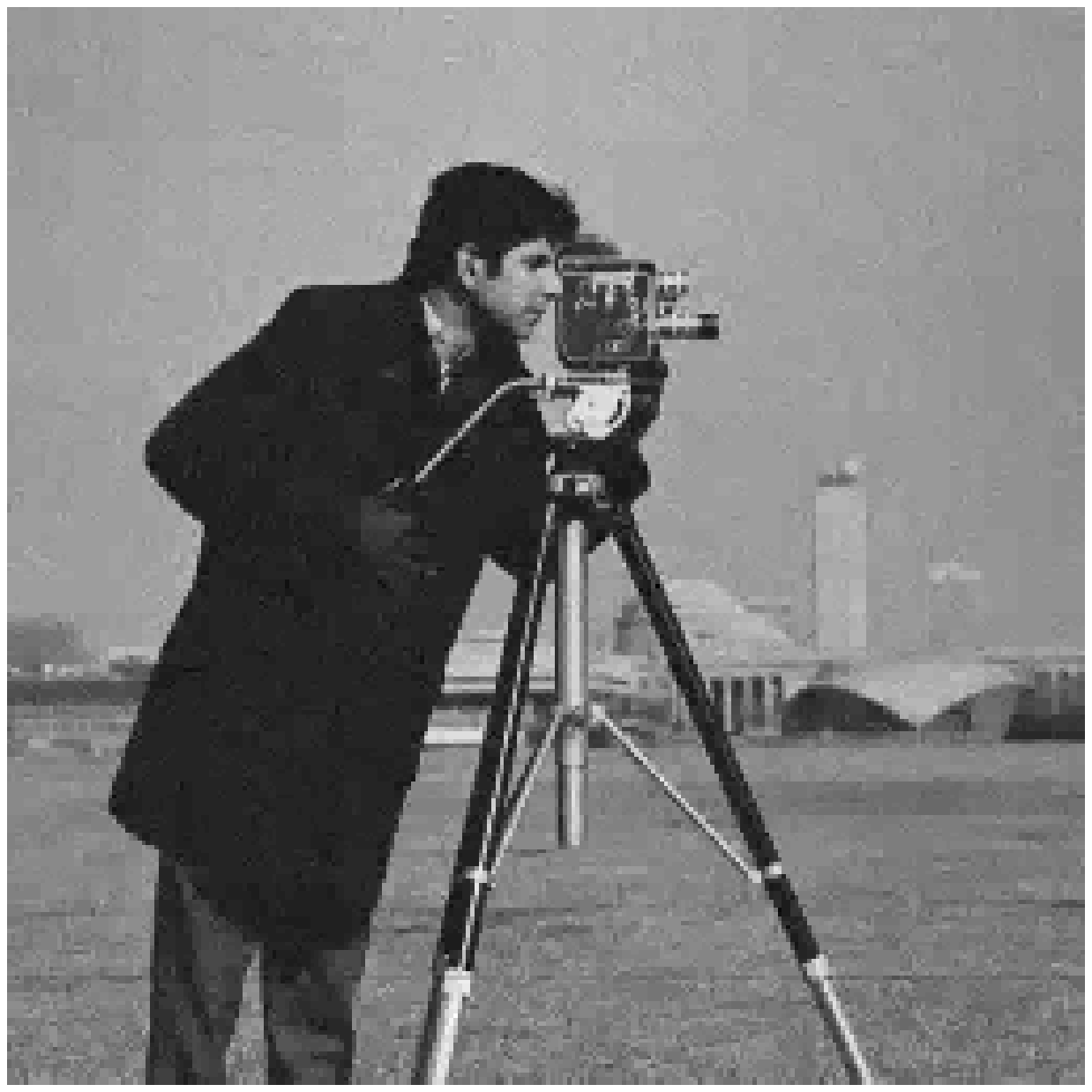} \label{Img:CameramanMPEMOPT}} 
\subfigure[\gls{MP-EM} ($\gls{PSNR} = 29.89$~dB)]{\includegraphics[width=1.67in]{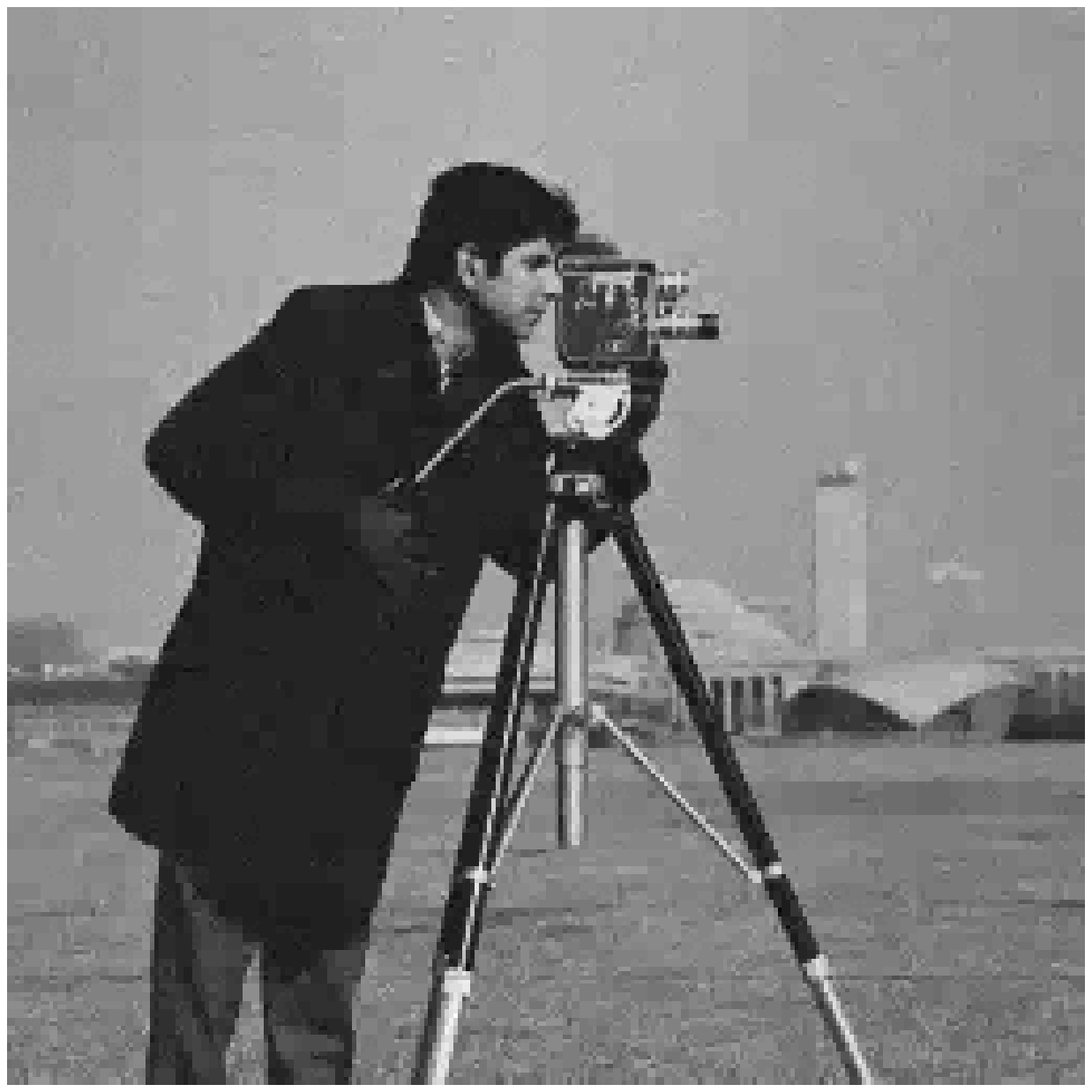} \label{Img:CameramanMPEM}} 
\subfigure[\gls{GPSR} ($\gls{PSNR} = 26.83$~dB)]{\includegraphics[width=1.67in]{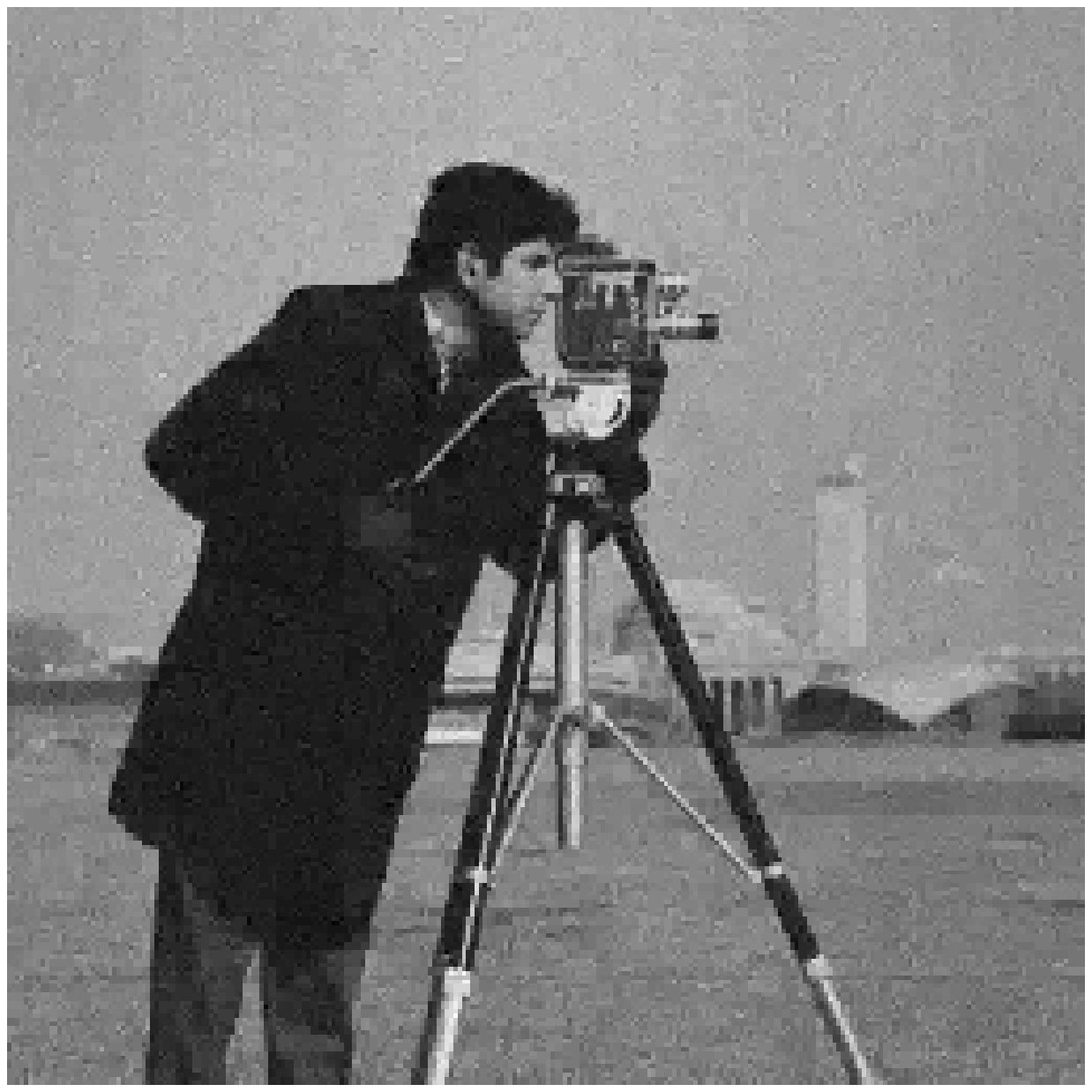}
\label{Img:CameramanGPSR} } 
\subfigure[\gls{FPC_AS} ($\gls{PSNR} = 26.75$~dB)]{\includegraphics[width=1.67in]{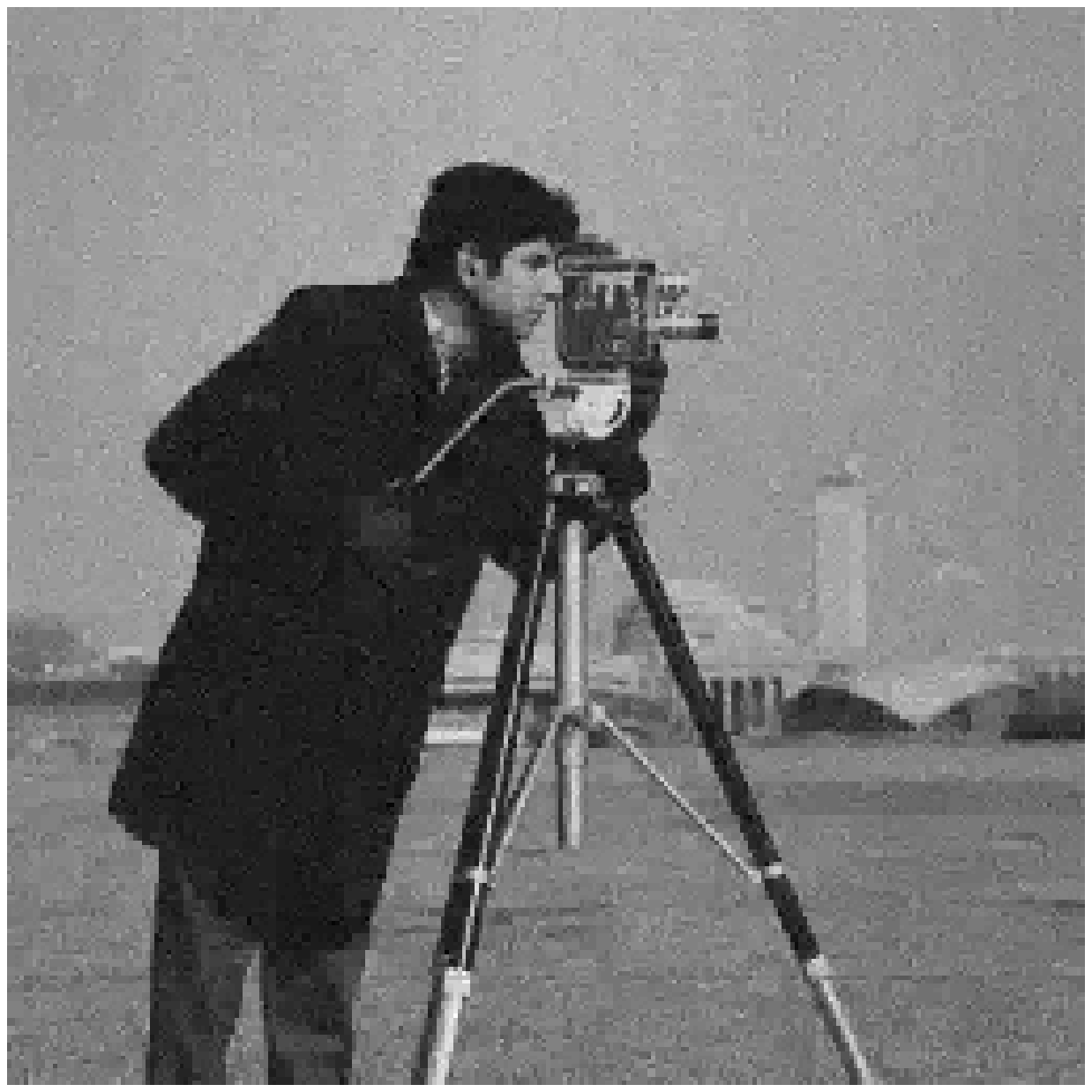}
\label{Img:CameramanFPCAS} } 
\subfigure[\gls{MB-IHT} ($\gls{PSNR} = 26.01$~dB)]{\includegraphics[width=1.67in]{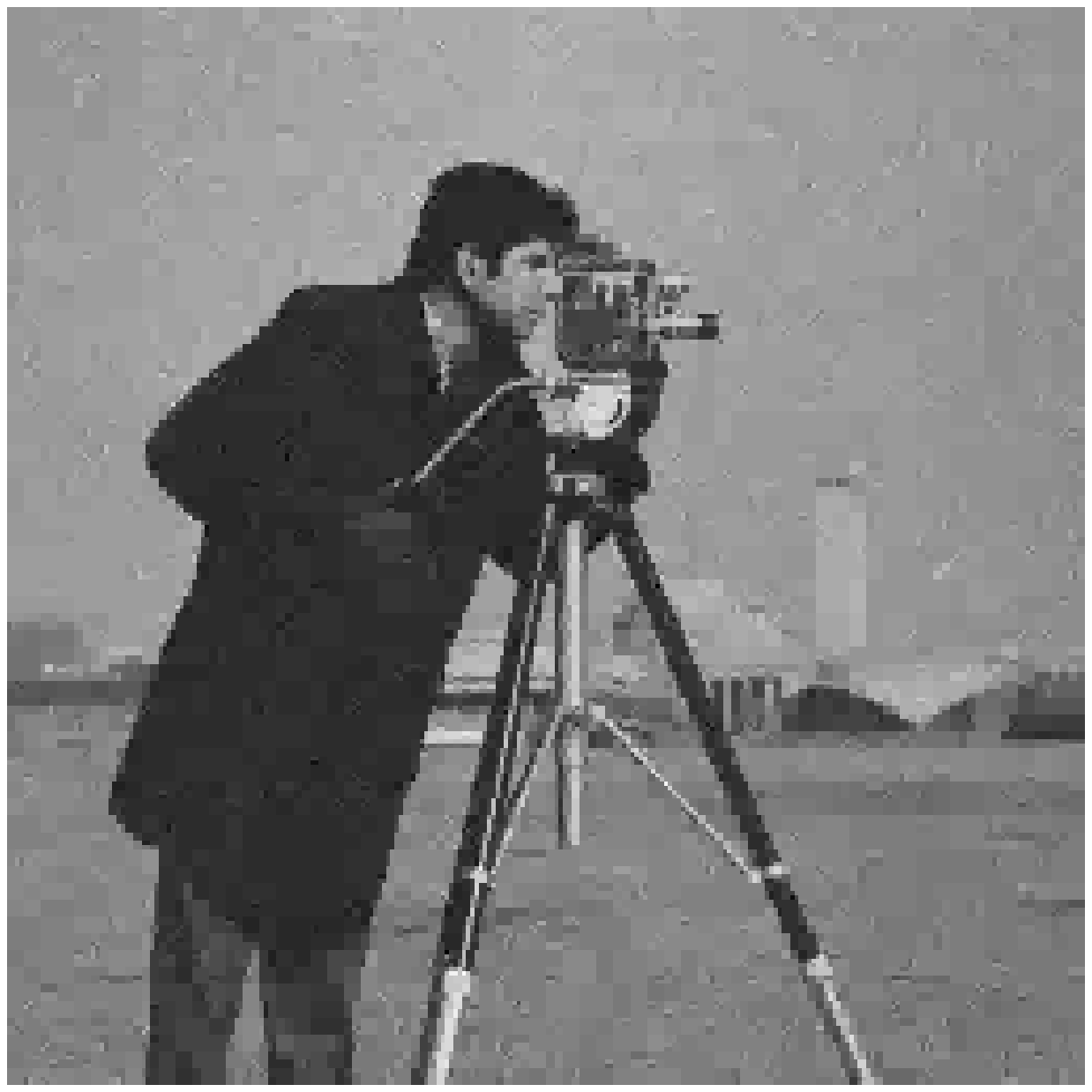}
\label{Img:CameramanMBIHT} } 
\subfigure[\gls{NIHT} ($\gls{PSNR} = 25.91$~dB)]{\includegraphics[width=1.67in]{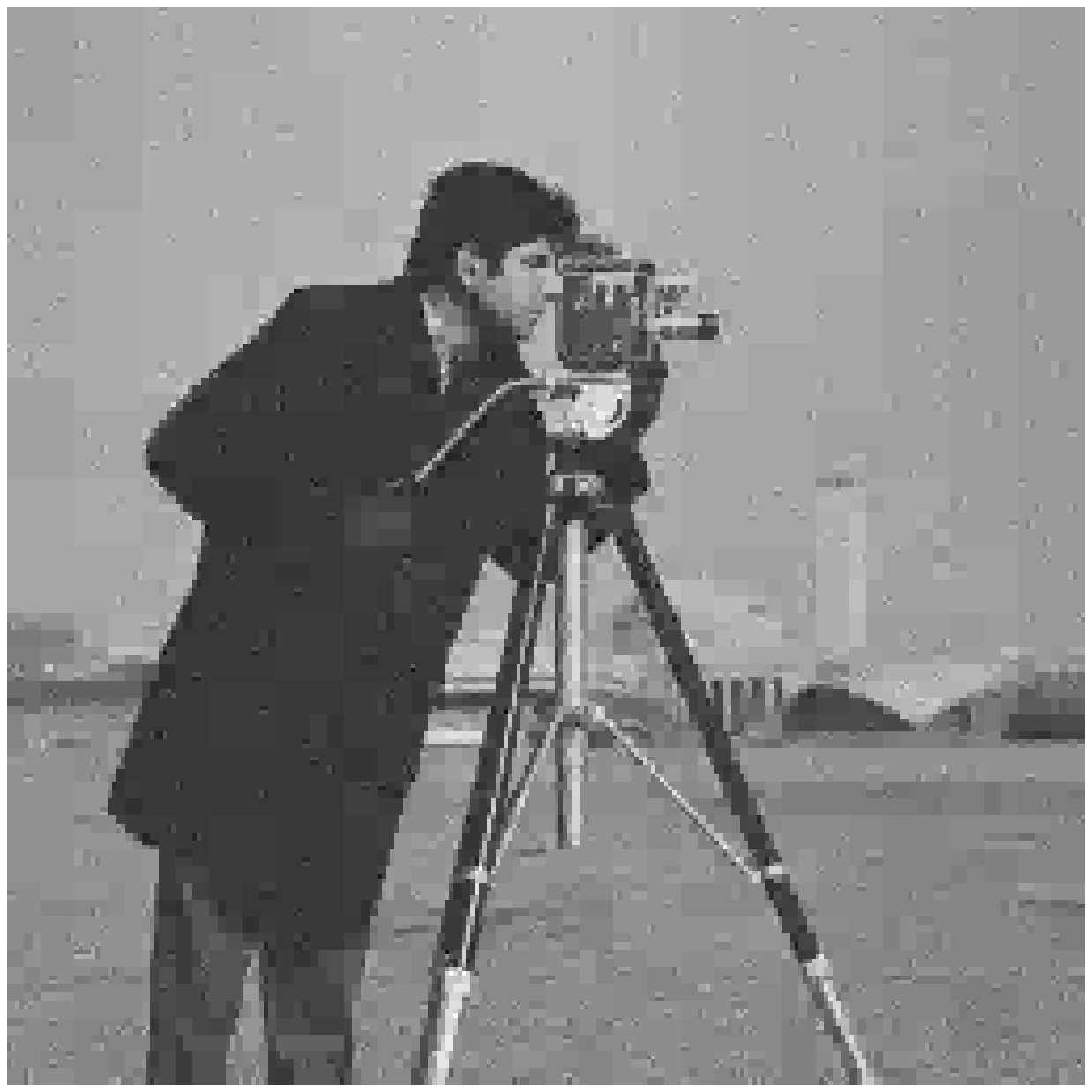}
\label{Img:CameramanNIHT} } 
\caption{The `Cameraman' image reconstructed by various methods for $N/p = 0.35$.}
\label{img:cameraman256} 
\end{figure}

Fig.~\ref{img:cameraman256} shows the reconstructed $256 \times 256$
`Cameraman' image by different methods for $N/p = 0.35$: In this case,
the \gls{turbo-AMP} algorithm achieves the best reconstructed image
quality compared with other methods, followed closely by \gls{MP-EM}
and \gls{MP-EM_OPT}; the reconstructions of all other methods are
clearly inferior to these schemes.

\section{Concluding Remarks}
\label{Conclusion}

\noindent We presented a Bayesian \gls{EM} algorithm for reconstructing
approximately sparse signal from compressive samples using a
Markov tree prior for the signal coefficients. We employed the
max-product belief propagation algorithm to implement the \gls{M} step
of the proposed \gls{EM} iteration. Compared with the existing message
passing algorithms in the compressive sampling area, our method
does not approximate the message form. The simulation results show
that our algorithm often outperforms existing algorithms for
simulated signals and standard test images with different sampling
operators and can successfully reconstruct signals collected by sampling
matrices with correlated elements and variable norms of rows and columns.

Our future work will include the convergence analysis of the \gls{MP-EM}
algorithm, incorporating other measurement models, using a more
general prior distribution for the binary state variables, and designing schemes
for learning the probabilistic Markov tree parameters from the measurements.

\appendices

\section{Derivation of the MP-EM Iteration and Proofs of Its Monotonicity
and Theorem~\ref{theorem}}
\label{app:EMalgorithm}

\renewcommand
    {\theequation}
    {A\arabic{equation}}
\setcounter{equation}{0}

\noindent We first determine the complete-data posterior distribution
and the distribution of the missing data $\bm{z}$ given the observed
data $\bm{y}$ and parameters $\bm{\theta}$ and $\sigma^2$. We then use
these distributions to derive the \gls{EM} iteration in
Section~\ref{sec:BayesianEM} following the standard approach outlined
in, e.g., \cite[Sec. 12.3]{Gelmanetal}. Finally, we prove the
monotonicity of the \gls{MP-EM} iteration in \eqref{eq:monotonicity}
and Theorem~\ref{theorem}.

 Consider the hierarchical two-stage model in
\eqref{eq:hierarchicalmodel}. The complete-data posterior
distribution for known $\sigma^2$ is \looseness=-1
\begin{subequations}
\label{eq:jointposteriorandcondmissinggivenobserveddataandparameter}
\begin{IEEEeqnarray}{rCl}
p_{\bm{\theta}, \bm{z} | \sigma^2, \bm{y}} (\bm{\theta}, \bz | \sigma^2, \bm{y}) 
&\propto& p_{\bm{y} | \bm{z}, \sigma^2} (\bm{y} | \bm{z}, \sigma^2) 
p_{\bm{z} | \bm{s}} (\bm{z} | \bm{s})  p_{\bm{s} | \bm{q}, \sigma^2} (\bm{s} |
\bm{q}, \sigma^2)  p_{\bm{q}} (\bm{q})  \notag\\
&\propto& \frac{\exp \{- 0.5 (\bm{y} - H \bz)^T [C (\sigma^2)]^{-
1} (\bm{y} - H \bz)\}}{\sqrt{\det [C (\sigma^2)]}} 
\Bigl(\frac{\epsilon^2}{\gamma^2}\Bigr)^{0.5  \sum_{i=1}^p q_i}  p_{\bm{q}}(\bq)
  \notag\\
  & & \cdot \exp[ - 0.5 
\| \bz - \bm{s} \|_2^2 / \sigma^2 - 0.5  \bm{s}^T  D^{-1}(\bq) 
  \bm{s} / \sigma^2 ]
    \label{eq:jointposterior}
\end{IEEEeqnarray}
where
\begin{equation}
\label{eq:C}
C (\sigma^2) = \sigma^2 (I_N - H H^T).
\end{equation}
Consequently, the distribution of the missing data $\bm{z}$ given the
observed data $\bm{y}$ and parameters $\bm{\theta}$ and $\sigma^2$ is
\begin{IEEEeqnarray}{rCl}
  \label{eq:condmissinggivenobserveddataandparameters}
p_{\bm{z} | \sigma^2, \bm{y}, \bm{\theta}} (\bz | \sigma^2, \bm{y}, \bm{\theta}) =
p_{\bm{z} | \sigma^2, \bm{y}, \bm{s}} (\bz | \sigma^2, \bm{y}, \bm{s}) = \mathcal{N}
\bigl(\bz | \Exp_{\bm{z} | \sigma^2, \bm{y}, \bm{s}} (\bz | \sigma^2, \bm{y},
\bm{s}), \cov_{\bm{z} | \sigma^2, \bm{y}, \bm{s}} (\bz | \sigma^2,
\bm{y}, \bm{s})\bigr) \notag\\
\end{IEEEeqnarray}
where
\begin{IEEEeqnarray}{rCl}
\label{eq:condmissinggivenobserveddataandparameters2}
\Exp_{\bm{z} | \sigma^2, \bm{y}, \bm{s}} (\bz | \sigma^2, \bm{y}, \bm{s}) &=&
\{H^T [C (\sigma^2)]^{- 1} H + I_p / \sigma^2\}^{- 1} \{H^T [C
(\sigma^2)]^{- 1} \bm{y} + \bm{s} / \sigma^2\} \\
\label{eq:condmissinggivenobserveddataandparameters3}
\cov_{\bm{z} | \sigma^2, \bm{y}, \bm{s}} (\bz | \sigma^2, \bm{y},
\bm{s}) &=& \{H^T [C (\sigma^2)]^{- 1} H + I_p / \sigma^2\}^{-1}.
\end{IEEEeqnarray}
\end{subequations}

By using the matrix inversion lemma \cite[eq.~(2.22),
p.~424]{Harville}:
\begin{subequations}
\begin{eqnarray}
(R + S T U)^{- 1} &=& R^{- 1} - R^{- 1} S (T^{- 1} + U R^{- 1}
S)^{ - 1} U R^{- 1}
\end{eqnarray}
and the following identity \cite[p.~425]{Harville}:
\begin{eqnarray}
(R + S T U)^{- 1} S T &=& R^{- 1} S (T^{- 1} + U R^{- 1} S)^{ - 1}
\end{eqnarray}
\end{subequations}
we simplify the conditional mean of the missing data in 
\eqref{eq:condmissinggivenobserveddataandparameters2} to the familiar
backprojection form:
\begin{equation}
\Exp_{\bm{z} | \sigma^2, \bm{y}, \bm{s}} [\bz | \sigma^2, \bm{y}, \bm{s}] =
\bm{s} + H^T (\bm{y} - H \bm{s}) \label{eq:conditionalexpectation}.
\end{equation}

We now derive the \gls{EM} iteration in Section~\ref{sec:BayesianEM}
by noting that the objective function $\ln p_{\bm{\theta} | \sigma^2,
  \bm{y}}( \bm{\theta} | \sigma^2, \bm{y} )$ that we aim to maximize
satisfies the following property [see e.g.,
\cite[eq. (12.4)]{Gelmanetal}]:
\begin{subequations}
\begin{IEEEeqnarray}{rCl}
  \ln p_{\bm{\theta} | \sigma^2, \bm{y}}( \bm{\theta} | \sigma^2,
  \bm{y} )  = \cQ (\bm{\theta} | \bm{\theta}^{(j)}) - \cH (\bm{\theta} |
  \bm{\theta}^{(j)})
  \label{eq:EMdecomposition}
\end{IEEEeqnarray}
where 
\begin{IEEEeqnarray}{rCl}
  \cQ (\bm{\theta} | \bm{\theta}^{(j)}) &\triangleq& \Exp_{\bm{z} | \sigma^2,
    \bm{y}, \bm{\theta}}  \bigl[ \ln p_{\bm{\theta}, \bm{z} | \sigma^2, \bm{y}}
  (\bm{\theta}, \bz | \sigma^2, \bm{y}) | \sigma^2, \bm{y}, \bm{\theta}^{(j)} \bigr]
  \\
  \cH (\bm{\theta} | \bm{\theta}^{(j)}) &\triangleq& \Exp_{\bm{z} | \sigma^2,
    \bm{y}, \bm{\theta}}  \bigl[ \ln p_{\bm{z} | \sigma^2, \bm{y}, \bm{\theta}}
  (\bz | \sigma^2, \bm{y}, \bm{\theta}) | \sigma^2, \bm{y}, \bm{\theta}^{(j)} \bigr]
\end{IEEEeqnarray}
\end{subequations}
are the expected complete-data log-posterior distribution and negative
entropy of the conditional missing data \gls{pdf}.  The expected
complete-data log-posterior $\cQ (\bm{\theta} | \bm{\theta}^{(j)})$
follows easily by taking the logarithm of the complete-data posterior
distribution \eqref{eq:jointposterior}, ignoring constant terms (not
functions of $\bm{\theta}$), and computing the conditional
expectation with respect to the missing data given the observed data
and parameters from the $j$th iteration:
\begin{subequations}
\begin{IEEEeqnarray}{rCl}
\label{eq:Q}
  \cQ (\bm{\theta} | \bm{\theta}^{(j)}) &=& \text{const} + \Exp_{\bm{z} | \sigma^2, \bm{y},
    \bm{\theta}}  \biggl\{   - 0.5  \frac{\| \bz - \bm{s} \|_2^2 + \bm{s}^T 
    D^{-1}(\bq)  \bm{s}}{\sigma^2}
+
  \ln[p_{\bm{q}} (\bm{q})] + 0.5 \ln\Bigl(\frac{\epsilon^2}{\gamma^2}\Bigr) \sum_{i=1}^p
  q_i \,\bigg|\, \sigma^2, \bm{y}, \bm{\theta}^{(j)} \biggr\}
  \notag\\
  &=& \text{const} - 0.5  \frac{\| \bz^{(j)} - \bm{s} \|_2^2 + \bm{s}^T 
    D^{-1}(\bq)  \bm{s}}{\sigma^2} + \ln[p_{\bm{q}}(\bq)] + 0.5 
  \ln\Bigl(\frac{\epsilon^2}{\gamma^2}\Bigr)  \sum_{i=1}^p q_i
  \label{eq:qfunctiondetail}
\end{IEEEeqnarray}
where const denotes the terms that are not functions of $\bm{\theta}$
and $\bz^{(j)}$ is the conditional mean of the missing data in
\eqref{eq:EstepBayes} that follows from
\eqref{eq:conditionalexpectation}. 
To determine the conditional
expectation in  \eqref{eq:qfunctiondetail}, we
only need the conditional mean of the missing data in
\eqref{eq:EstepBayes}, which therefore constitutes the \gls{E} step.
Now, the \gls{M} step requires maximization of $\cQ (\bm{\theta} |
\bm{\theta}^{(j)})$ with respect to $\bm{\theta}$:
\begin{equation}
  \label{eq:Mstepdef}
\bm{\theta}^{(j+1)} =  \arg \max_{\bm{\theta}} \cQ(\bm{\theta} | \bm{\theta}^{(j)})
\end{equation}
and \eqref{eq:MstepBayes1} follows from \eqref{eq:qfunctiondetail}.
\end{subequations}

The monotonicity of the \gls{MP-EM} iteration in \eqref{eq:monotonicity} follows from
\begin{IEEEeqnarray}{rCl}
\ln p_{\bm{\theta} | \sigma^2, \bm{y}}(\bm{\theta}^{(j+1)} | \sigma^2,
  \bm{y}) - \ln p_{\bm{\theta} | \sigma^2, \bm{y}}(\bm{\theta}^{(j)} |
  \sigma^2, \bm{y}) &=& 
    \cH( \bm{\theta}^{(j)} | \bm{\theta}^{(j)} ) - \cH(
    \bm{\theta}^{(j+1)} | \bm{\theta}^{(j)})
\nonumber
\\ & &
+  \cQ( \bm{\theta}^{(j+1)} | \bm{\theta}^{(j)} ) - \cQ(
      \bm{\theta}^{(j)} | \bm{\theta}^{(j)} )
\IEEEyessubnumber
\label{eq:marginalloglikelihooddifference0}
\\
&=& 
 \Div{ p_{\bm{z} | \sigma^2, \bm{y}, \bm{\theta}}( \bm{z} |
  \sigma^2, \bm{y}, \bm{\theta}^{(j)} )   }{   
p_{\bm{z} | \sigma^2, \bm{y}, \bm{\theta}}( \bm{z} | \sigma^2, \bm{y}, \bm{\theta}^{(j+1)} )
  } 
\nonumber
\\ & &
+  \cQ( \bm{\theta}^{(j+1)} | \bm{\theta}^{(j)} ) - \cQ(  \bm{\theta}^{(j)} | \bm{\theta}^{(j)} )
  \geq 0
\IEEEyessubnumber
 \label{eq:marginalloglikelihooddifference}
\end{IEEEeqnarray}
by the nonnegativity of \gls{KL} divergence \cite[Theorem
2.8.1]{Murphy}, \cite[Theorem 8.6.1]{CoverThomas} and the fact that
$\cQ(\bm{\theta}^{(j + 1)} | \bm{\theta}^{(j)}) - \cQ
(\bm{\theta}^{(j)} | \bm{\theta}^{(j)}) \geq 0$ because $\cQ
(\bm{\theta} | \bm{\theta}^{(j)})$ is maximized at $\bm{\theta}^{(j +
  1)}$. Here, \eqref{eq:marginalloglikelihooddifference} follows from
\eqref{eq:marginalloglikelihooddifference0} by using the identity
$\cH( \bm{\theta} | \bm{\theta} ) - \cH( \bm{\theta}' | \bm{\theta}) =
\Div{ p_{\bm{z} | \sigma^2, \bm{y}, \bm{\theta}}( \bm{z} | \sigma^2,
  \bm{y}, \bm{\theta} ) }{ p_{\bm{z} | \sigma^2, \bm{y}, \bm{\theta}}(
  \bm{z} | \sigma^2, \bm{y}, \bm{\theta}' ) }$.

\begin{IEEEproof}[Proof of Theorem~\ref{theorem}]
  For a given $\bq$, \eqref{eq:qfunctiondetail} is a quadratic
  function of $\bm{s}$ that is easy to maximize with respect to
  $\bm{s}$ [see also \eqref{eq:theta}]:
  \begin{equation}
    \arg \max_{\bm{s}} \cQ(\bm{\theta} | \bm{\theta}^{(j)}) = \big[ D^{-1}(\bq) + I_p \big]^{- 1}  \bz^{(j)}.
  \end{equation}
  Therefore, the estimates of $\bm{s}$ and $\bq$ obtained upon
  convergence of the \gls{EM} iteration in Section~\ref{sec:BayesianEM} to
  its fixed point satisfy:
  \begin{IEEEeqnarray}{rCl}
   \bm{s}^{(+\infty)} &=& \big[ D^{-1}(\bq^{(+\infty)}) + I_p \big]^{- 1}  \bz^{(+ \infty)}
   \notag\\
   &=& \big[ D^{-1}(\bq^{(+\infty)}) + I_p \big]^{- 1}  \big[ \bm{s}^{(+\infty)} +  H^T   (\bm{y} -  H  \bm{s}^{(+\infty)}) \big]
   \label{eq:Mstepsinfty}
   \end{IEEEeqnarray}
where the second equality follows by using \eqref{eq:EstepBayes}.
Solving \eqref{eq:Mstepsinfty} for $\bm{s}^{(+\infty)}$ yields
\begin{IEEEeqnarray}{rCl}
   \bm{s}^{(+\infty)} &=& \big[ D^{-1}(\bq^{(+\infty)}) + H^T H \big]^{-1} H^T \bm{y}
   \label{eq:Msteps1}
\end{IEEEeqnarray}
and \eqref{eq:sinftyandshat} follows.
\end{IEEEproof}

\renewcommand
    {\theequation}
    {B\arabic{equation}}

\setcounter{equation}{0}

\renewcommand{\thesubsection}{\Alph{section}-\Roman{subsection}}
\renewcommand{\thesubsubsection}{\Alph{section}-\Roman{subsection}
\arabic{subsubsection}}
\renewcommand {\thesubsectiondis}{\Roman{subsection}}

\section{Derivation of the Messages and Beliefs in Section~\ref{sec:BP}}
\label{app:max-productalgorithmderivation}

\noindent
Before we proceed, note the following useful identities:
\begin{subequations}
\begin{IEEEeqnarray}{rCl}
  \label{eq:appmodeofproductofGaussians}
  \arg \max_{s_i} \mathcal{N}( z_i | s_i, \sigma^2 ) 
  \mathcal{N}( s_i | 0, \tau^2 ) &=& \frac{\tau^2  z_i}{\sigma^2
    + \tau^2}
  \\
\label{eq:appmaximizedproductofGaussians}
  \max_{s_i} \mathcal{N}( z_i | s_i, \sigma^2 )  \mathcal{N}(
  s_i | 0, \tau^2 ) &=& \frac{1}{\sqrt{2  \uppi  \sigma^2 } 
    \sqrt{2  \uppi  \tau^2 }}  \exp\biggl(-0.5  \frac{z_i^2}{\sigma^2 +
  \tau^2}\biggr).
\end{IEEEeqnarray}
\end{subequations}

\subsection{Upward Messages}
\label{app:Upwardmessagederivation}

\subsubsection{Upward Messages from Leaf Nodes}
\label{app:Upwardmessagesfromleafnodes}

\noindent
When passing upward messages from the leaf nodes $i \in \cT_{\sleaf}$, we set
the multiplicative term $\prod_{k \in \ch (i)} m_{k \rightarrow i}
(q_i)$ to one, yielding [see \eqref{eq:upwardmsg}]
\begin{IEEEeqnarray}{rCl}
  \label{eq:upwardmsgleaf}
m_{i \rightarrow \pi(i)}( q_{\pi (i)} ) &=& \alpha  \max_{\bm{\theta}_i}
  \Bigl\{ \mathcal{N}( z_i | s_i, \sigma^2 ) 
  [\mathcal{N}( s_i | 0, \gamma^2 \sigma^2)]^{q_i}  [\mathcal{N}( s_i
| 0, \epsilon^2 \sigma^2)]^{1-q_i}
\notag
\\
&& \cdot [P_{\tH}^{q_i}  (1-P_{\tH})^{1-q_i}]^{q_{\pi(i)}} 
   [P_{\tL}^{q_i}  (1 - P_{\tL})^{1-q_i}]^{1-q_{\pi(i)}}
 \Bigr\}.
\end{IEEEeqnarray}
For $q_{\pi(i)}=0$, we have
\begin{subequations}
\begin{equation}
  \label{eq:muiu0}
  m_{i \rightarrow \pi(i)}(0) = \mu_i^{\tu}(0) = \alpha_1  \max\biggl\{ (1-P_{\tL})  \exp\Bigl( - 0.5 
  \frac{z_i^2}{ \sigma^2+\sigma^2 \epsilon^2}\Bigr) \big/ \epsilon, 
  P_{\tL}  \exp\Bigl( - 0.5  \frac{z_i^2}{ \sigma^2+\sigma^2
    \gamma^2}\Bigr) \big/ \gamma \biggr\}
\end{equation}
and, for $q_{\pi(i)}=1$, we have
\begin{equation}
  \label{eq:muiu1}
  m_{i \rightarrow \pi(i)}(1) = \mu_i^{\tu}(1) = \alpha_1  \max\biggl\{ (1-P_{\tH})  \exp\Bigl(- 0.5 
  \frac{z_i^2}{ \sigma^2+\sigma^2 \epsilon^2}\Bigr) \big/ \epsilon, 
  P_{\tH}  \exp\Bigl(- 0.5  \frac{z_i^2}{ \sigma^2+\sigma^2
    \gamma^2}\Bigr) \big/ \gamma \biggr\}
 \end{equation}
\end{subequations}
where we have used \eqref{eq:appmaximizedproductofGaussians} with
$\tau^2=\sigma^2 \epsilon^2$ and $\tau^2=\sigma^2 \gamma^2$ and
$\alpha_1 > 0$ is an appropriate normalizing constant. It follows from
\eqref{eq:appmodeofproductofGaussians} that the only two
candidates of $\bm{\theta}_i$ to maximize
(\ref{eq:upwardmsgleaf}) are $[0, \widehat{s}_i(0)]^T$ and $[1,
\widehat{s}_i(1)]^T$.

\subsubsection{Upward Messages from Non-Leaf Nodes}

\noindent
For $i \in \cT \backslash \cT_{\sleaf}$, we can use induction to
simplify the multiplicative term $\prod_{k \in \ch (i)} m_{k
  \rightarrow i} (q_i)$ in \eqref{eq:upwardmsg} as follows:
\begin{IEEEeqnarray}{rCl}
\prod_{k \in \ch (i)} m_{k \rightarrow i} (q_i) =
\biggl[\prod_{k \in \ch (i)} \mu_k^{\tu} (0)\biggr]^{1 - q_i}  \biggl[\prod_{k
\in \ch (i)} \mu_k^{\tu} (1)\biggr]^{q_i} \label{eq:prodmsg}
\end{IEEEeqnarray}
see also Fig.~\ref{fig:msgup}.

Substituting \eqref{eq:prodmsg} into \eqref{eq:upwardmsg} yields
\begin{IEEEeqnarray}{rCl}
  m_{i \rightarrow \pi(i)}( q_{\pi (i)} ) &=&  \alpha  \max_{\bm{\theta}_i}
  \Biggl\{ \mathcal{N}( z_i | s_i, \sigma^2 ) 
  [\mathcal{N}( s_i | 0, \gamma^2 \sigma^2)]^{q_i}  [\mathcal{N}( s_i
| 0, \epsilon^2 \sigma^2)]^{1-q_i}
 [P_{\tH}^{q_i}  (1-P_{\tH})^{1-q_i}]^{q_{\pi(i)}} 
 \notag
 \\
 & & \cdot
  [P_{\tL}^{q_i}  (1 - P_{\tL})^{1-q_i}]^{1-q_{\pi(i)}} \biggl[\prod_{k \in
\ch (i)} \mu_k^{\tu} (0)\biggr]^{1 - q_i}  \biggl[\prod_{k \in \ch (i)}
\mu_k^{\tu} (1)\biggr]^{q_i}
 \Biggr\}.
 \label{eq:upwardmsgremaining}
\end{IEEEeqnarray}
For $q_{\pi(i)}=0$, we have
\begin{subequations}
\begin{IEEEeqnarray}{rCl}
  m_{i \rightarrow \pi(i)}(0) &=& \alpha_1  \max\Biggl\{ (1-P_{\tL})  \biggl[\prod_{k \in \ch (i)}
\mu_k^{\tu} (0)\biggr]  \exp\biggl( - 0.5  \frac{z_i^2}{
  \sigma^2+\sigma^2 \epsilon^2}\biggr) \big/ \epsilon,
  \notag
  \\
&&  P_{\tL}  \biggl[\prod_{k \in \ch (i)} \mu_k^{\tu} (1)\biggr] 
\exp\biggl( - 0.5  \frac{z_i^2}{
  \sigma^2+\sigma^2 \gamma^2}\biggr) \big/ \gamma \Biggr\}
\end{IEEEeqnarray}
and, for $q_{\pi(i)}=1$, we have
\begin{IEEEeqnarray}{rCl}
  m_{i \rightarrow \pi(i)}(1) &=& \alpha_1  \max\Biggl\{ (1-P_{\tH})  \biggl[\prod_{k \in \ch (i)}
\mu_k^{\tu} (0)\biggr]  \exp\Bigl( - 0.5  \frac{z_i^2}{
  \sigma^2+\sigma^2 \epsilon^2}\Bigr) \big/ \epsilon,
\notag
\\
&&  P_{\tH}  \biggl[\prod_{k \in \ch (i)} \mu_k^{\tu} (1)\biggr] 
\exp\Bigl( - 0.5  \frac{z_i^2}{
  \sigma^2+\sigma^2 \gamma^2}\Bigr) \big/ \gamma \Biggr\}
\end{IEEEeqnarray}
\end{subequations}
where we have used \eqref{eq:appmaximizedproductofGaussians} with
$\tau^2=\sigma^2 \epsilon^2$ and $\tau^2=\sigma^2 \gamma^2$ and
$\alpha_1 > 0$ is an appropriate normalizing constant. It follows
from \eqref{eq:appmodeofproductofGaussians} that the only two
candidates of $\bm{\theta}_i$ to maximize
\eqref{eq:upwardmsgremaining} are $[0, \widehat{s}_i(0)]^T$ and $[1,
\widehat{s}_i(1)]^T$.

\subsection{Downward Messages}
\label{app:Downwardmessagederivation}

\noindent
Based on the results in Section~\ref{subsubsec:upmsg} and
Appendix~\ref{app:Upwardmessagederivation}, we simplify the product of
upward messages sent from the siblings of node $i$ in
(\ref{eq:downwardmsg}) as follows [see \eqref{eq:msgupwardform}]:
\begin{equation}
  \label{eq:upchildren}
  \prod_{k \in \sib (i)} m_{k \rightarrow \pi(i)} (q_{\pi (i)}) = \biggl[\prod_{k \in \sib
    (i)} \mu^{\tu}_k (0)\biggr]^{1-q_{\pi (i)}}   \biggl[\prod_{k \in \sib
    (i)} \mu^{\tu}_k (1)\biggr]^{q_{\pi (i)}}
\end{equation}
see also Fig.~\ref{fig:msgdown}.

\subsubsection{Downward Messages from Root Nodes}

\noindent For the node $\pi (i) \in \cT_{\troot}$, we set the
message $m_{\gp(i) \rightarrow \pi(i)}(q_{\pi(i)})$ to one,
yielding [see \eqref{eq:downwardmsg}]
\begin{IEEEeqnarray}{rCl}
  \label{eq:downwardmsgroot}
  m_{\pi(i) \rightarrow i}( q_i ) = \alpha \max_{\bm{\theta}_{\pi(i)}}
  \biggl\{ \psi_{\pi(i)}(\bm{\theta}_{\pi(i)})  \psi_{i,\pi(i)}( q_i,
  q_{\pi(i)})  \prod_{k \in \sib(i)} m_{k \rightarrow \pi(i)}(
  q_{\pi(i)} ) \biggr\}.
\end{IEEEeqnarray}
Substituting \eqref{eq:upchildren} into \eqref{eq:downwardmsgroot}
yields
\begin{IEEEeqnarray}{rCl}
\label{eq:downwardmsgroot2}
 m_{\pi(i) \rightarrow i}( q_i ) &=&  \alpha  \max_{\bm{\theta}_{\pi(i)}}
\biggl\{ \mathcal{N}( z_{\pi (i)} |
  s_{\pi (i)}, \sigma^2 )  [P_{\troot} \mathcal{N}( s_{\pi (i)}
 | 0, \gamma^2 \sigma^2)]^{q_{\pi (i)}}  [(1 - P_{\troot})
  \mathcal{N}( s_{\pi (i)} | 0, \epsilon^2 \sigma^2)]^{1-q_{\pi
      (i)}} \notag
  \\ & &
  \cdot [P_{\tH}^{q_i}  (1-P_{\tH})^{1-q_i}]^{q_{\pi(i)}} 
  [P_{\tL}^{q_i}  (1 - P_{\tL})^{1-q_i}]^{1-q_{\pi(i)}}  \biggl[\prod_{k
    \in \sib (i)} \mu_k^{\tu} (0)\biggr]^{1 - q_{\pi (i)}}  \biggl[\prod_{k \in
    \sib (i)} \mu_k^{\tu} (1)\biggr]^{q_{\pi (i)}} \biggr\}.
\hspace{0.35in}
\end{IEEEeqnarray}
For $q_i = 0$, we have
\begin{subequations}
\begin{IEEEeqnarray}{rCl}
  m_{\pi(i) \rightarrow i}( 0 ) &=& \alpha_1 \max \biggl\{ (1 -
  P_{\troot}) (1 - P_{\tL}) \biggl[ \prod_{k \in \sib (i)} \mu_k^{\tu} (0)\biggr]
  \exp\Bigl( - 0.5 \frac{z_{\pi (i)}^2}{ \sigma^2+\sigma^2 \epsilon^2}
  \Bigr) \big/ \epsilon, \notag
  \\
  & & P_{\troot} (1-P_{\tH}) \biggl[ \prod_{k \in \sib (i)} \mu_k^{\tu} (1)\biggr]
  \exp\Bigl( - 0.5 \frac{z_{\pi (i)}^2}{ \sigma^2+\sigma^2 \gamma^2}
  \Bigr) \big/ \gamma \biggr\}
\end{IEEEeqnarray}
and for $q_i = 1$, we have
\begin{IEEEeqnarray}{rCl}
m_{\pi(i) \rightarrow i}( 1 ) &=& \alpha_1 \max \biggl\{ (1 - P_{\troot})  P_{\tL}  \biggl[ \prod_{k \in
\sib (i)} \mu_k^{\tu} (0)\biggr] 
\exp\Bigl( - 0.5  \frac{z_{\pi (i)}^2}{
  \sigma^2+\sigma^2 \epsilon^2} \Bigr) \big/ \epsilon, 
  \notag
  \\
  & &  P_{\troot}  P_{\tH}  \biggl[ \prod_{k \in \sib (i)} \mu_k^{\tu}
  (1)\biggr] \exp\Big( - 0.5  \frac{z_{\pi (i)}^2}{
  \sigma^2+\sigma^2 \gamma^2} \Big)
\big/ \gamma
 \biggr\}
\end{IEEEeqnarray}
\end{subequations}
where we have used \eqref{eq:appmaximizedproductofGaussians} with
$\tau^2=\sigma^2 \epsilon^2$ and $\tau^2=\sigma^2 \gamma^2$ and
 $\alpha_1 > 0$ is an appropriate normalizing constant.
It follows
from \eqref{eq:appmodeofproductofGaussians} that the only two
candidates of $\bm{\theta}_{\pi(i)}$ to maximize \eqref{eq:downwardmsgroot2} are
$[0, \widehat{s}_{\pi(i)}(0)]^T$ and $[1,
\widehat{s}_{\pi(i)}(1)]^T$.

\subsubsection{Downward Messages from Non-Root Nodes}

\noindent
For the node $\pi (i) \in (\cT \backslash \cT_{\troot})
\backslash \cT_{\sleaf}$, using the same strategy as above,
\eqref{eq:downwardmsg} simplifies as
\begin{IEEEeqnarray}{rCl}
m_{\pi(i) \rightarrow i}( q_i ) &=& \alpha  \max_{\bm{\theta}_{\pi(i)}}
  \biggl\{ \mathcal{N}( z_{\pi (i)} | s_{\pi (i)}, \sigma^2 ) 
  [ \mathcal{N}( s_{\pi (i)} | 0, \gamma^2 \sigma^2)]^{q_{\pi (i)}}  [ \mathcal{N}(
  s_{\pi (i)} | 0, \epsilon^2 \sigma^2)]^{1-q_{\pi (i)}}
 \notag
 \\
 & & \cdot [P_{\tH}^{q_i}  (1-P_{\tH})^{1-q_i}]^{q_{\pi(i)}} 
  [P_{\tL}^{q_i}  (1 - P_{\tL})^{1-q_i}]^{1-q_{\pi(i)}}  \biggl[\prod_{k \in
\sib (i)} \mu_k^{\tu} (0)\biggr]^{1 - q_{\pi (i)}}  \biggl[\prod_{k \in
\sib (i)} \mu_k^{\tu} (1)\biggr]^{q_{\pi (i)}}
\notag
\\
& & \cdot [\mu_{\pi (i)}^{\td}(0)]^{1-q_{\pi (i)}} 
[\mu^{\td}_{\pi (i)}(1)]^{q_{\pi (i)}} 
 \biggr\} .
\label{eq:downwardmsgnonroot}
\end{IEEEeqnarray}
For $q_i = 0$, we have
\begin{subequations}
\begin{IEEEeqnarray}{rCl}
  m_{\pi(i) \rightarrow i}( 0 ) &=& \alpha_1 \max \biggl\{\mu_{\pi
    (i)}^{\td}(0) (1 - P_{\tL}) \biggl[ \prod_{k \in \sib (i)}
  \mu_k^{\tu} (0)\biggr] \exp\Bigl( - 0.5 \frac{z_{\pi (i)}^2}{
    \sigma^2+\sigma^2 \epsilon^2} \Bigr) \big/ \epsilon, \notag
  \\
  & & \mu_{\pi (i)}^{\td}(1) (1-P_{\tH}) \biggl[ \prod_{k \in \sib
    (i)} \mu_k^{\tu} (1)\biggr] \exp\Bigl( - 0.5 \frac{z_{\pi (i)}^2}{
    \sigma^2+\sigma^2 \gamma^2} \Bigr) \big/ \gamma \biggr\}
\end{IEEEeqnarray}
and for $q_i = 1$, we have
\begin{IEEEeqnarray}{rCl}
m_{\pi(i) \rightarrow i}( 1 ) &=& \alpha_1 \max \biggl\{\mu_{\pi (i)}^{\td}(0)  P_{\tL}  \biggl[ \prod_{k \in
\sib (i)} \mu_k^{\tu} (0)\biggr]  \exp\Bigl( - 0.5  \frac{z_{\pi
(i)}^2}{ \sigma^2+\sigma^2 \epsilon^2} \Bigr) \big/ \epsilon, \;
  \notag
  \\
  & & \mu_{\pi (i)}^{\td}(1)  P_{\tH}  \biggl[ \prod_{k \in \sib (i)} \mu_k^{\tu}
  (1)\biggr] \exp\Bigl( - 0.5  \frac{z_{\pi (i)}^2}{
  \sigma^2+\sigma^2 \gamma^2} \Bigr) \big/ \gamma
 \biggr\}
\end{IEEEeqnarray}
\end{subequations}
where we have used \eqref{eq:appmaximizedproductofGaussians} with
$\tau^2=\sigma^2 \epsilon^2$ and $\tau^2=\sigma^2 \gamma^2$ and
$\alpha_1 > 0$ is an appropriate normalizing constant. It follows
from \eqref{eq:appmodeofproductofGaussians} that the only two
candidates to maximize \eqref{eq:downwardmsgnonroot} are $[0,
\widehat{s}_{\pi(i)}(0)]^T$ and $[1, \widehat{s}_{\pi(i)}(1)]^T$.

\subsection{Beliefs}
\label{app:Beliefsderivation}

\noindent Define the vector $\bbeta_i = [\beta_i (0), 
\beta_i (1)]^T$ as
\begin{equation}
  \label{eq:betai}
  \beta_i(0) = \max_{s_i} b( [0, s_i]^T ), \quad \beta_i(1) =
  \max_{s_i} b( [1, s_i]^T )
\end{equation}
where $b(\bm{\theta}_i)$ are the beliefs defined in \eqref{eq:belief}.

\subsubsection{Beliefs for the Root Nodes}

\noindent
For root nodes $i \in \cT_{\troot}$, the beliefs $b(\bm{\theta}_i)$ in
\eqref{eq:belief} become
\begin{IEEEeqnarray}{rCL}
  \label{eq:beliefroot}
  b(\bm{\theta}_i) &=& \alpha  \mathcal{N}( z_i | s_i, \sigma^2 )  [P_{\troot} 
  \mathcal{N}( s_i | 0, \gamma^2 \sigma^2 )]^{q_i} 
  [(1-P_{\troot})  \mathcal{N}( s_i | 0, \epsilon^2 \sigma^2
  )]^{1-q_i}
\notag
\\
& & \cdot  \big[\prod_{k \in \ch (i)}
\mu^{\tu}_k(0)\big]^{1-q_i} 
  \big[\prod_{k \in \ch (i)} \mu^{\tu}_k(1)\big]^{q_i}.
\end{IEEEeqnarray}
and \eqref{eq:betai} simplify to [see \eqref{eq:appmaximizedproductofGaussians}]
\begin{subequations}
\begin{IEEEeqnarray}{rCL}
  \label{eq:beliefroot2}
  \beta_i(0) &=& \alpha  \frac{1}{\sqrt{2  \uppi  \sigma^2 }  \sqrt{2 
      \uppi  \epsilon^2 \sigma^2 }}  \exp\big( - 0.5  \frac{z_i^2}{
  \sigma^2+\sigma^2 \epsilon^2}\big)  (1-P_{\troot})  \prod_{k \in \ch (i)} \mu^{\tu}_k(0)
\\
 \beta_i(1) &=&  \alpha  \frac{1}{\sqrt{2  \uppi  \sigma^2 }  \sqrt{2 
      \uppi  \gamma^2 \sigma^2 }}  \exp\big( - 0.5  \frac{z_i^2}{
  \sigma^2+\sigma^2 \gamma^2}\big)  P_{\troot}   \prod_{k \in \ch (i)} \mu^{\tu}_k(1)
\end{IEEEeqnarray}
\end{subequations}
yielding $\bbeta_i = [\beta_i (0), \beta_i (1) ]^T = \alpha_1 [ 1 -
P_{\troot}, P_{\troot} ]^T \odot \bphi(z_i)  \odot
\etabold_i^{\tu}$.
\vspace{0.1in}

\subsubsection{Beliefs for the Non-Root Non-Leaf Nodes}

\noindent
For $i \in ( \cT  \backslash  \cT_{\troot} )  \backslash 
\cT_{\sleaf}$, the beliefs $b(\bm{\theta}_i)$ in \eqref{eq:belief} become
\begin{IEEEeqnarray}{rCL}
  \label{eq:beliefnorootnoleaf}
  b(\bm{\theta}_i) &=& \alpha \mathcal{N}( z_i | s_i, \sigma^2 )
  [\mathcal{N}( s_i | 0, \gamma^2 \sigma^2 )]^{q_i} [\mathcal{N}( s_i
  | 0, \epsilon^2 \sigma^2 )]^{1-q_i} [\mu_i^{\td}(0)]^{1-q_i}
  [\mu^{\td}_i(1)]^{q_i} \notag
  \\
  & & \cdot \biggl[\prod_{k \in \ch (i)} \mu^{\tu}_k(0)\biggr]^{1-q_i}
  \biggl[\prod_{k \in \ch (i)} \mu^{\tu}_k(1)\biggr]^{q_i}
\end{IEEEeqnarray}
and \eqref{eq:betai} simplify to [see \eqref{eq:appmaximizedproductofGaussians}]
\begin{subequations}
\begin{IEEEeqnarray}{rCL}
  \label{eq:beliefnorootnoleaf2}
  \beta_i(0) &=& \alpha \frac{1}{\sqrt{2 \uppi \sigma^2 } \sqrt{2
      \uppi \epsilon^2 \sigma^2 }} \exp\Bigl( - 0.5 \frac{z_i^2}{
    \sigma^2+\sigma^2 \epsilon^2}\Bigr) \mu_i^{\td}(0) \prod_{k \in
    \ch (i)} \mu^{\tu}_k(0)
  \\
  \beta_i(1) &=& \alpha \frac{1}{\sqrt{2 \uppi \sigma^2 } \sqrt{2
      \uppi \gamma^2 \sigma^2 }} \exp\Bigl( - 0.5 \frac{z_i^2}{
    \sigma^2+\sigma^2 \gamma^2}\Bigr) \mu_i^{\td}(1) \prod_{k \in \ch
    (i)} \mu^{\tu}_k(1)
\end{IEEEeqnarray}
\end{subequations}
yielding $\bbeta_i = [\beta_i (0), \beta_i (1) ]^T = \alpha_1 \bphi(z_i)
\odot \bmu_i^{\td} \odot \etabold_i^{\tu}$.

\subsubsection{Beliefs for the Leaf Nodes}

\noindent
For $i \in \cT_{\sleaf}$, the beliefs $b(\bm{\theta}_i)$ in
\eqref{eq:belief} become
\begin{IEEEeqnarray}{rCL}
  \label{eq:beliefleaf}
  b(\bm{\theta}_i) &=& \alpha  \mathcal{N}( z_i | s_i, \sigma^2 ) 
  [\mathcal{N}( s_i | 0, \gamma^2 \sigma^2 )]^{q_i} 
  [\mathcal{N}( s_i | 0, \epsilon^2 \sigma^2 )]^{1-q_i} 
[\mu_i^{\td}(0)]^{1-q_i}  [\mu^{\td}_i(1)]^{q_i} \notag
  \\
\end{IEEEeqnarray}
and \eqref{eq:betai} simplify to [see \eqref{eq:appmaximizedproductofGaussians}]
\begin{subequations}
\begin{IEEEeqnarray}{rCL}
  \label{eq:beliefleaf2}
  \beta_i(0) &=& \alpha  \frac{1}{\sqrt{2  \uppi  \sigma^2 }  \sqrt{2 
      \uppi  \epsilon^2 \sigma^2 }}  \exp\Bigl( - 0.5  \frac{z_i^2}{
  \sigma^2+\sigma^2 \epsilon^2}\Bigr)  \mu_i^{\td}(0)
\\
 \beta_i(1) &=&  \alpha  \frac{1}{\sqrt{2  \uppi  \sigma^2 }  \sqrt{2 
      \uppi  \gamma^2 \sigma^2 }}  \exp\Bigl( - 0.5  \frac{z_i^2}{
  \sigma^2+\sigma^2 \gamma^2}\Bigr)  \mu_i^{\td}(1)
\end{IEEEeqnarray}
\end{subequations}
yielding $\bbeta_i = [\beta_i (0), \beta_i (1) ]^T =  \alpha_1 \bphi(z_i)
\odot \bmu_i^{\td}$.

Consequently, the mode $\widehat{\bm{\theta}}_i$ is computed as
\begin{equation}
  \widehat{\bm{\theta}}_i =  \arg
  \max_{\bm{\theta}_i} b(\bm{\theta}_i) =
\begin{cases} [1,\widehat{s}_i(1)]^T, & \beta_i (1)
\geq \beta_i (0) \\ [0,\widehat{s}_i(0)]^T, & \text{otherwise}
\end{cases}
\end{equation}
which follows from \eqref{eq:appmodeofproductofGaussians}.

Note that the normalizing constants $\alpha$ and $\alpha_1$ in
the above upward and downward messages and beliefs have been set so that $m_{i
  \rightarrow \pi(i)}( 0 ) + m_{i \rightarrow \pi(i)}( 1 ) = 1$,
$m_{\pi(i) \rightarrow i}( 0 ) + m_{\pi(i) \rightarrow i}( 1 ) = 1$,
and $\beta_i (0) + \beta_i (1) = 1$ respectively.

\nocite{sd12SPIE}

\bibliography{IEEEabrv,bpem}   %

\end{document}